\documentclass[letterpaper, 10 pt, conference]{ieeeconf}  

\IEEEoverridecommandlockouts                              

\usepackage{amsmath}
\usepackage{amssymb}
\usepackage{graphicx}
\usepackage{textcomp}
\usepackage{xcolor}
\usepackage{pgfplots}
\usepackage{subfigure}
\usepgfplotslibrary{groupplots}
\usepackage{tikz}
\usepackage{tikzscale}
\usepackage{cleveref}
\usepackage{float}
\usepackage{multirow}
\usepackage{algorithm}
\usepackage{algpseudocode} 
\usepackage{siunitx}
\usepackage{comment}
\usepackage{booktabs}
\usepackage{alphalph}

\crefname{figure}{Fig.}{Fig.}
\Crefname{figure}{Fig.}{Fig.}
\crefname{equation}{Eq.}{Eq.}
\Crefname{equation}{Eq.}{Eq.}

\newcommand{\ie}{\textit{i}.\textit{e}., }
\newcommand{\eg}{\textit{e}.\textit{g}., }
\newcommand{\st}{\text{s.t. }}

\title{\LARGE \bf
Proactive Human-Robot Co-Assembly: \\ Leveraging Human Intention Prediction and Robust Safe Control
}

\author{Ruixuan Liu$^{1}$, Rui Chen$^{1}$, Abulikemu Abuduweili$^{1}$ and Changliu Liu$^{1}$ 
\thanks{*This work is in part supported by Ford Motor Company.}
\thanks{$^{1}$Ruixuan Liu, Rui Chen, Abulikemu Abuduweili, and Changliu Liu are with Robotics Institute,
	Carnegie Mellon University,
	Pittsburgh, PA, 15213, USA.
        {\tt\small ruixuanl, ruic3, abulikea, cliu6@andrew.cmu.edu}}%
}

\begin{document}

\maketitle
\thispagestyle{empty}
\pagestyle{empty}

\begin{abstract}
Human-robot collaboration (HRC) is one key component to achieving flexible manufacturing to meet the different needs of customers. 
However, it is difficult to build intelligent robots that can proactively assist humans in a safe and efficient way due to several challenges.
First, it is challenging to achieve efficient collaboration due to diverse human behaviors and data scarcity.
Second, it is difficult to ensure interactive safety due to uncertainty in human behaviors.
This paper presents an integrated framework for proactive HRC.
A robust intention prediction module, which leverages prior task information and human-in-the-loop training, is learned to guide the robot for efficient collaboration. 
The proposed framework also uses robust safe control to ensure interactive safety under uncertainty.
The developed framework is applied to a co-assembly task using a Kinova Gen3 robot.
The experiment demonstrates that our solution is robust to environmental changes as well as different human preferences and behaviors. In addition, it improves task efficiency by approximately 15-20\%.
Moreover, the experiment demonstrates that our solution can guarantee interactive safety during proactive collaboration.
\end{abstract}

\section{INTRODUCTION}
\begin{figure}
\subfigure[\footnotesize Co-assembly \cite{7487476}.]{\includegraphics[width=0.32\linewidth]{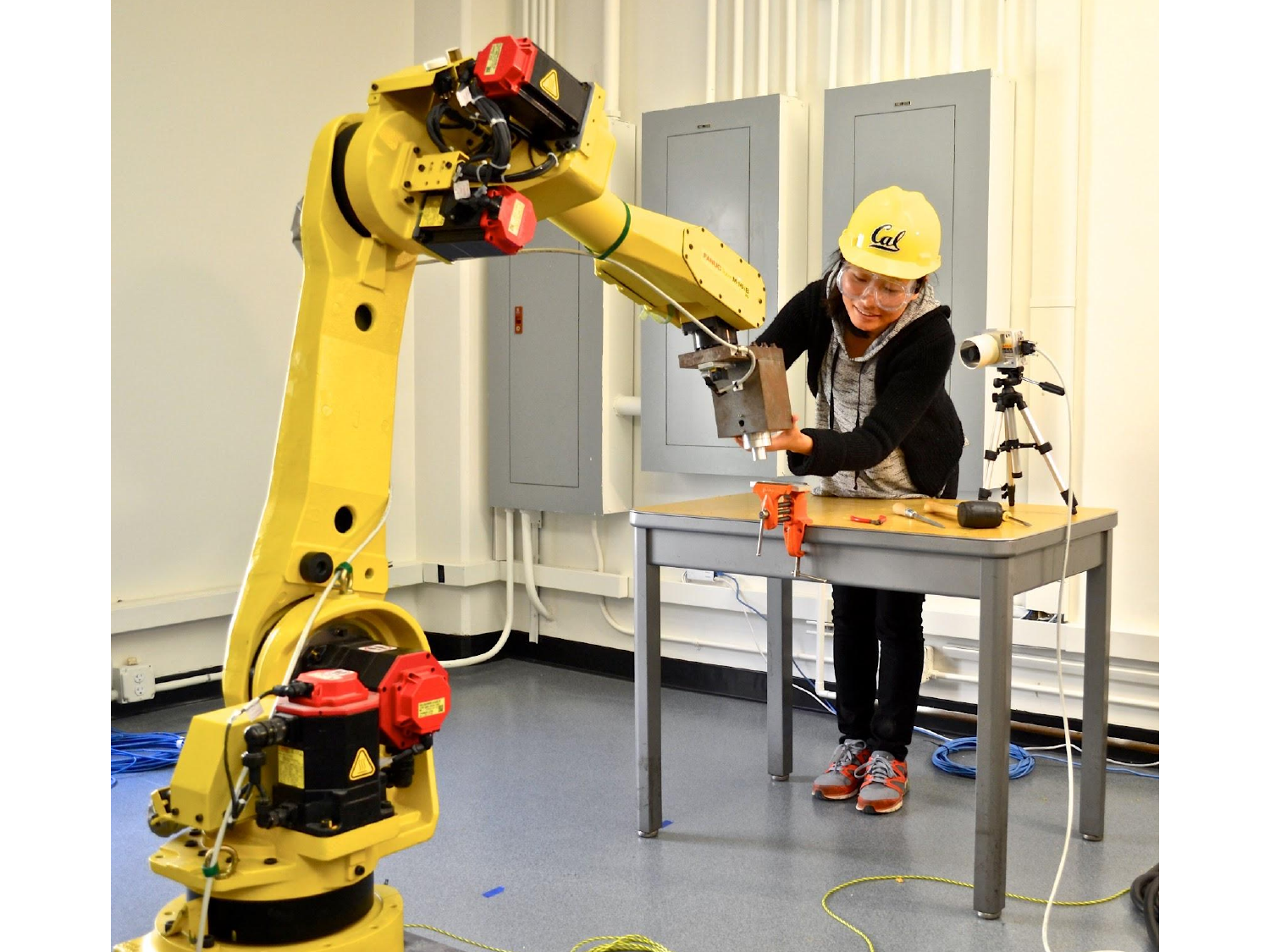}\label{fig:hri1}}\hfill
\subfigure[\footnotesize Co-handling \cite{cohandle}.]{\includegraphics[width=0.32\linewidth]{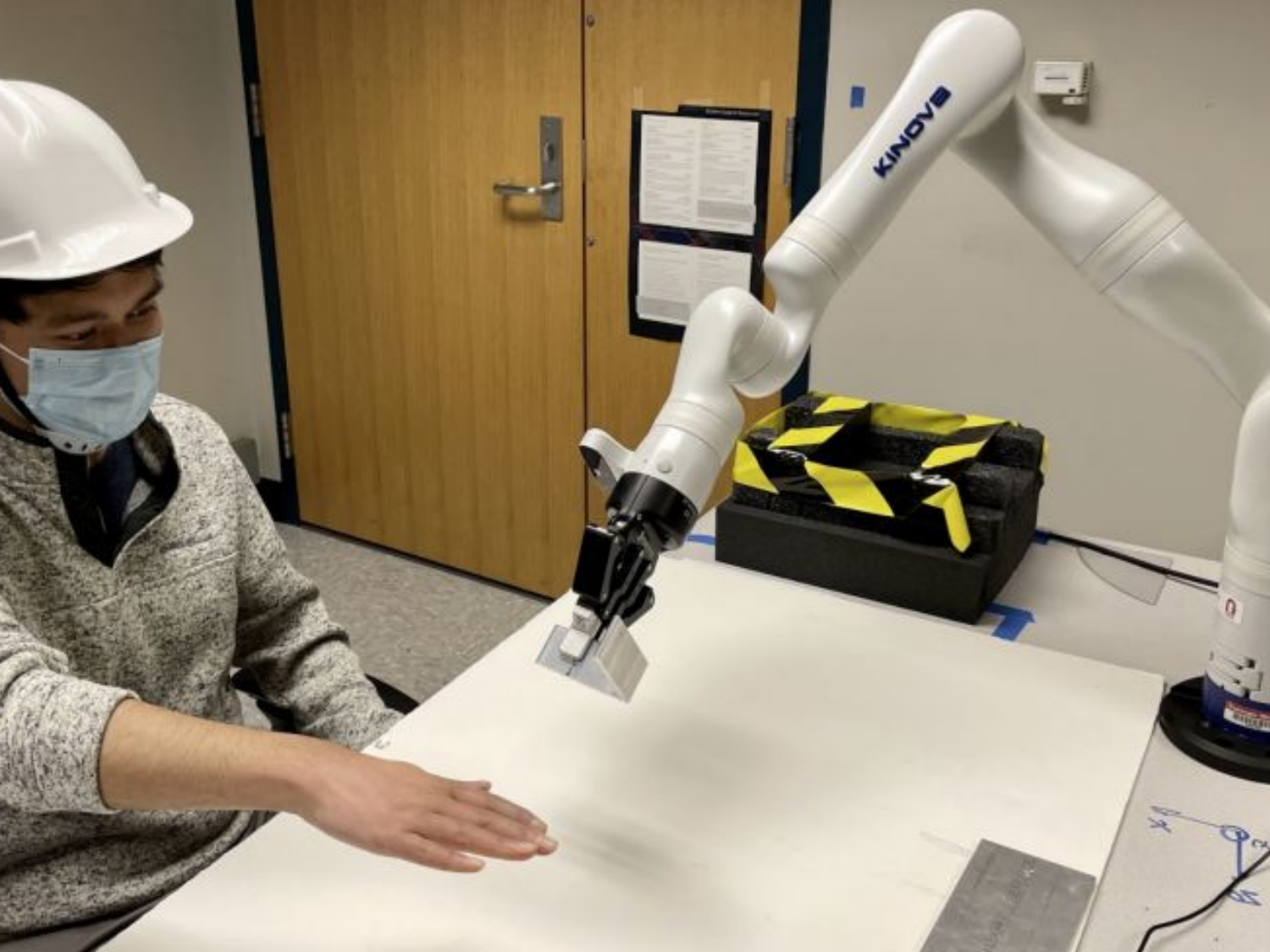}\label{fig:hri2}}\hfill
\subfigure[\footnotesize Handover \cite{jssa, jpc}.]{\includegraphics[width=0.32\linewidth]{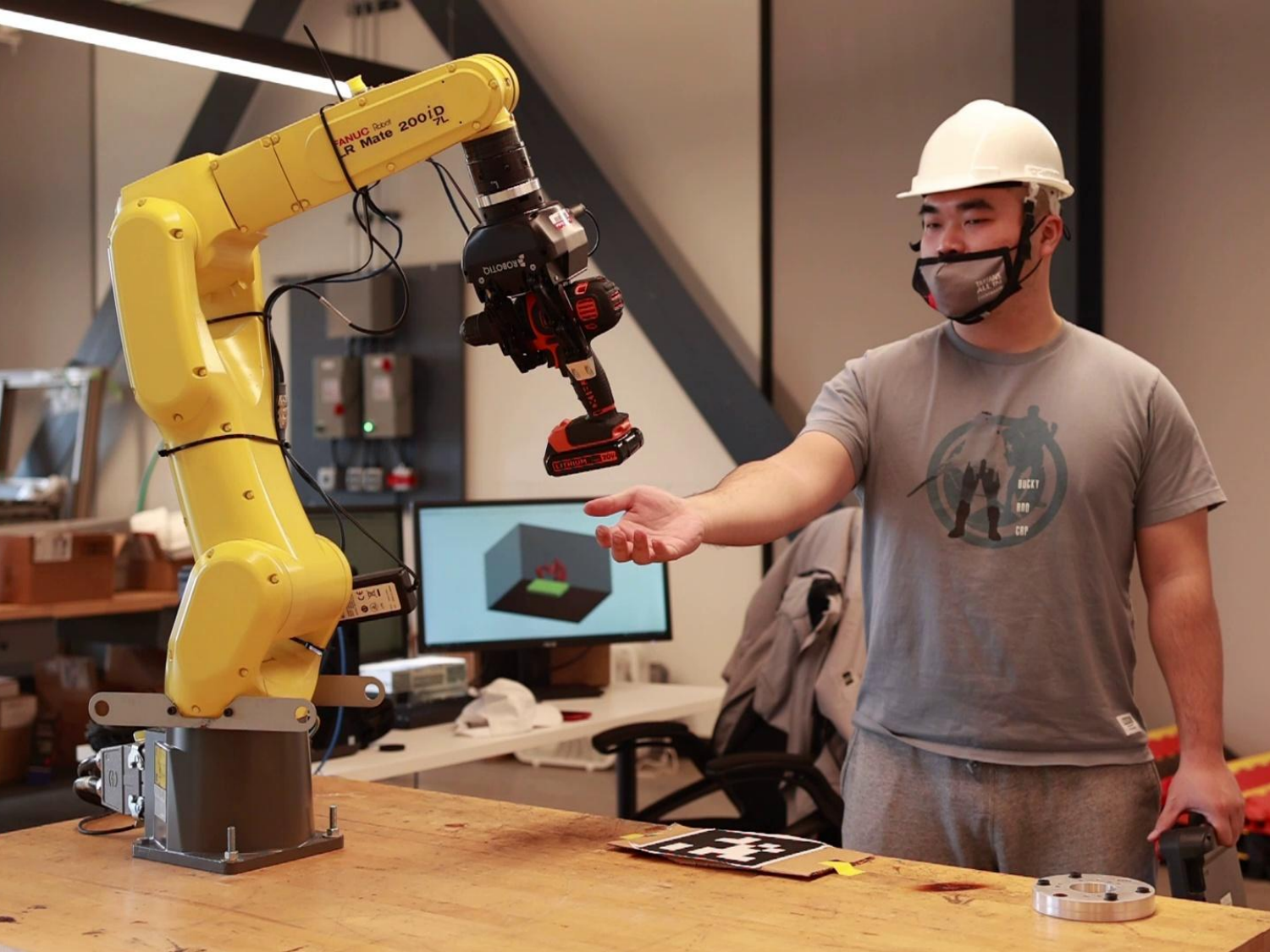}\label{fig:hri3}}
\vspace{-5pt}
    \caption{\footnotesize Examples of HRC. (a) The human and the robot assemble an object collaboratively. (b) The human uses gestures to control the robot to manipulate an object, which is not directly operable by the human. (c) The robot delivers the human-desired tool to the human partner. \label{fig:hri}}
    \vspace{-20pt}
\end{figure}

In contemporary manufacturing, robots are widely used in many applications, such as automotive assembly \cite{MICHALOS201081} and additive manufacturing \cite{URHAL2019335}.
However, the need for mass customization is calling for robots to work efficiently with human workers (HRC) to improve the efficiency and flexibility of production lines and satisfy the various needs of the customers.
HRC can be deployed in various situations as shown in \cref{fig:hri}. In \cref{fig:hri1}, the robot co-assembles an object with the human; in \cref{fig:hri2}, the human uses gestures to jointly handle an object with a robot; in \cref{fig:hri3}, the robot delivers a tool (\eg a power drill) to the human.

In fact, HRC has recently been widely studied \cite{WANG2019701, WANG20175, MICHALOS2018194}.
However, there remain many challenges in deploying HRC in real-world applications.
One major challenge is \textbf{intent recognition} \cite{ctx7484641100004436}, where the robot needs to understand human behaviors in order to better collaborate.
Existing works \cite{8990024,  CASALINO2018194} use intention prediction to schedule and plan the robot collaboration strategy.
However, different humans have different behaviors in different circumstances.
The diverse and complex nature of human behaviors makes it challenging to construct an intent recognition model that is \textbf{robust} to different humans.
It is expensive (if not impossible) to obtain human demonstrations for all possible situations.
Some works \cite{9281312, abuduweili2019robust} leverage online adaptation techniques to fine-tune the prediction model. But due to the nature of intent recognition (\ie classification problem), it is difficult to obtain ground-truth labels online for adaptation.
The data scarcity (\ie limited human demonstrations) makes it challenging for robust intention prediction.

Another challenge for HRC is \textbf{interactive safety} under uncertainty (from both humans and the environment).
Conventional robots in industrial settings are strictly caged for safety concerns \cite{hrcinindustry}. 
To free robots, recent works set safety zones running different velocity profiles and halt the robot if anything gets too close \cite{s21124113}.
That approach might be too conservative and harm efficiency.
The recent emergence of collaborative robots has integrated force sensors to reduce the impact upon collision (\eg FANUC CR series).
However, reducing the collision impact only improves the worst case, rather than guaranteeing general interactive safety. 
It is desired that unexpected collisions can be avoided and the robot can collaborate safely in the dynamic environment.

To address the challenges, this paper presents an integrated framework to improve collaboration efficiency while guaranteeing interactive safety for HRC. 
Given the environmental information, our framework has an intention prediction module to understand human behaviors.
In particular, to tackle the data scarcity challenge, this paper leverages prior knowledge to infer task progress, and human-in-the-loop data augmentation to learn robust intention classification models.
Based on the inference of human intention and task progress, the system plans the appropriate collaboration action.
To ensure interactive safety under uncertainty, the proposed framework uses robust safe control to safeguard the execution of the planned collaboration action in real-time.
To demonstrate our system, we perform a human-robot co-assembly task on a Kinova Gen3 robot.
Experiments show that the framework improves task efficiency by approximately 15-20\% on average. 
Moreover, our pipeline can guarantee interactive safety (\ie no collision) during active collaboration.
Finally, the proposed solution is robust to environmental changes as well as different human preferences and behaviors.

\begin{figure}
    \centering
    \includegraphics[width=\linewidth]{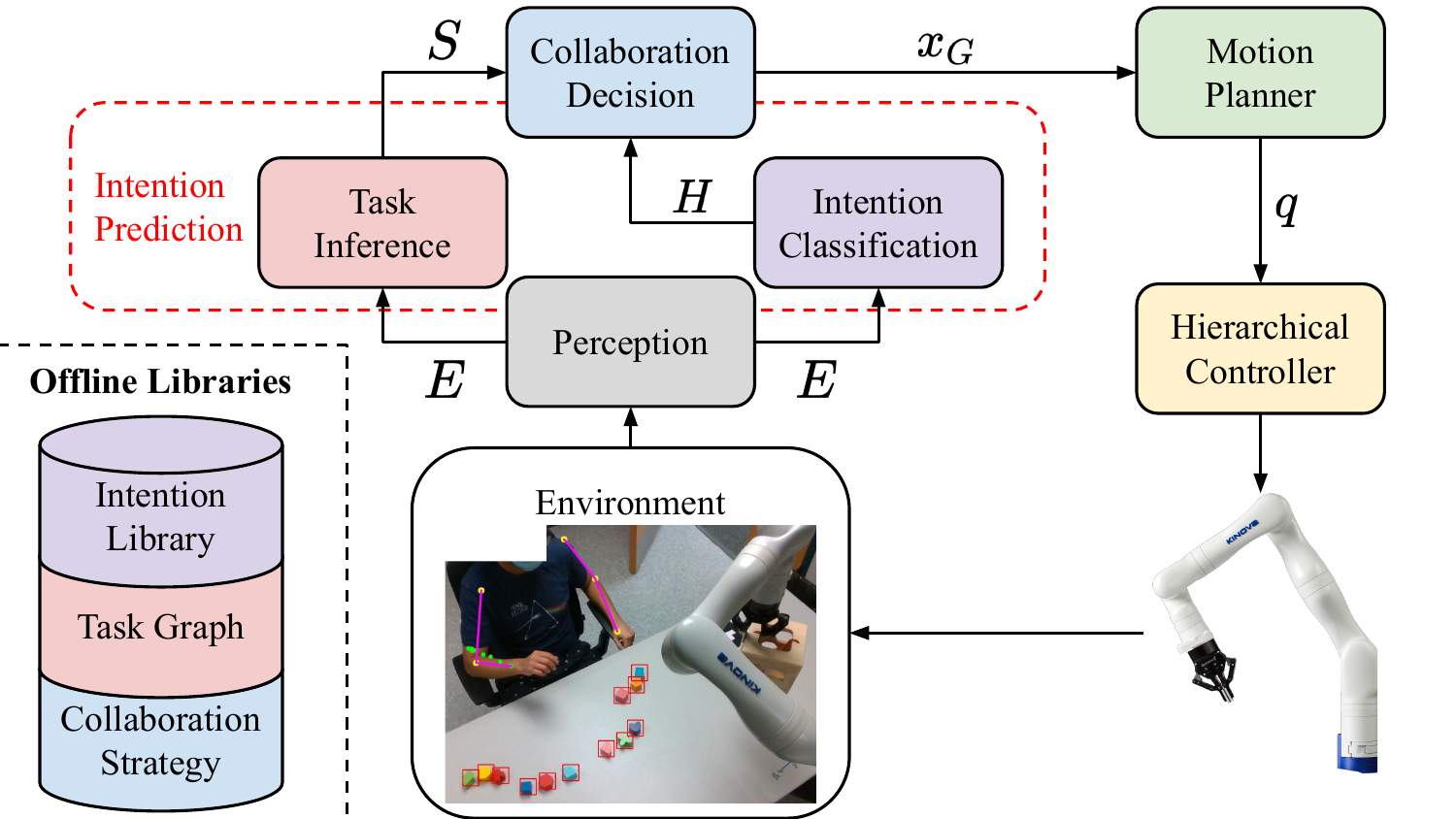}
    \caption{\footnotesize Illustration of the system architecture. \label{fig:archi}}
    \vspace{-20pt}
\end{figure}

\section{PROBLEM FORMULATION}
This paper considers the HRC settings where a single human collaborates with a single robot arm. 
In addition, this paper mainly considers assembly tasks that are done by sequential operations.
The robot collaboration task can be decomposed into multiple subtasks corresponding to the human sequential operations.
Each robot subtask can be formulated as constrained optimization \cite{taskagnostic}.
The objective is to find an efficient robot motion, $q=[q_1;q_2;\dots;q_n]$ where $n$ denotes the planning horizon, such that safety constraints are satisfied. 
The constrained optimization can be formulated as $\min_q~L_T(q, ~x_G)~ \st~  C_\mathrm{safety}$,
where $x_G$ denotes the robot task goal and $L_T$ represents the robot task-specific objective, \eg tracking cost function or travel time cost.
$C_\mathrm{safety}$ guarantees safety and encodes dynamics and kinematics constraints.
In the context of HRC for co-assembly, the constrained optimization can be expanded as
\begin{equation}\label{eq:prob}
    \begin{split}
        \min_q~&L_T(q, ~x_G)\\
        \st~&x_G=f(~\underbrace{\min_H~L_H(H, E), ~ \max_S~L_S(S)}_{H_{intention}}~),\\
        &\underbrace{\forall k\in[1,n], ~\mathbb{E}\left[d(q_k,E_k)\right]\geq d_{min}}_{C_\mathrm{safety}},
    \end{split}
\end{equation}
where $H_{intention}$ denotes the human intention. 
The function $f(\cdot)$ maps $H_{intention}$ to the corresponding robot collaboration strategy $x_G$.
Since the human intention is complex in nature, $H_{intention}$ is decomposed to human operation $H$ and task progress $S$.
$L_H$ and $L_S$ are the corresponding objective functions.
The outer optimization solves the optimal robot motion $q$ given the goal $x_G$. 
The inner optimizations predict human behavior and infer the task progress.
The safety constraint $C_{safety}$ ensures the solved robot motion $q$ is collision-free throughout the task.
$d(\cdot)$ is a distance function that calculates the minimum distance between the robot and the environment objects using the capsule representation \cite{7487476}.
$d_{min}$ is the user-specified safety margin. 
$E$ denotes the environment (\ie obstacles and the human).
$E_k$ is the environment at timestep $k$ obtained using perception algorithms, \eg OpenPose \cite{openpose} and YOLO \cite{yolov3}, as shown in \cref{fig:perception}. 
Notably, we consider \cref{eq:prob} in a probabilistic sense due to the uncertainty (\eg sensor uncertainty, sudden change of mind, etc) in human motions. 
Hence, we take expectations of the quantities in constraints over possible environment states.

\Cref{fig:archi} illustrates the architecture of the proposed framework.
Existing motion planners \cite{cfs, dscc_cfs} can solve $q$ in \cref{eq:prob} efficiently.
To achieve safe and efficient human-robot co-assembly, we need to ensure that 1) with data scarcity, $H_{intention}$ can be robustly predicted for different humans in different environments, and 2) $q$ can be safely executed, subject to $C_\mathrm{safety}$, in real-time under uncertainty.

\section{Human Intention Prediction}
Human intention prediction is essential to HRC since the robot needs to understand the human in order to intelligently collaborate \cite{10.1007/s11263-022-01594-9, SHARMA2022120,8990024,CASALINO2018194}.
However, human intention is difficult to predict since different humans have different behaviors in different circumstances (\eg task progress, environment setups, etc).
It is expensive (if not impossible) to obtain human demonstrations for all possible situations.
As a result, it is especially challenging to construct robust intention prediction due to data scarcity (\ie limited human demonstrations).
To tackle this challenge, this paper leverages prior task knowledge and human-in-the-loop training to improve the robustness of intention prediction.
In particular, as shown in \cref{fig:archi}, we decompose the intention prediction problem into two subtasks, 1) task inference, which encodes prior knowledge of the task, and 2) intention classification, which leverages human-in-the-loop data augmentation to learn robust intention classification model with scarce data.

\subsection{Task Progress Inference}
\begin{figure}
    \centering
    \includegraphics[width=1\linewidth]{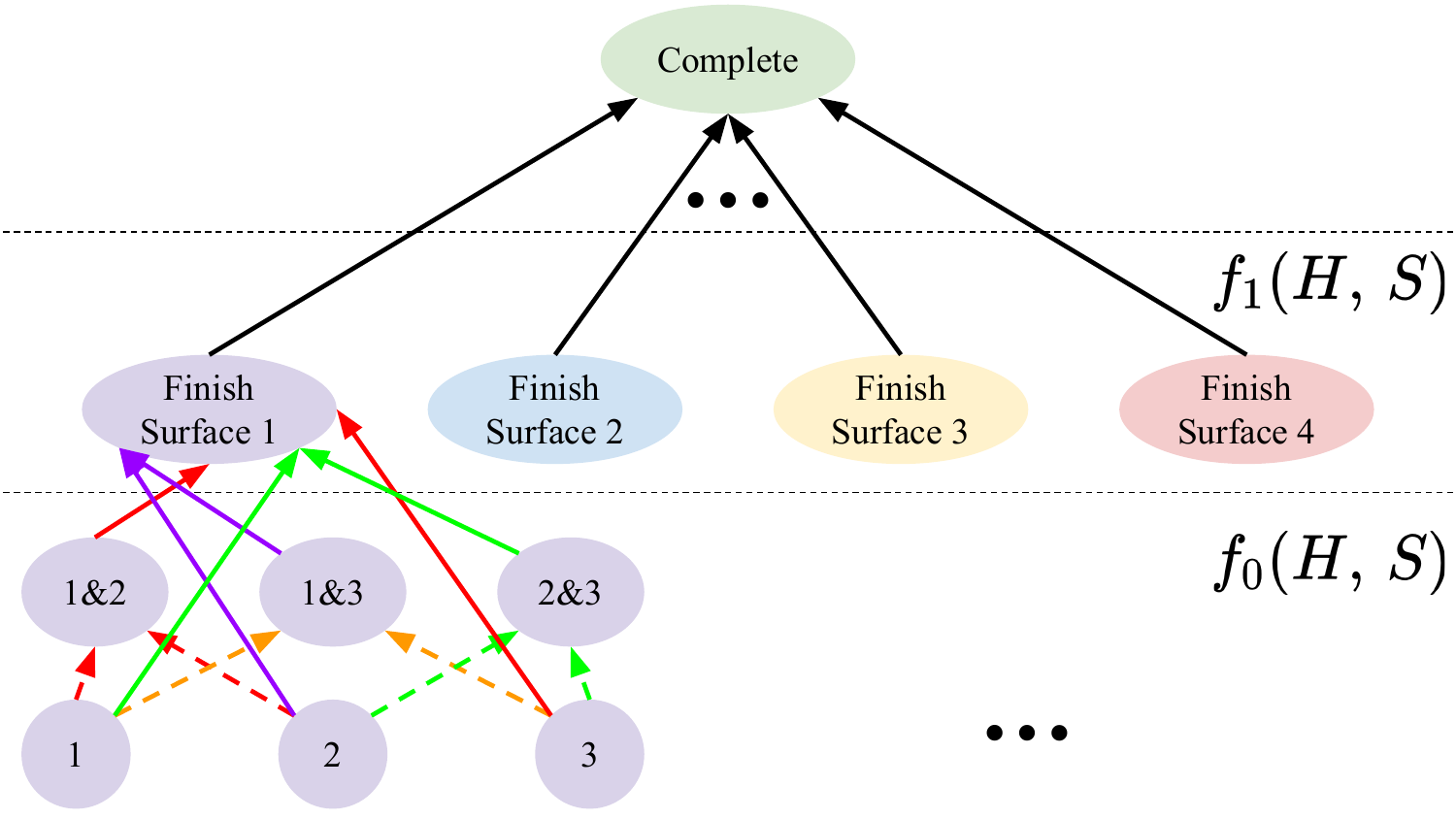}
    \caption{\footnotesize Illustration of the task graph corresponding to the co-assembly task in \cref{fig:container}. Identical arrows indicate AND connections. Different arrows at a node represent OR connections. $f_0(\cdot)$ and $f_1(\cdot)$ denote the collaboration policy corresponding to the nodes at the level of the graph. \label{fig:taskgraph}}
    \vspace{-20pt}
\end{figure}

Human intention heavily depends on the progress of the task.
This paper mainly considers assembly tasks that can be accomplished by sequential operations.
In real-world applications (\eg engine assembly), there are instructions that must be followed (\eg assemble sequence), while there also exists certain flexibility (\eg the order of fastening screws).
Therefore, we use AND/OR graphs to represent assembly plans \cite{54734}.
\Cref{fig:taskgraph} illustrates an example AND/OR task graph for the co-assembly task shown in \cref{fig:container}, in which the human inserts 12 different blocks into a container via holes on 4 surfaces.
Identical arrows indicate AND connections, while arrows with different colors or types denote OR connections.
The numbers in the nodes at the first two levels indicate the IDs of the blocks being inserted.

Given an assembly task, an AND/OR task graph is designed offline based on prior task knowledge and stored in the offline libraries as shown in \cref{fig:archi}.
The task graph is denoted as $S_T=\{S_1, S_2, \dots,S_{N_T}\}$, where $S_i$ denotes nodes in the graph that represent the task states and $N_T$ denotes the number of possible states.
The task progress $S\in S_T$ is determined by the maximization in \cref{eq:prob}, which can be written as
\begin{equation}\label{eq:taskgraphmax}
    \begin{split}
        \max_S~L_S(S) &=\max_S P(S|s_{1:i})\\
        &\propto P(s_{1:i}|S)P(S),
    \end{split}
\end{equation}
where $s_{1:i}$ denotes the prior task nodes sequence.
Given the environment observation $E$, \cref{eq:taskgraphmax} can be solved by tracing the task graph $S_T$ and $S$ can be inferred.

\subsection{Intention Classification}
Given the observation of the environment $E$, such as the human trajectory and the object poses, the intention classification outputs the intention type $H\in H_M=\{H_1, H_2,\dots,H_{N_H}\}$, where $N_H$ denotes the number of intention types and $H_M$ is the intention library defined offline as shown in \cref{fig:archi}.
In the co-assembly context, $H_i$ could be either the human reaching a certain object, assembling, disassembling, etc.
The intention $H\in H_M$ is solved by the inner minimization in \cref{eq:prob}, which can be written as 
\begin{equation}\label{eq:minhuman}
    \min_H ~L_H(H, H')~\st ~H=h_{\theta}(E),
\end{equation}
where $h_{\theta}(\cdot)$ denotes the classification function and $H'$ represents the ground-truth intention.
$L_H(\cdot)$ defines the similarity between $H_i$ and $H_j$.
Due to the complex nature of human behavior, this paper uses a data-driven method (\ie neural networks) to solve \cref{eq:minhuman}. 
However, due to the limited amount of training data, it is difficult to learn an $h_{\theta}(\cdot)$ that is robust in real-world deployment, \eg with different humans, different environments, etc.

To tackle the challenge, this paper uses iterative adversarial data augmentation (IADA) \cite{iada} to improve the model performance during offline training.
Given an initial training dataset $D_0=\{e\in E, h\in H_M\}$, IADA leverages neural network verification to attack and find the weakest parts of the current neural network model. 
Specifically, it finds adversarial data points and expands the original dataset. 
It queries human experts for the labels of the adversaries, and then, augments the training dataset with $D_0'$ (\ie expert verified data points) and $D_{adv}$ (pseudo-labeled data points) to improve the learned model performance.
Human verification is essential in learning $h_{\theta}(\cdot)$ since $e\in E$ is vulnerable to adversarial attacks. 
Slight perturbations would potentially change the intention encoded in the human operation.
Thus, it is important to leverage expert knowledge to ensure the model training correctness.

Given the recognized human intention $H$ and inferred task progress $S$, the robot goal $x_G=f(H,S)$ is determined, where $f(\cdot)$ is the task-specific collaboration policy defined offline (shown in \cref{fig:archi}).
\Cref{fig:taskgraph} shows examples of collaboration policies $f_0(\cdot)$ and $f_1(\cdot)$, which correspond to purple nodes at level 1 and 2, and the nodes at level 3 respectively.
In particular, the policies can be written as
\begin{equation}
    \begin{split}
        f_0(H, S)&=\begin{cases}
      Alert & H\notin \{R1\wedge R2 \wedge R3\}\\
      x_G=x_C & Otherwise
    \end{cases} \\
    \\
    f_1(H, S)&=\begin{cases}
      DS_1 & H\in \{R_1\vee R_2 \vee R_3\}\\
      DS_2 & H\in \{R_4\vee R_5 \vee R_6\}\\
      DS_3 & H\in \{R_7\vee R_8 \vee R_9\}\\
      DS_4 & H\in \{R_{10}\vee R_{11} \vee R_{12}\}\\
      x_G=x_C & Otherwise
    \end{cases} \\
    \end{split}
\end{equation}
where $x_C$ denotes the current robot task, $DS_i$ represents the robot task \textit{Displaying surface $i$}, and $R_i$ indicates the human intention \textit{Reaching block i}.  
The robot task $q$ is then solved in \cref{eq:prob} based on $x_G$ using a motion planner as shown in \cref{fig:archi}.

\begin{figure}
    \centering
    \includegraphics[width=0.8\linewidth]{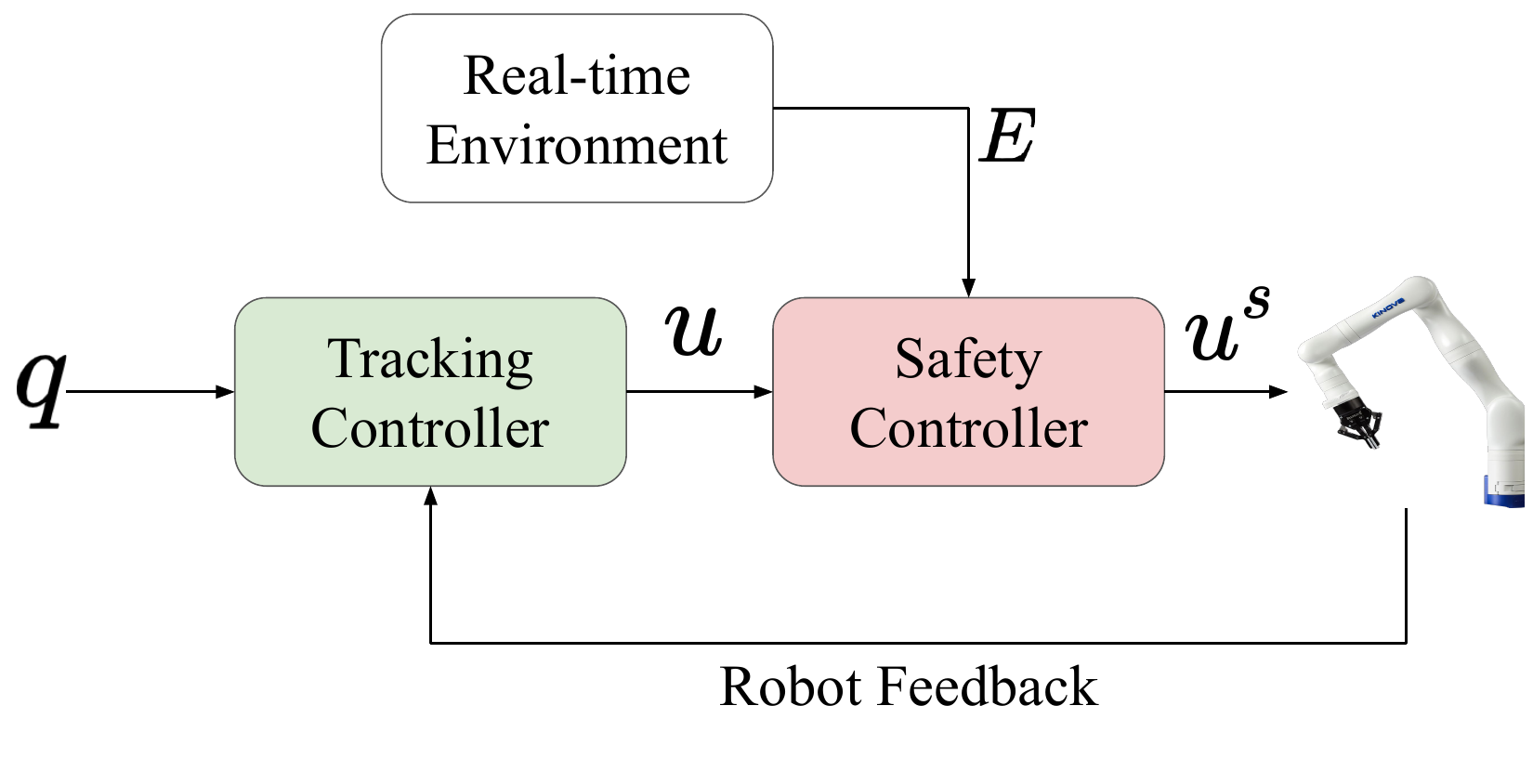}
    \vspace{-15pt}
    \caption{\footnotesize Illustration of the hierarchical controller architecture. \label{fig:controller}}
    \vspace{-20pt}
\end{figure}

\section{Hierarchical Controller}
Given the planned robot motion $q$, it is important that the controller executes $q$ safely in real time due to the dynamic environment (\ie human involvement and uncertainty). 
\Cref{fig:controller} illustrates the hierarchical design of the robot controller.
Given the robot task $q$ solved in \cref{eq:prob}, the hierarchical controller has a tracking controller to execute $q$. And a safety controller monitors the nominal control from the tracking controller and ensures interactive safety in real time according to the environment.
This paper uses the Kinova Gen3 robot, but the hierarchical controller design can be easily extended to other robot platforms.

\paragraph{Tracking Controller}
We implement the tracking controller as a feedback controller with an update rate of approximately 30Hz. 
Specifically, at a given time $k$, given a joint tracking reference $q_k$, the tracking controller receives the robot feedback and uses a PD controller to generate a desired joint acceleration $u$ as the nominal tracking control to execute the task.

\paragraph{Safety Controller}
It is highly possible that although $q$ is safe at the time when it is planned, it might lead to safety hazards during the execution due to the dynamic environment.
Note that solving $q$ in \cref{eq:prob} requires a non-trivial computation time. 
Therefore, it is important to safeguard the controller output in the control loop in order to ensure interactive safety.
The hierarchical controller has the safety controller considering the real-time environment status and uncertainty to safeguard the system. 

\begin{figure}
\subfigure[]{\includegraphics[width=0.48\linewidth]{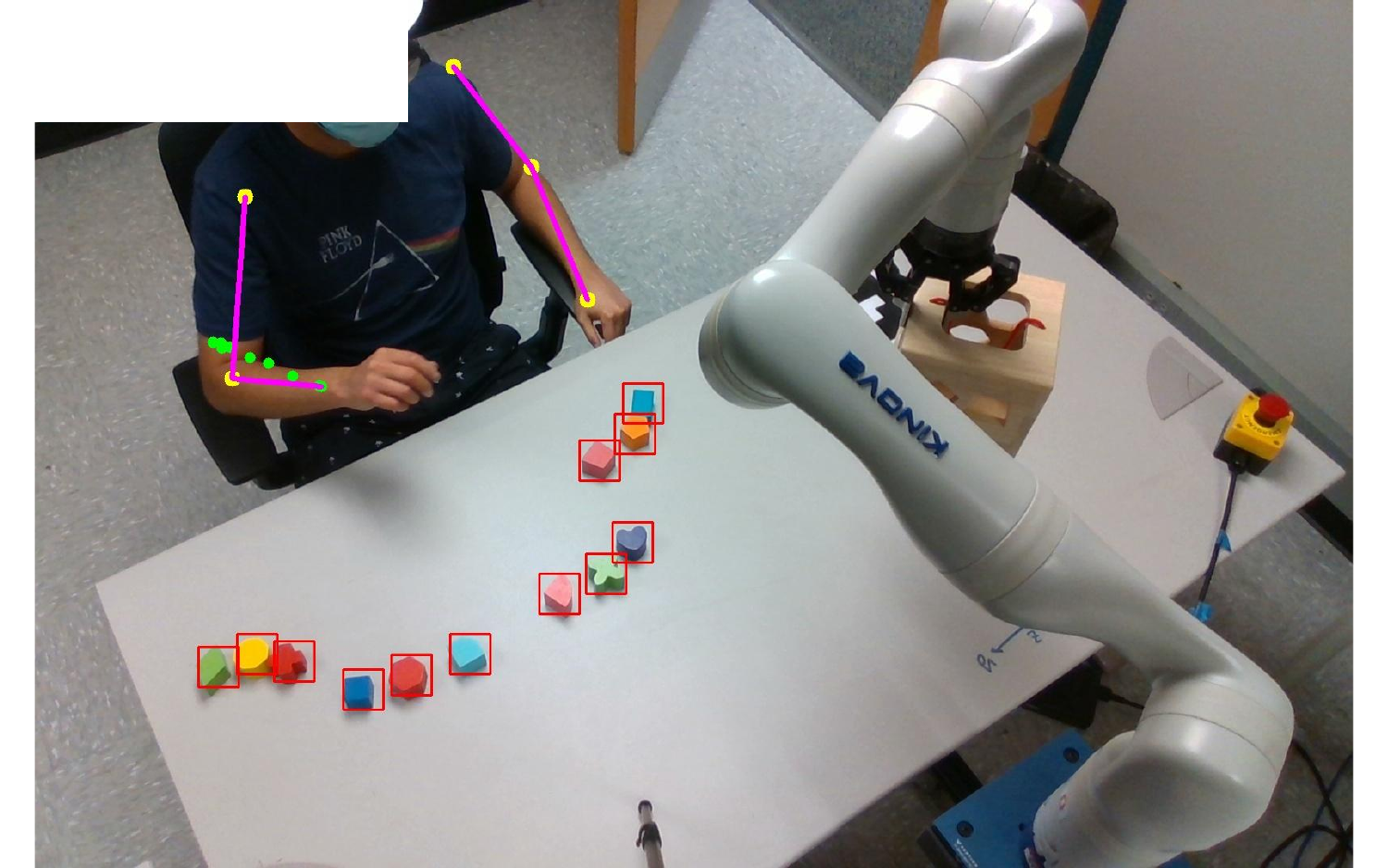}\label{fig:perception}}\hfill
\subfigure[]{\includegraphics[width=0.48\linewidth]{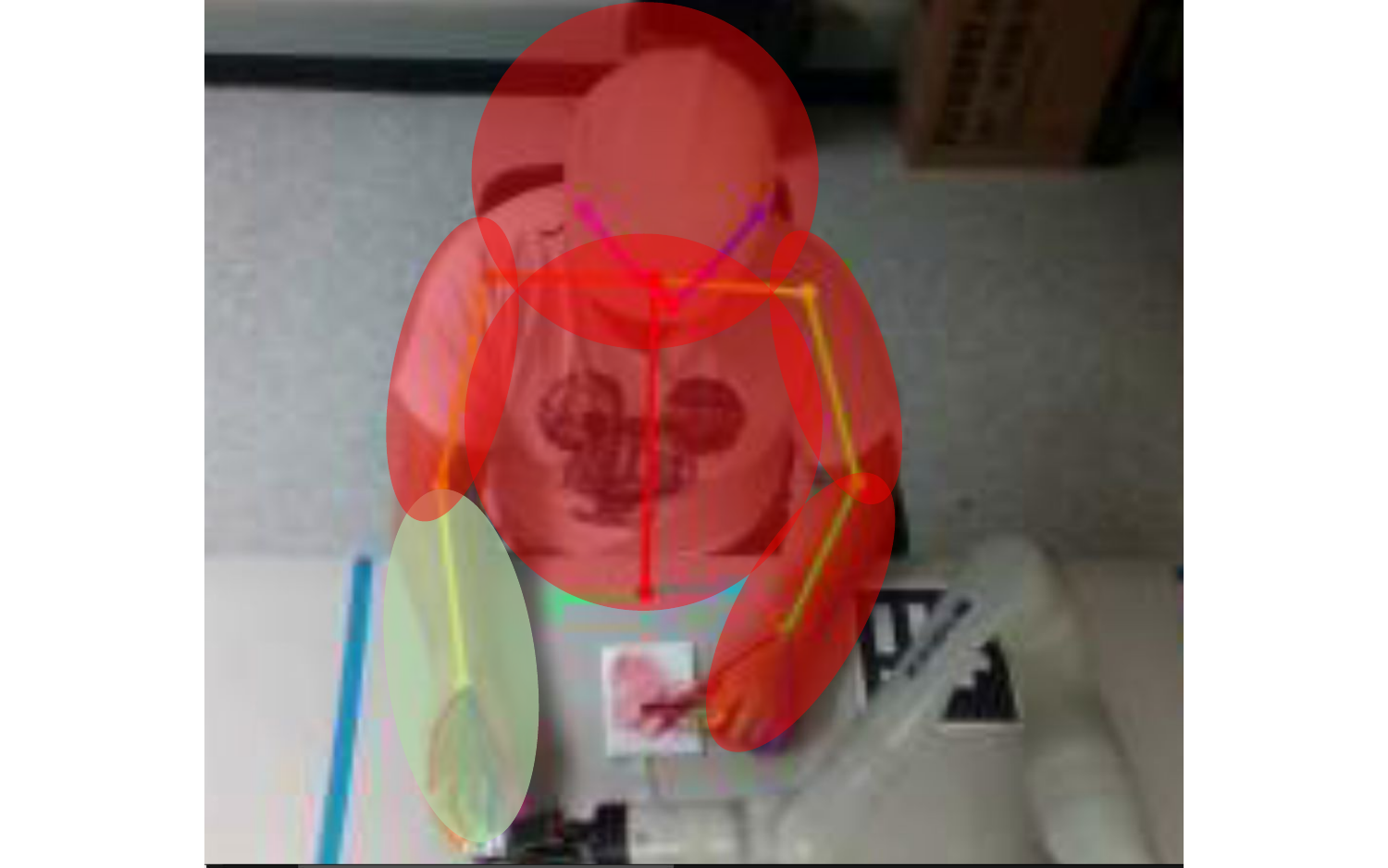}\label{fig:safety_specification}}
\vspace{-10pt}
    \caption{\footnotesize Left: Perception of the environment. Red bounding boxes: objects detected in the environment. Green dots: historical right wrist trajectory. Yellow dots: human body keypoints. Purple line segments: human skeleton links between keypoints. Right: Safety specification. Red: body parts that need to be avoided. Green: body parts that are allowed for close contact. \label{fig:perception_safety}}
    \vspace{-20pt}
\end{figure}

Given the nominal control $u$, we incorporate the safe set algorithm (SSA) \cite{ssa, jssa} at the downstream to adjust $u$ if necessary according to the real-time environment.
This paper considers discrete-time systems.
At a given time step $k$, we have the robot state $q^r_k$ and the environment $E_k$.
A scalar function $\phi_k=\Phi(q^r_k, E_k)$, which is derived from user safety specification as shown in \cref{fig:safety_specification} \cite{sos, jssa, ssa}, defines the safety status of the entire system.
$\Phi(\cdot)$ is defined such that the system is safe when $\phi \leq 0$ and unsafe when $\phi > 0$. 
The safety index value $\phi$ is evaluated online in the control loop, and the safe control $u^s$ is calculated as $\min_{u_k^s}||u_k^s-u_k||, \st \phi_{k+1} \leq 0$.
The resulting safe control $u_k^s$ is then sent to the robot for execution.

\section{EXPERIMENTS}
\begin{figure}
\subfigure[]{\includegraphics[width=0.49\linewidth]{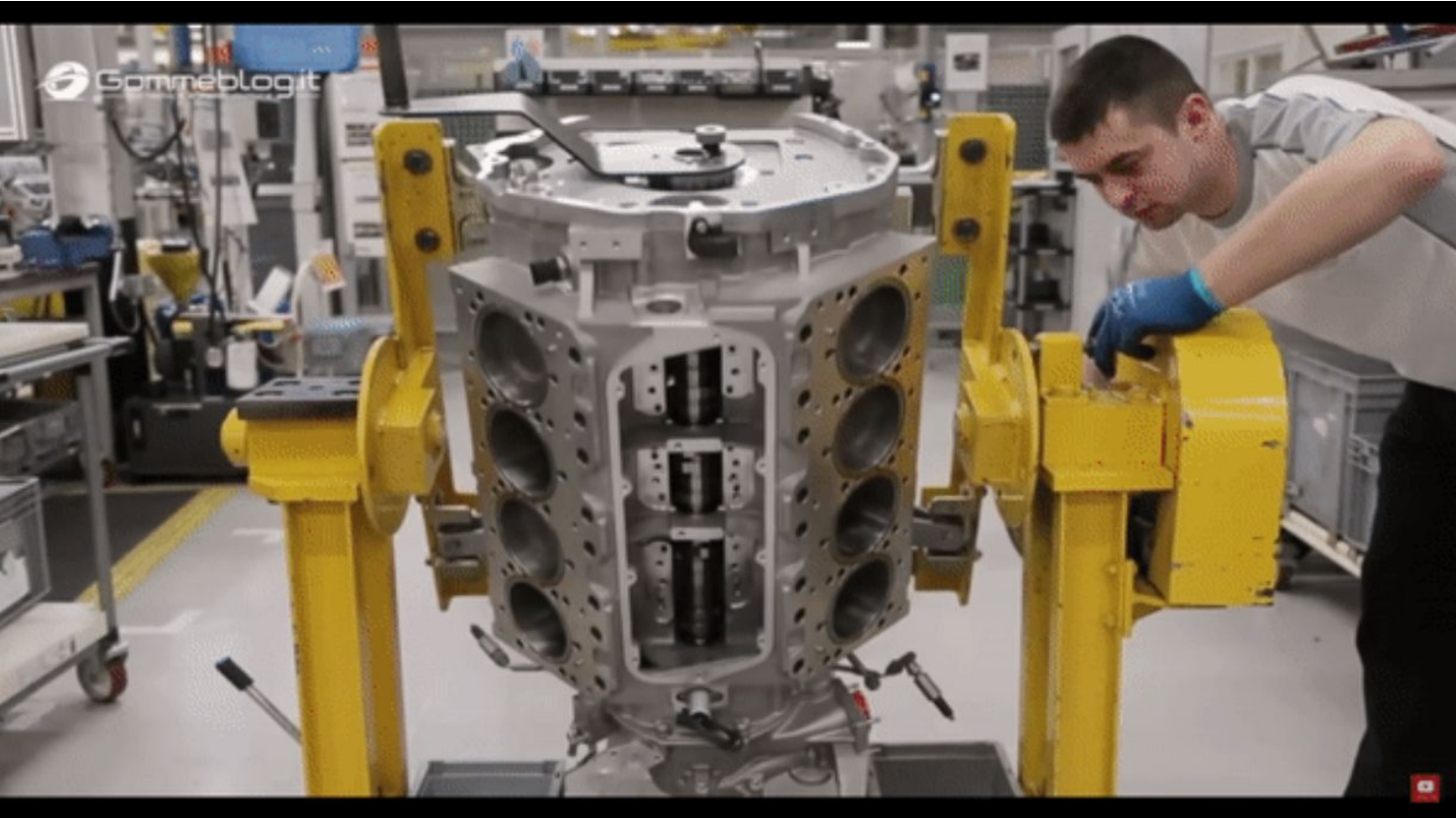}\label{fig:engine_example}}\hfill
\subfigure[]{\includegraphics[width=0.49\linewidth]{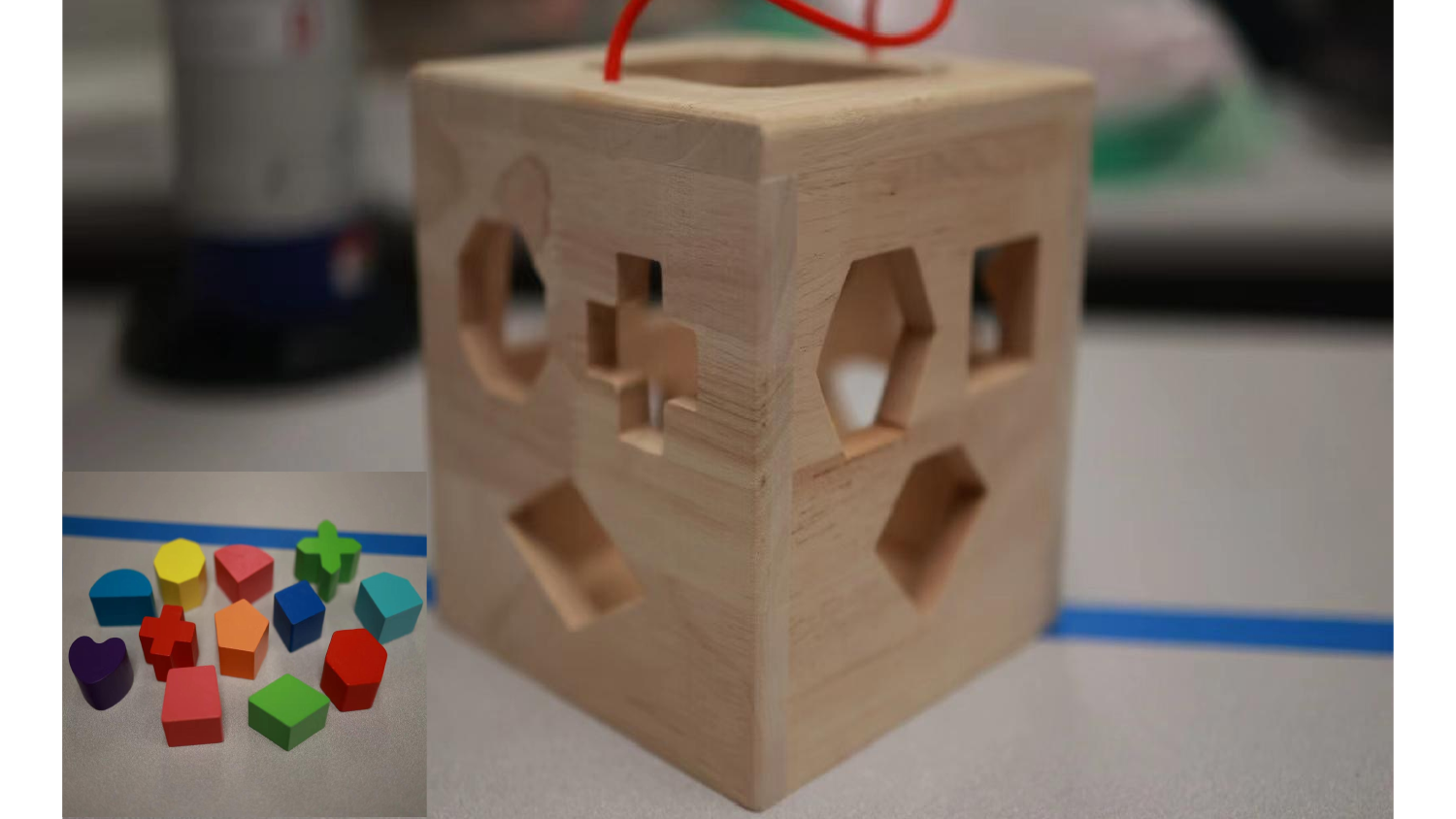}\label{fig:container}}
\vspace{-10pt}
    \caption{\footnotesize Illustration of assembly tasks. Left: A real-world engine assembly task\textsuperscript{1}. Right: A simplified co-assembly task. \label{fig:task}}
    \vspace{-20pt}
\end{figure}

Automotive engine assembly is a real-world manufacturing task, which currently heavily requires manual operations \cite{TROMMNAU2019387} as shown in \cref{fig:engine_example}.
In such tasks, the human worker needs to operate on different sides of the object (\ie the engine), which is not directly movable by humans due to its large size and heavy weight.
Therefore, the conventional approach is to have a fixture holding the object, and the human operator either manipulates the fixture in order to rotate the object, or maneuvers around to reach different sides.
HRC is a feasible solution to improve task efficiency by having a robot hold the object. 
Instead of having the human worker manually control the robot to operate the object, the proposed framework enables the robot to intelligently operate the object based on its inference of human behavior. 

\footnotetext[1]{Picture from https://youtu.be/aHAMNpYxjPc}

Inspired by the real-world manufacturing task (\ie engine assembly in \cref{fig:engine_example}), this paper considers a simplified assembly task as shown in \cref{fig:container}. The human is given blocks with different shapes as shown in the bottom-left of \cref{fig:container}. The goal of the assembly is to insert the blocks into the container using holes on different surfaces. 
It is assumed that the container is heavy, and thus, not directly operable by humans.
With HRC, we have a robot manipulating the container and a human worker inserting the blocks. 
The robot and the human collaborate in a way that the robot displays the corresponding surface to the human for easy insertion.

The task shown in \cref{fig:container} shares similarities with the manufacturing task in \cref{fig:engine_example}.
First, the assembly requires sequential operations.
Second, it requires operations on different sides of a non-operable object.
More importantly, it has instructions that the human must follow, but at the same time, has flexibility for the human to decide based on preferences.
For example, human workers need to finish one surface at a time, but they can choose which surface to operate first, and the order of block insertion within a surface. 

We compare our solution with a baseline approach, in which the robot passively waits for human command and the human operator manually controls the robot to move when needed. This corresponds to the current manufacturing approach. 
On the other hand, our solution has the robot proactively collaborating with the human worker by inferring the task progress and human intention without explicit human command.
We use a 3-layer FCNN (2 Relu hidden layers with 32 neurons and 1 linear output layer) to model $h_{\theta}(\cdot)$ with IADA training \cite{iada}. 
The training data includes demonstrations from two human subjects, and each human demonstrated for two trials.
Each human demonstrated the task based on his insertion preference and the clusters of blocks are placed randomly.
The experiment results include five human subjects (including the two demonstrators), and each human did the task for two trials.

\subsection{Robust Collaboration}
It is important that the robot can robustly collaborate with the human worker.
This paper mainly considers robustness in three aspects, 1) human preferences, 2) environment setups, and 3) different humans.

\begin{figure*}
\subfigure[]{\includegraphics[width=0.12\linewidth]{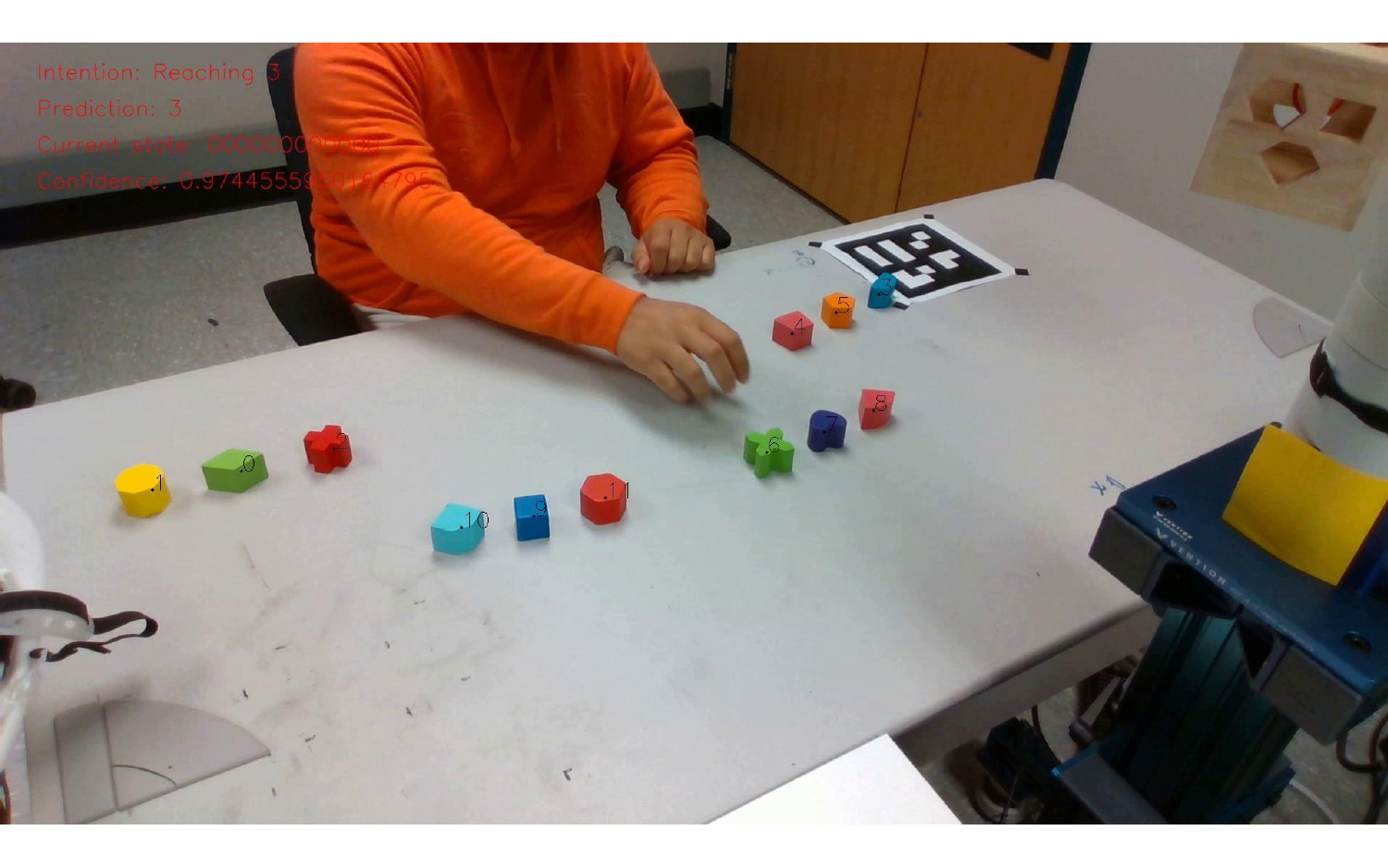}\label{fig:1_1}}\hfill
\subfigure[]{\includegraphics[width=0.12\linewidth]{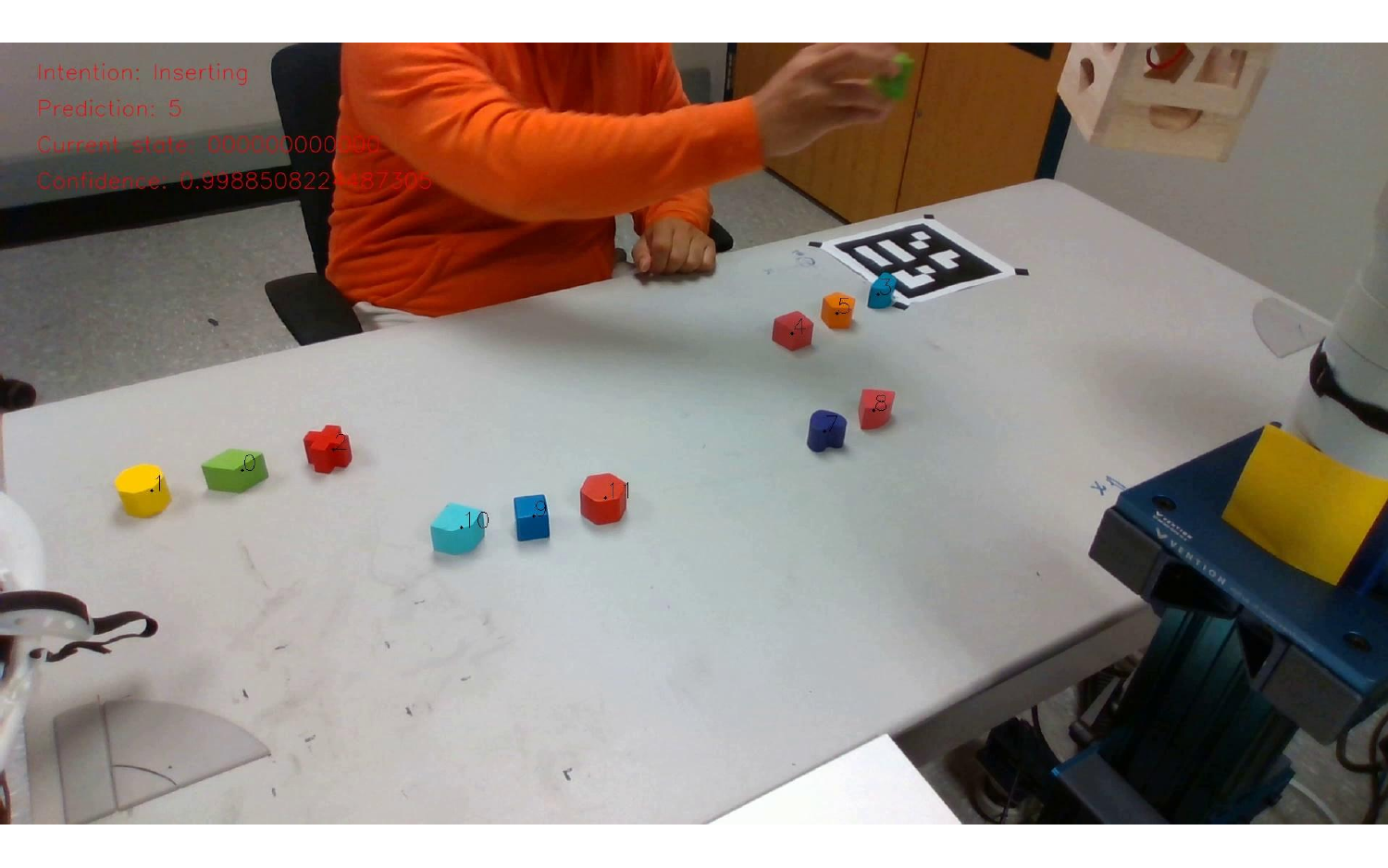}\label{fig:1_2}}\hfill
\subfigure[]{\includegraphics[width=0.12\linewidth]{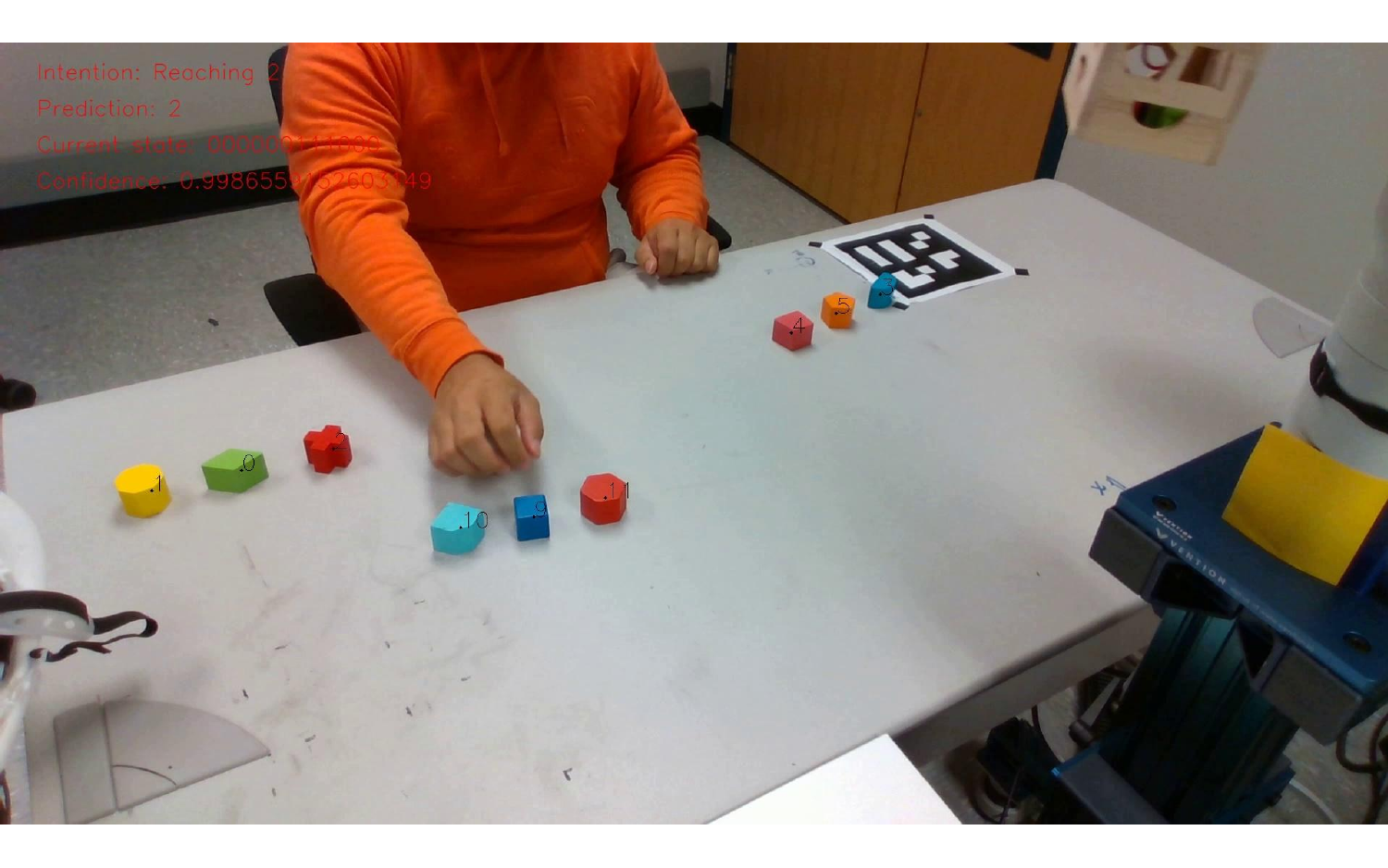}\label{fig:1_3}}\hfill
\subfigure[]{\includegraphics[width=0.12\linewidth]{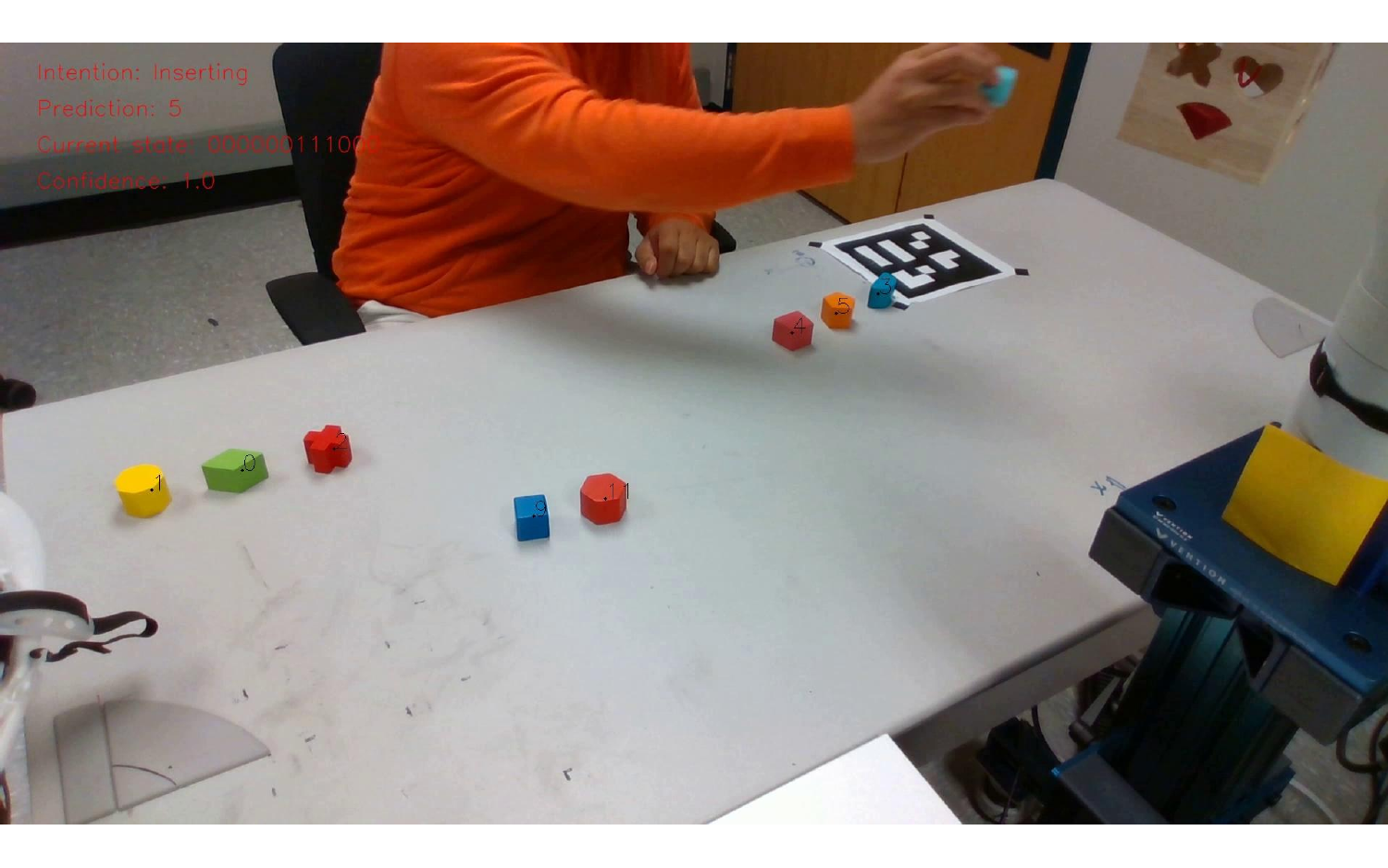}\label{fig:1_4}}\hfill
\subfigure[]{\includegraphics[width=0.12\linewidth]{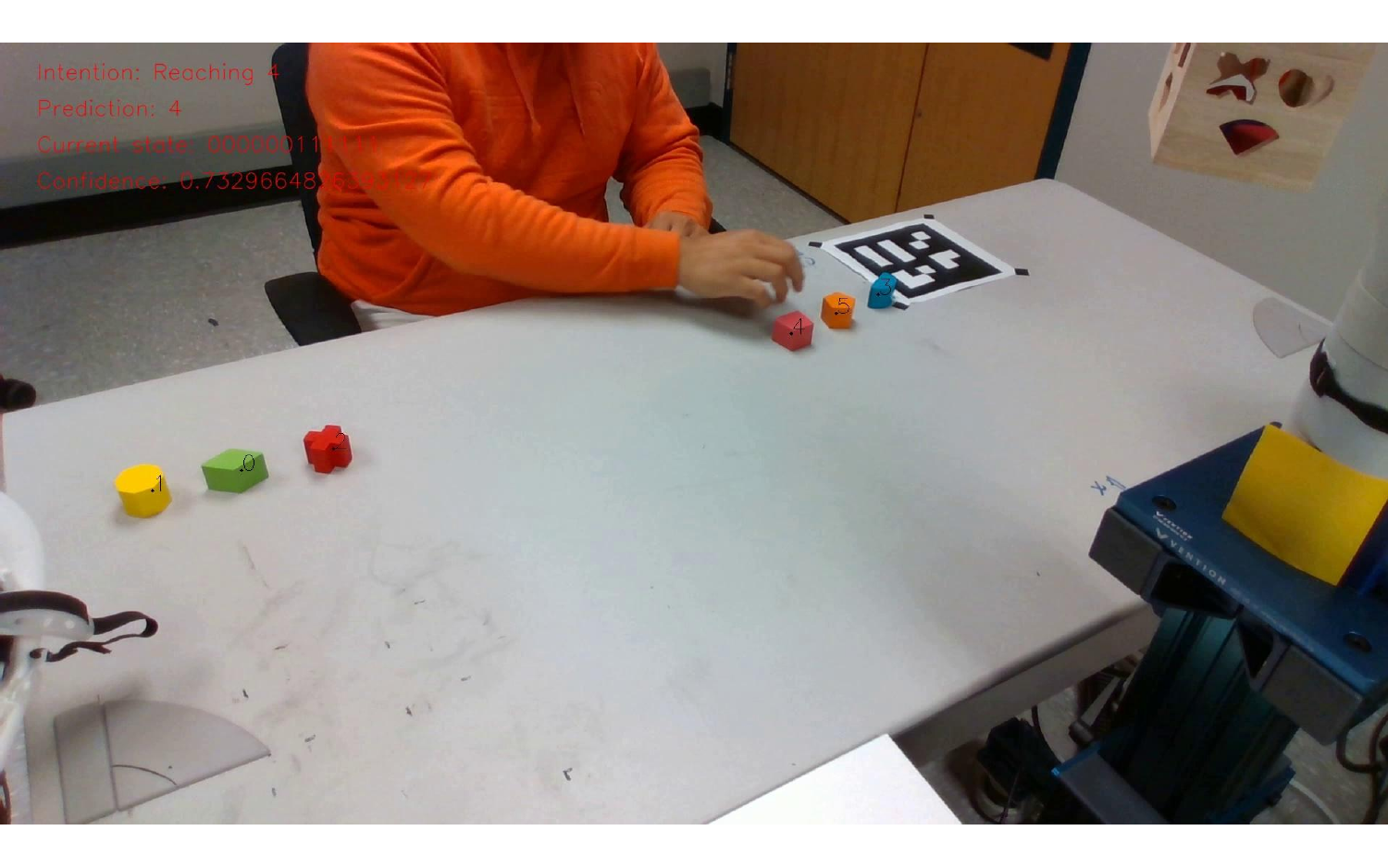}\label{fig:1_5}}\hfill
\subfigure[]{\includegraphics[width=0.12\linewidth]{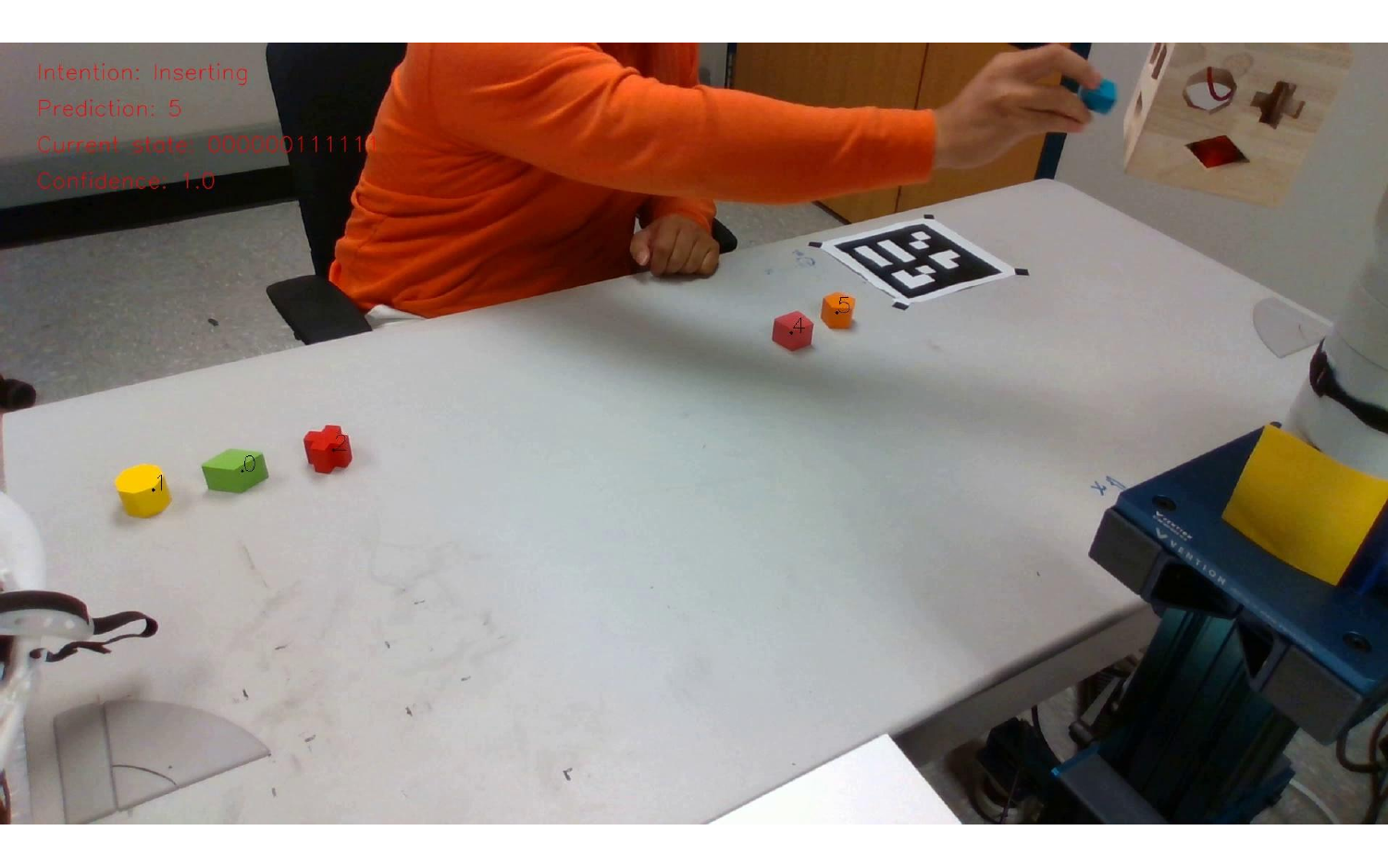}\label{fig:1_6}}\hfill
\subfigure[]{\includegraphics[width=0.12\linewidth]{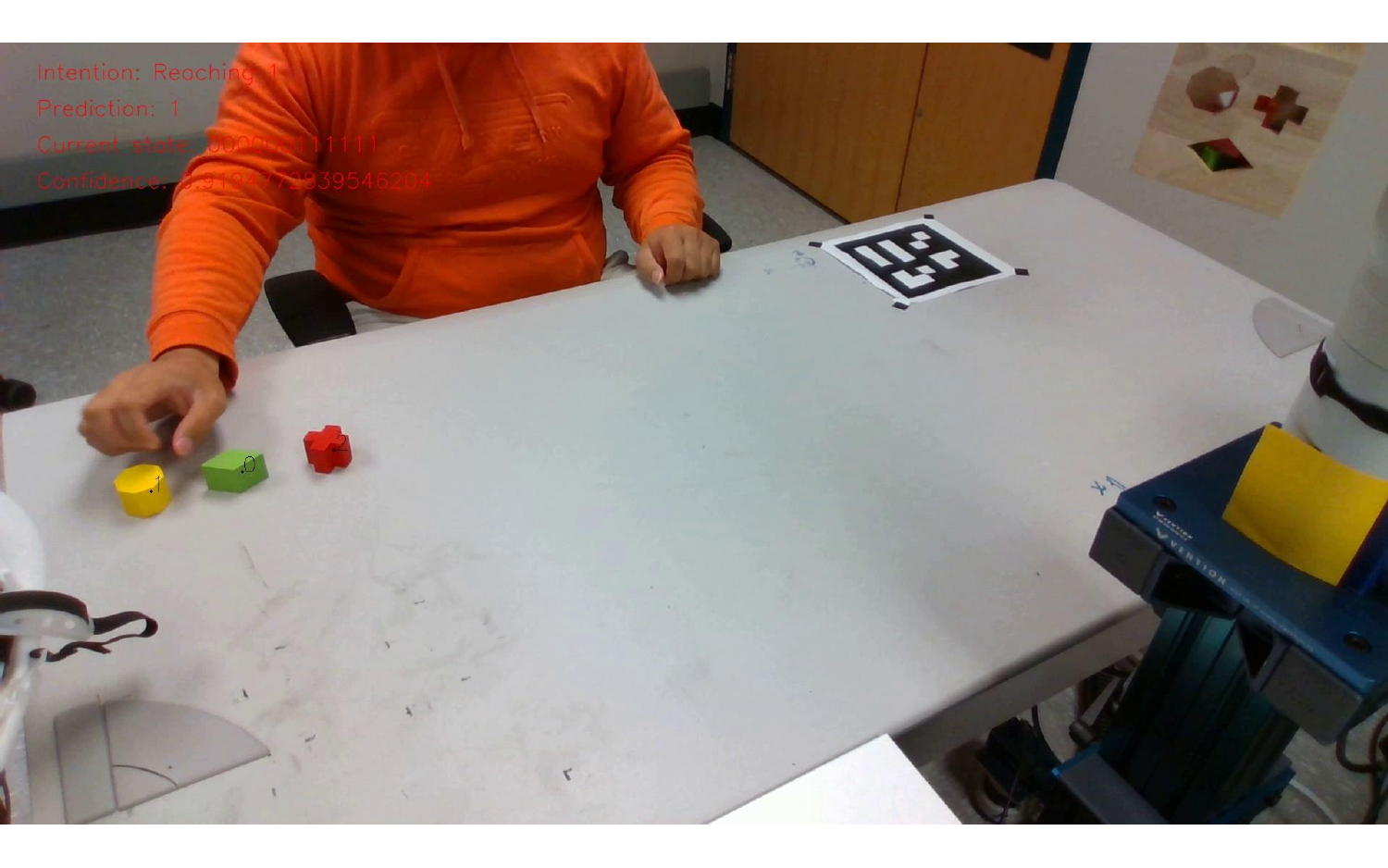}\label{fig:1_7}}\hfill
\subfigure[]{\includegraphics[width=0.12\linewidth]{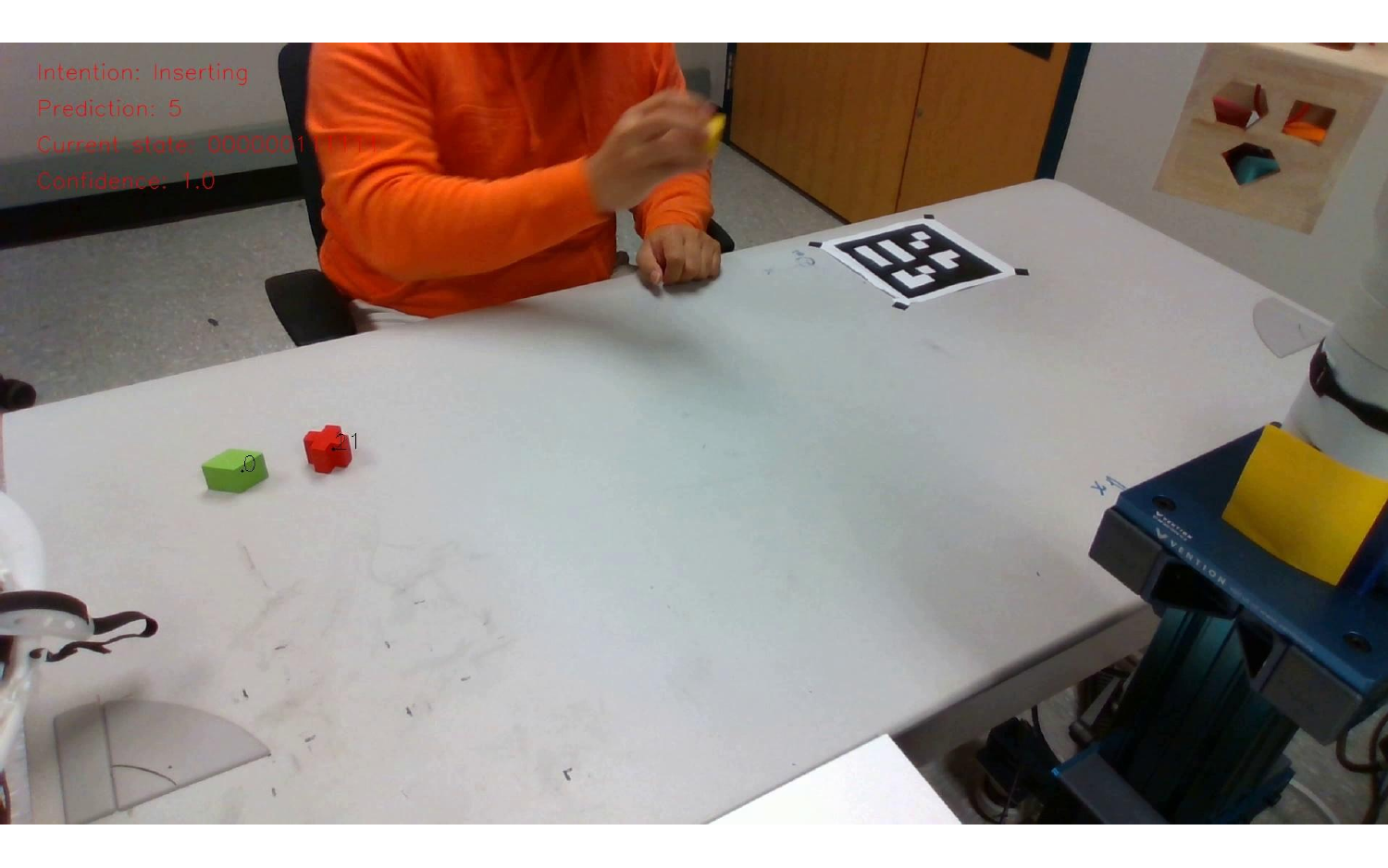}\label{fig:1_8}}
\vspace{-10pt}\\
\subfigure[]{\includegraphics[width=0.12\linewidth]{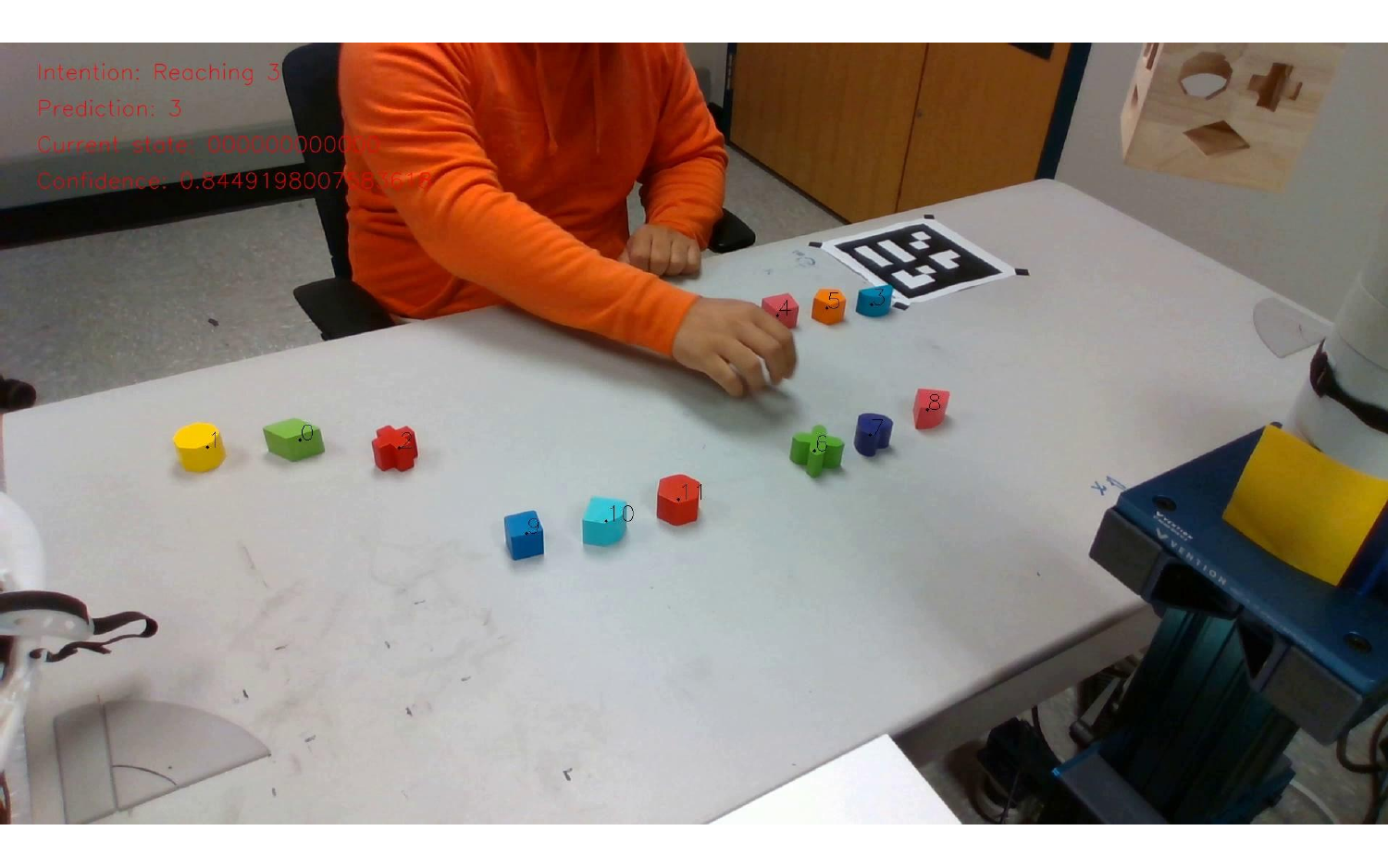}\label{fig:2_1}}\hfill
\subfigure[]{\includegraphics[width=0.12\linewidth]{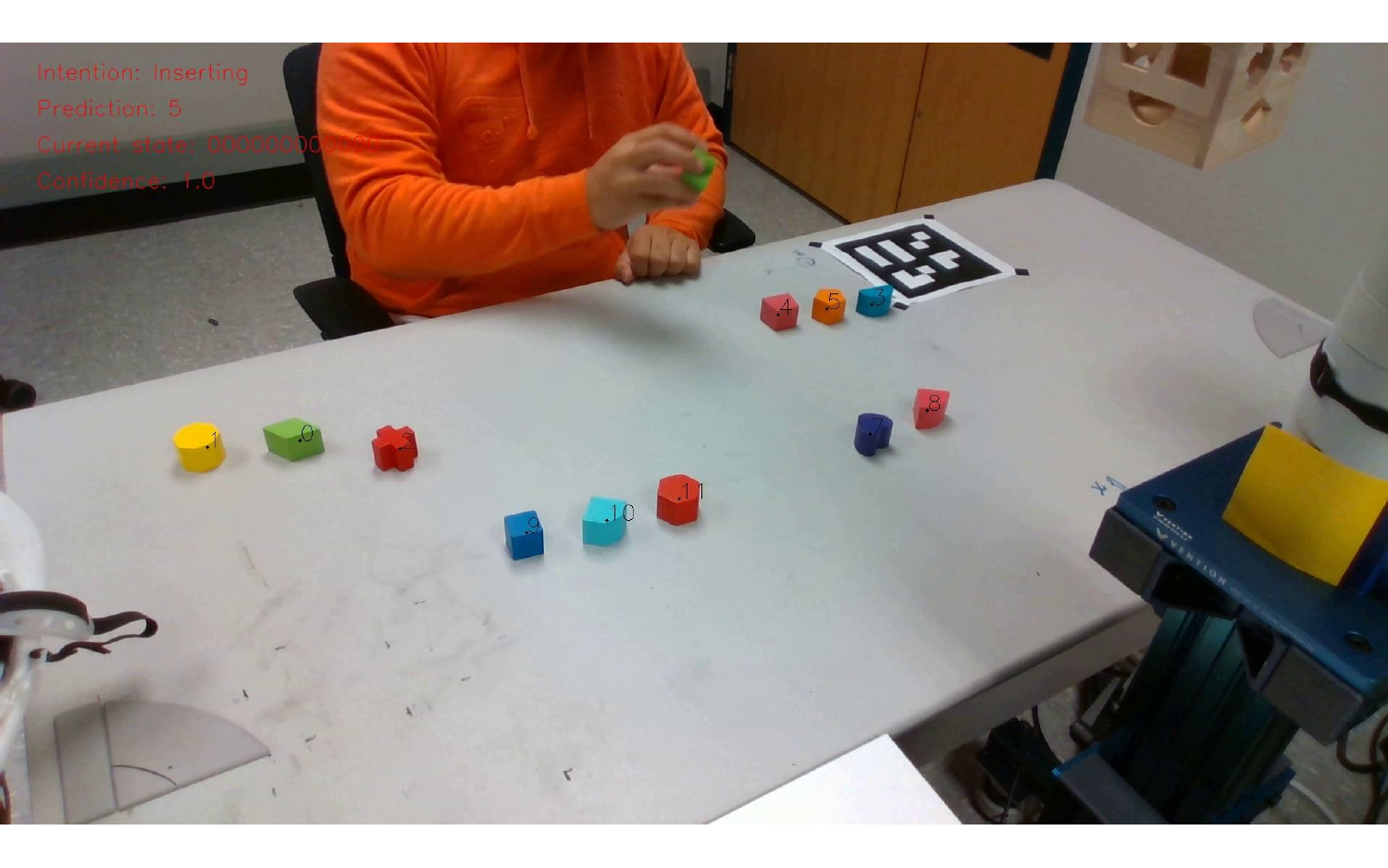}\label{fig:2_2}}\hfill
\subfigure[]{\includegraphics[width=0.12\linewidth]{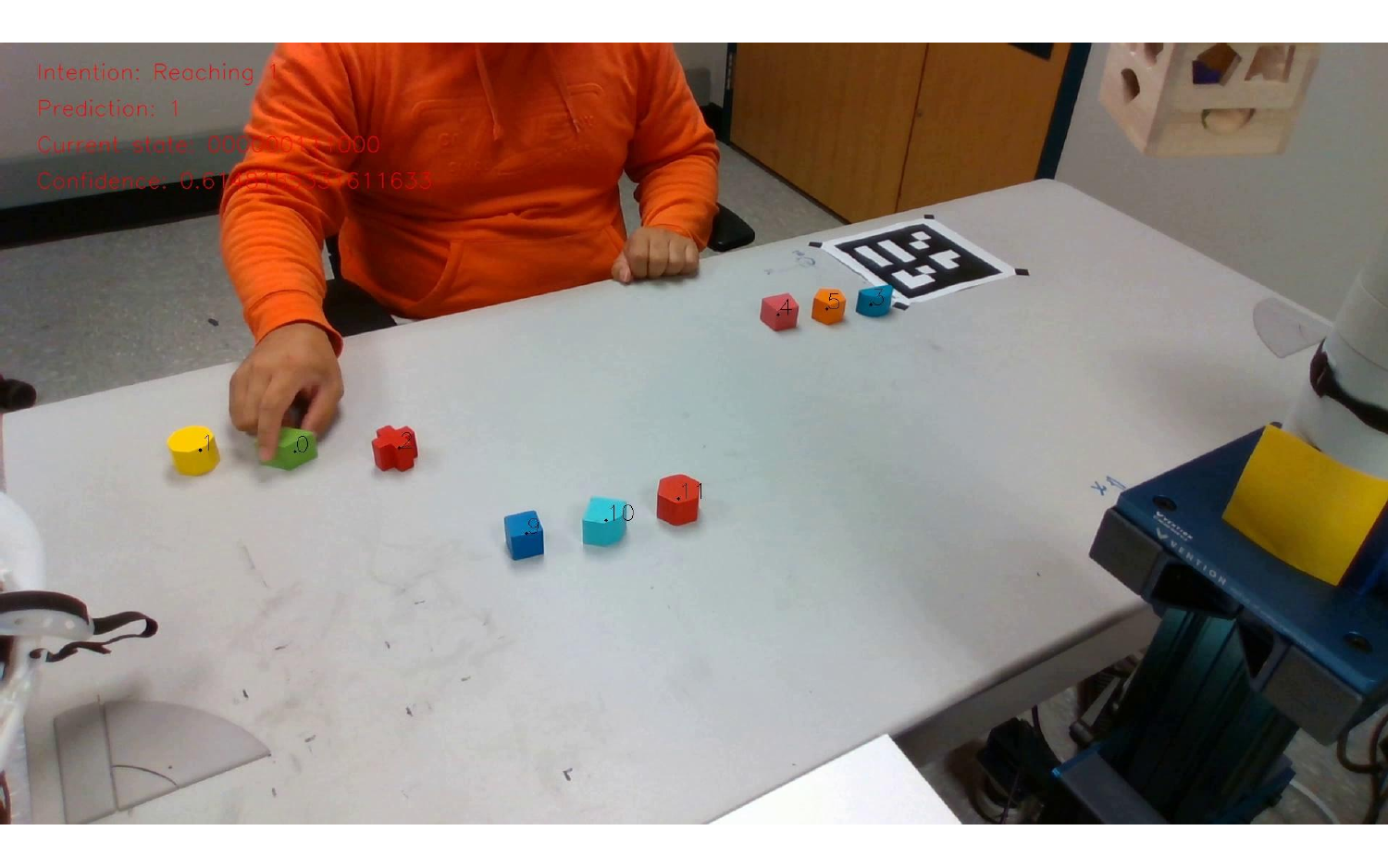}\label{fig:2_3}}\hfill
\subfigure[]{\includegraphics[width=0.12\linewidth]{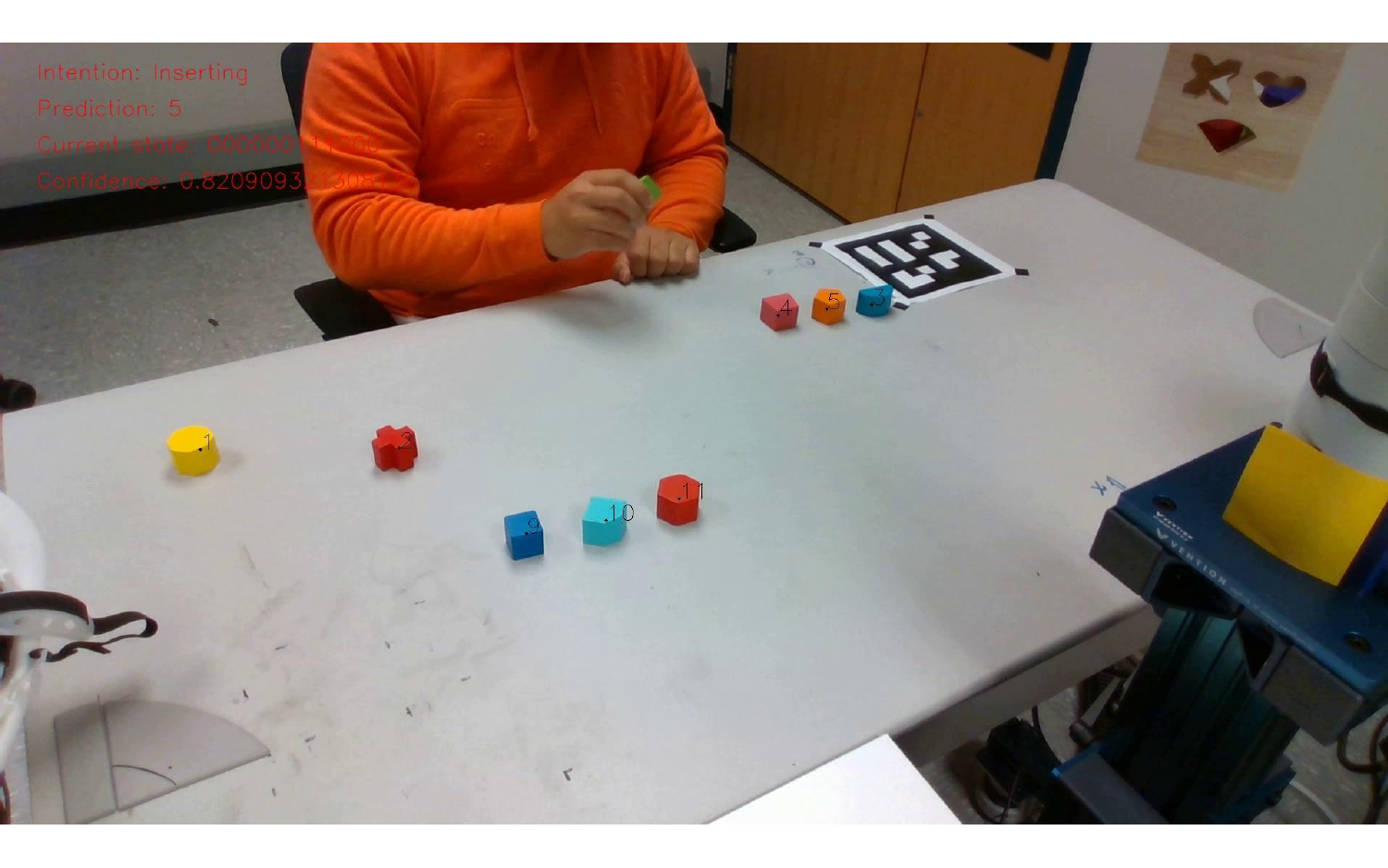}\label{fig:2_4}}\hfill
\subfigure[]{\includegraphics[width=0.12\linewidth]{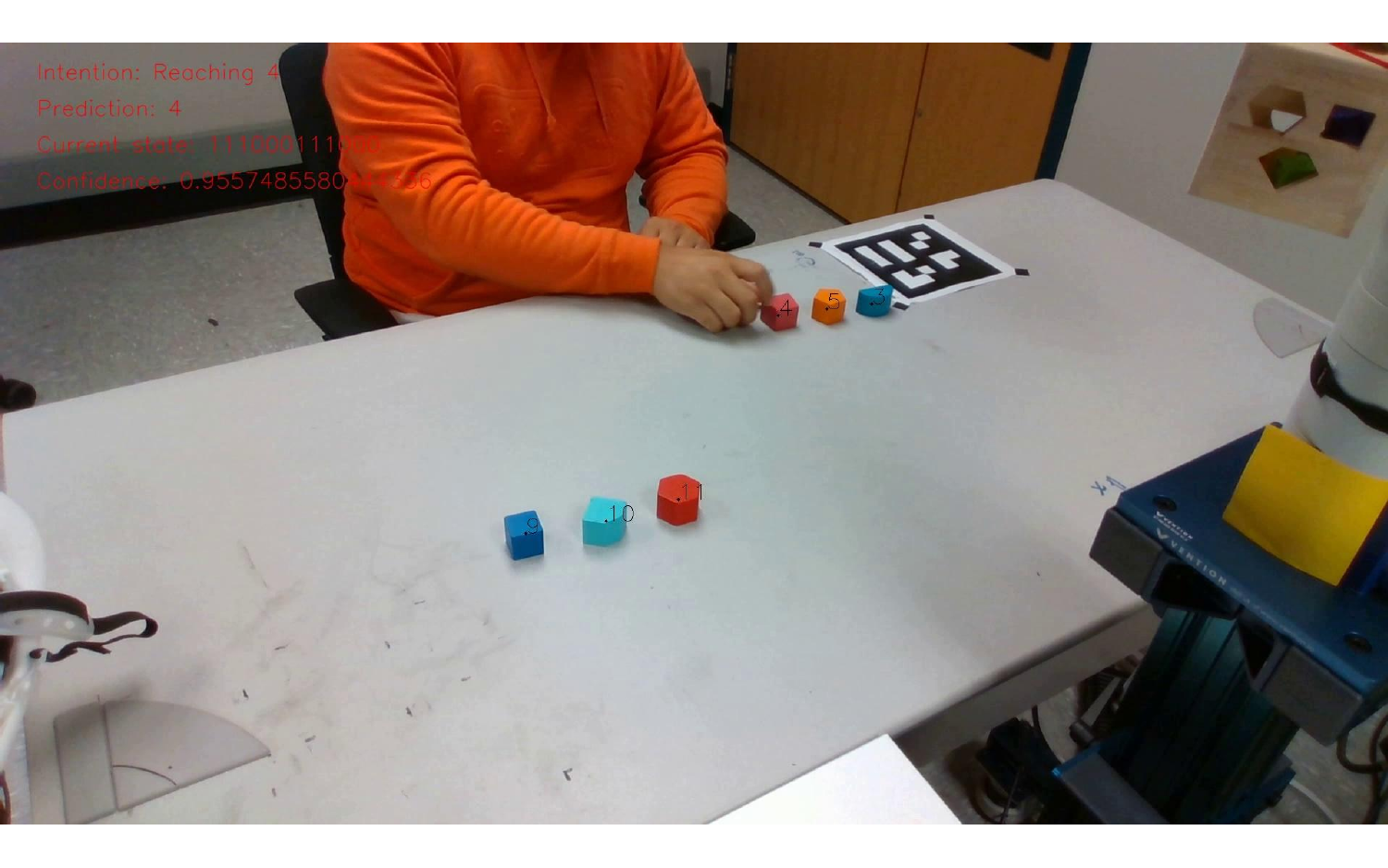}\label{fig:2_5}}\hfill
\subfigure[]{\includegraphics[width=0.12\linewidth]{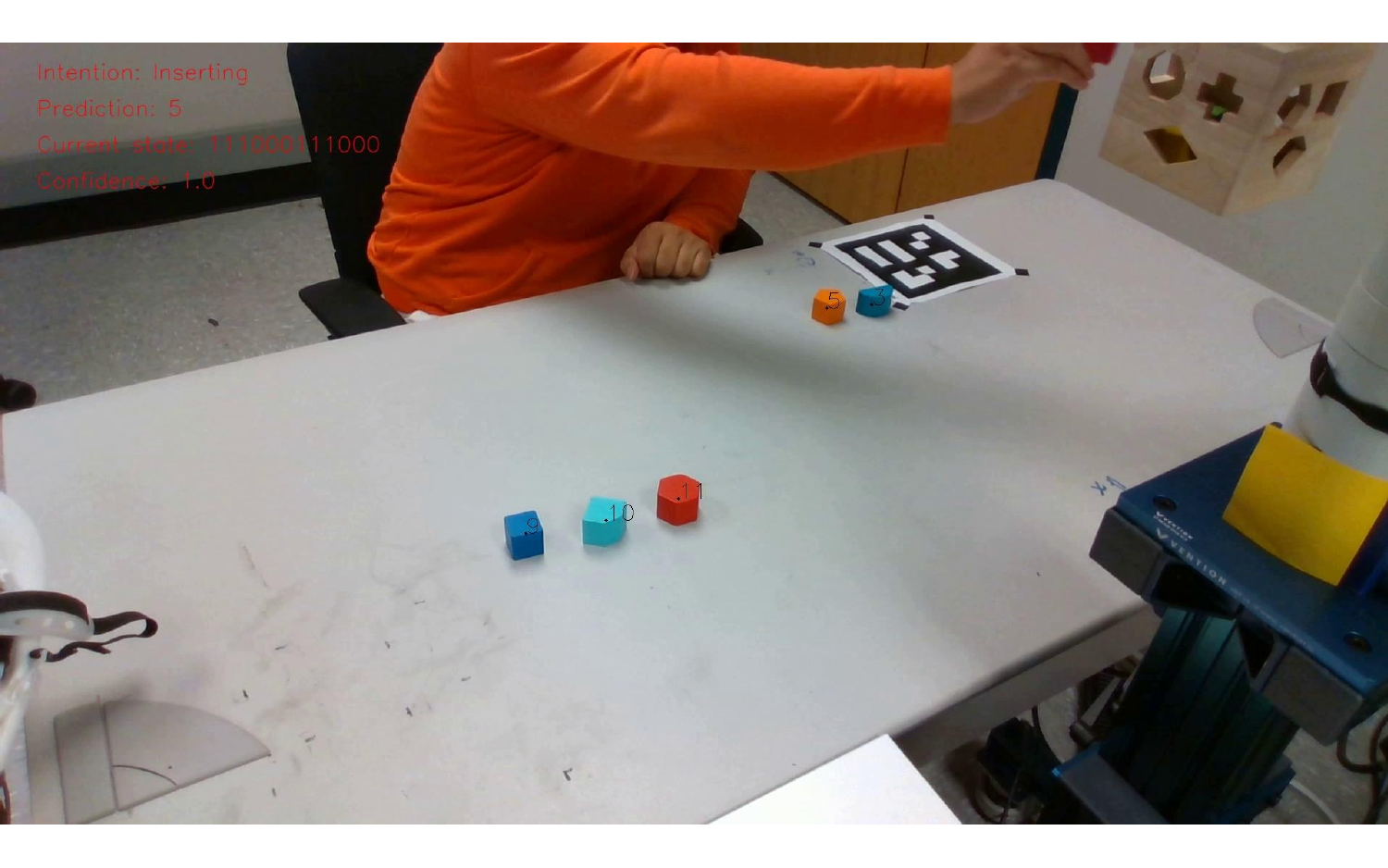}\label{fig:2_6}}\hfill
\subfigure[]{\includegraphics[width=0.12\linewidth]{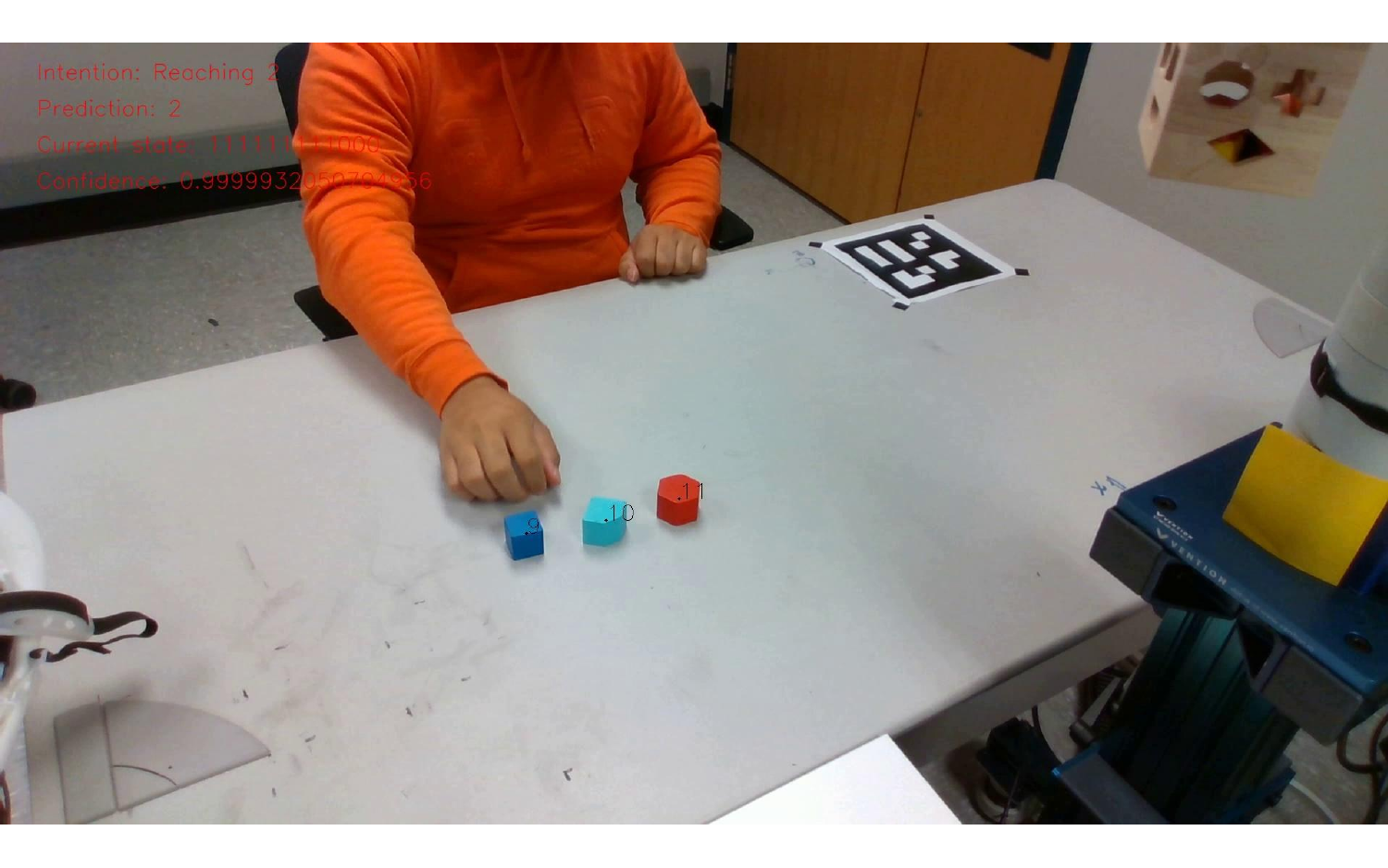}\label{fig:2_7}}\hfill
\subfigure[]{\includegraphics[width=0.12\linewidth]{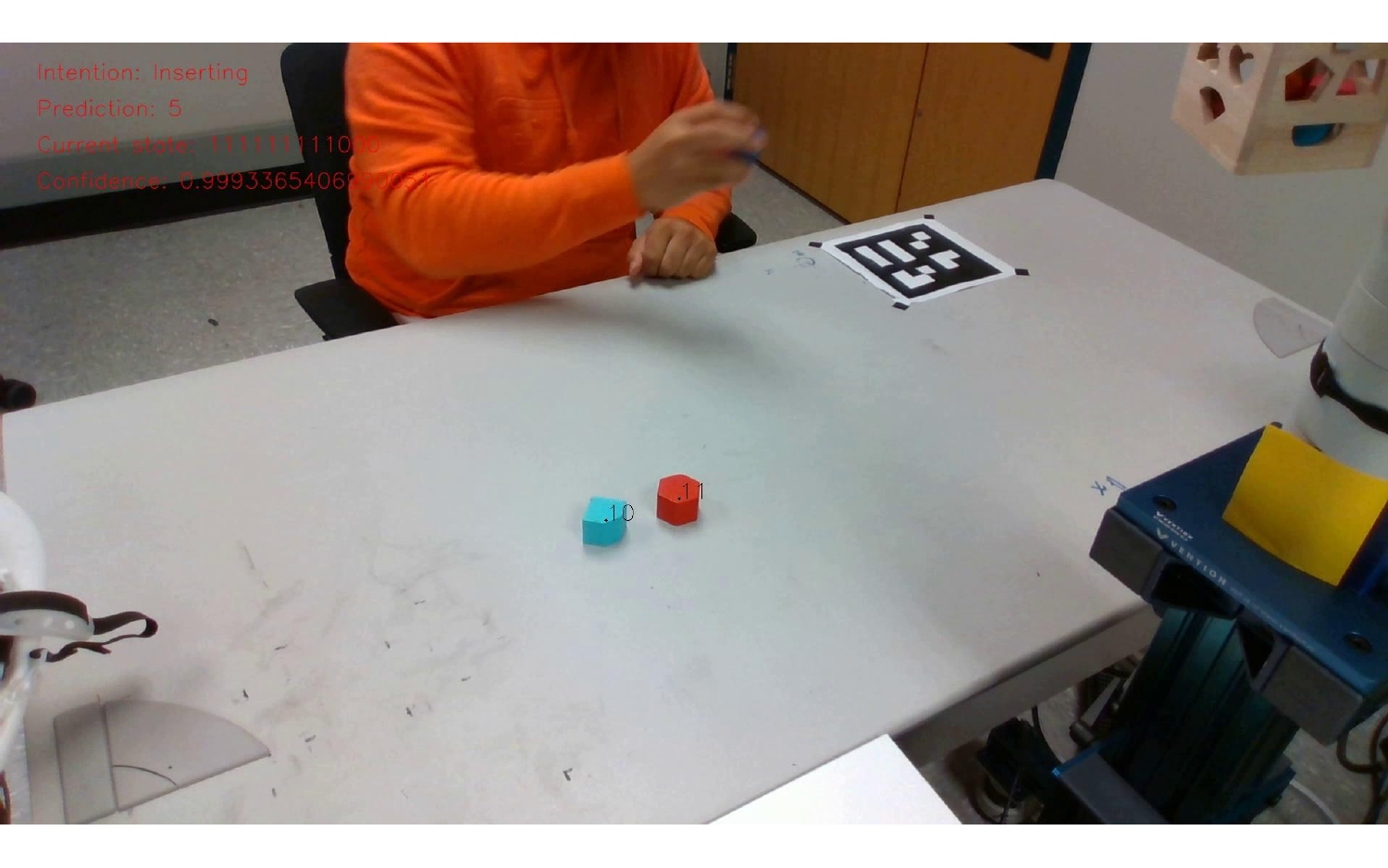}\label{fig:2_8}}
\vspace{-10pt}\\
\subfigure[]{\includegraphics[width=0.12\linewidth]{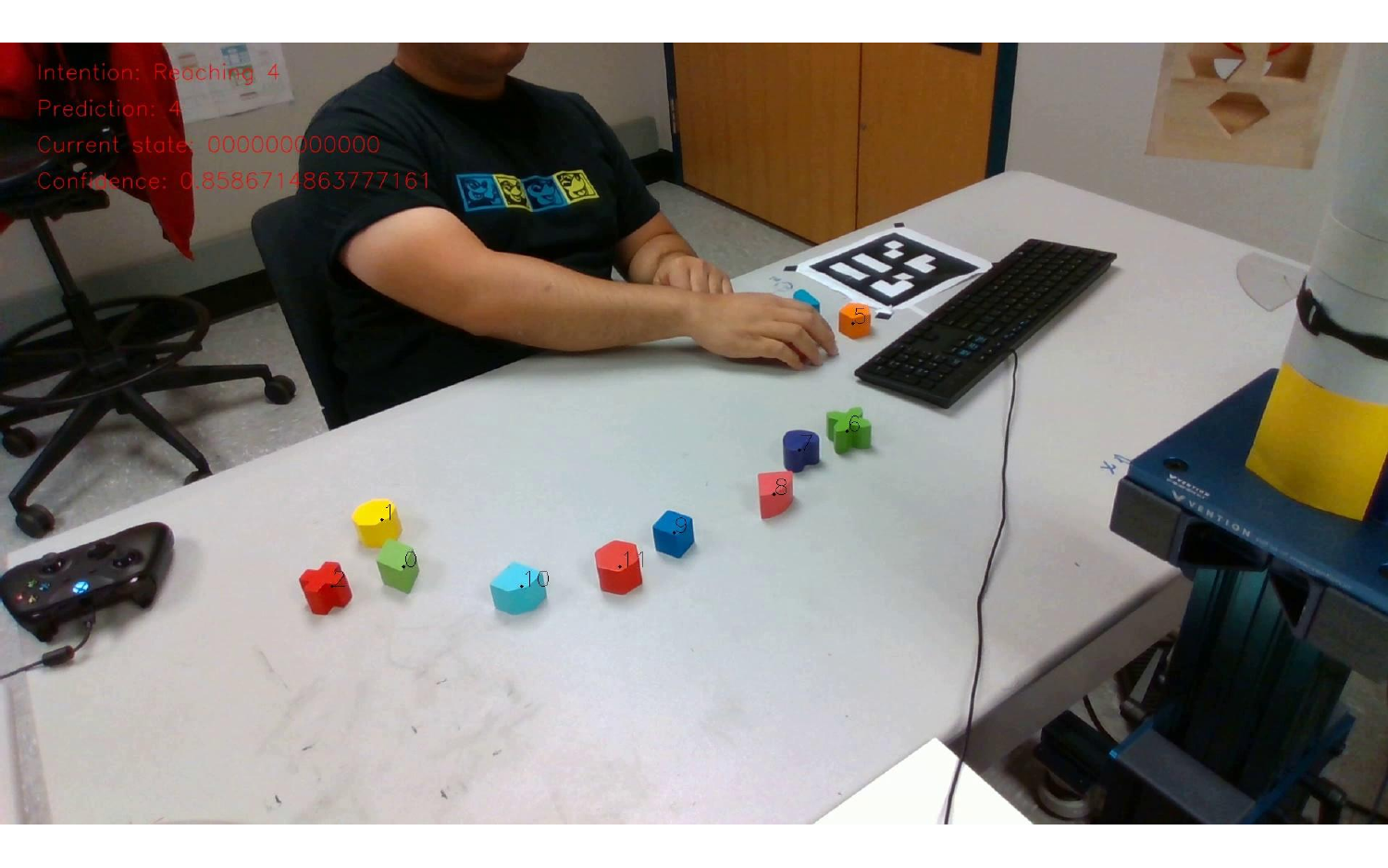}\label{fig:3_1}}\hfill
\subfigure[]{\includegraphics[width=0.12\linewidth]{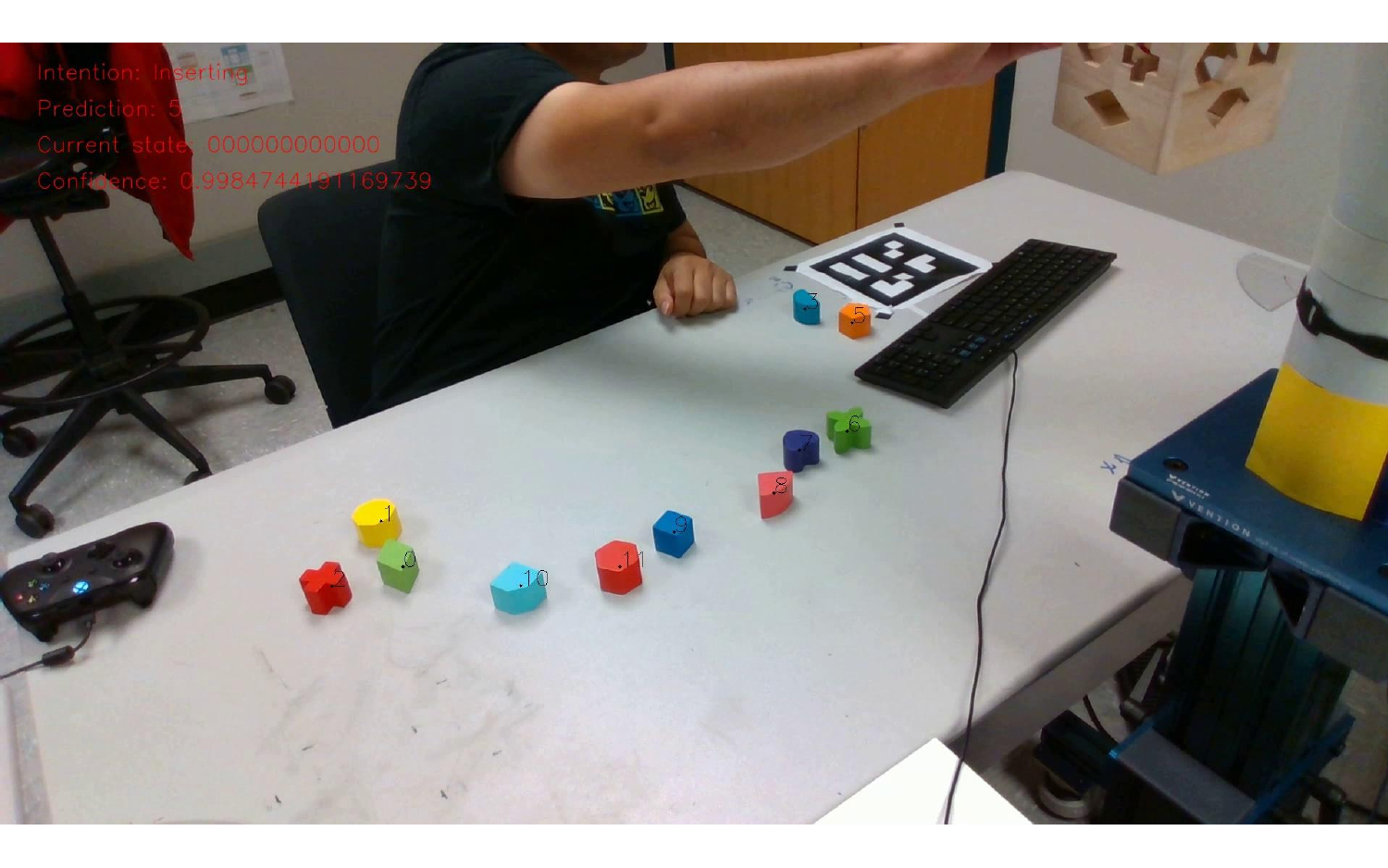}\label{fig:3_2}}\hfill
\subfigure[]{\includegraphics[width=0.12\linewidth]{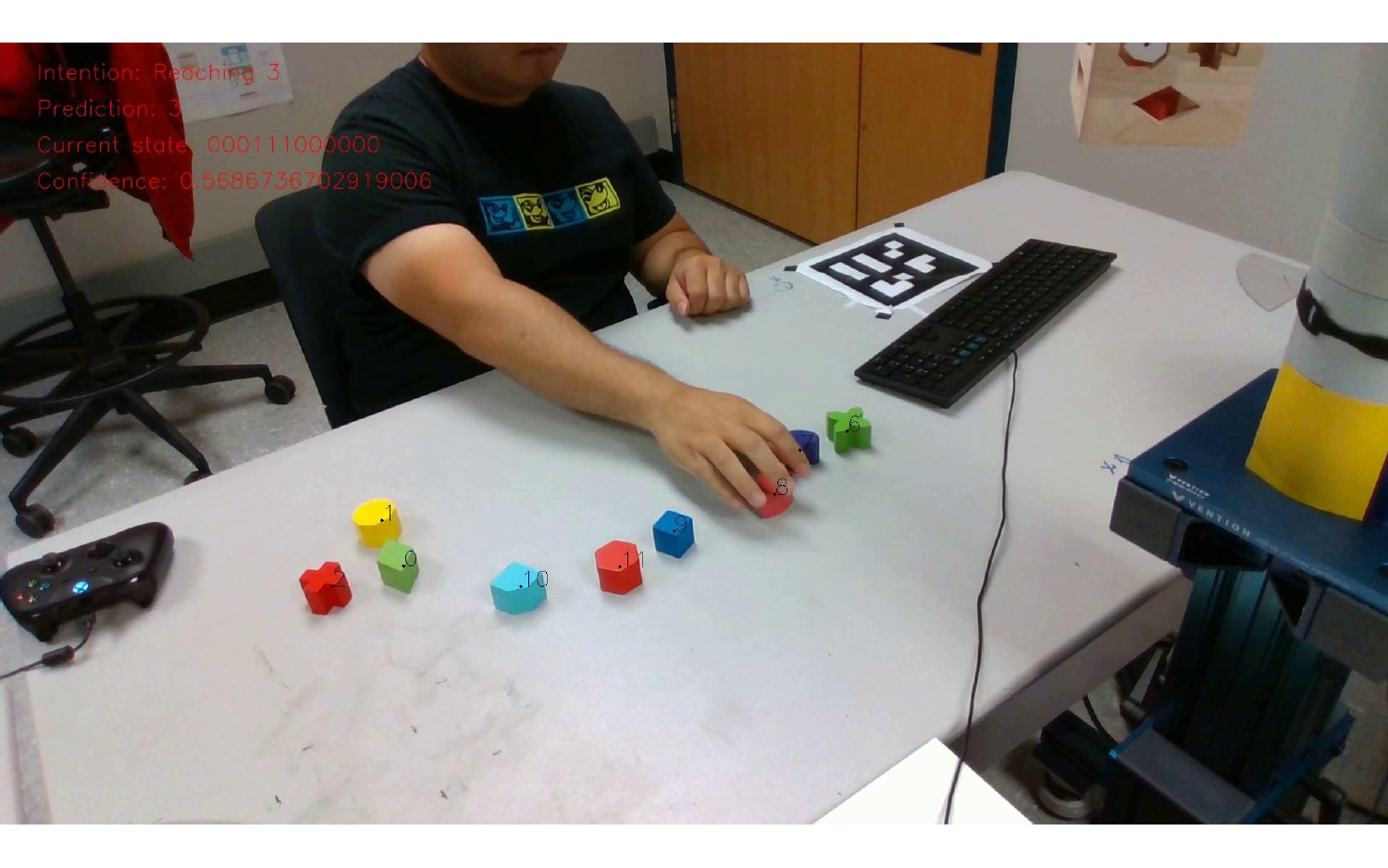}\label{fig:3_3}}\hfill
\subfigure[]{\includegraphics[width=0.12\linewidth]{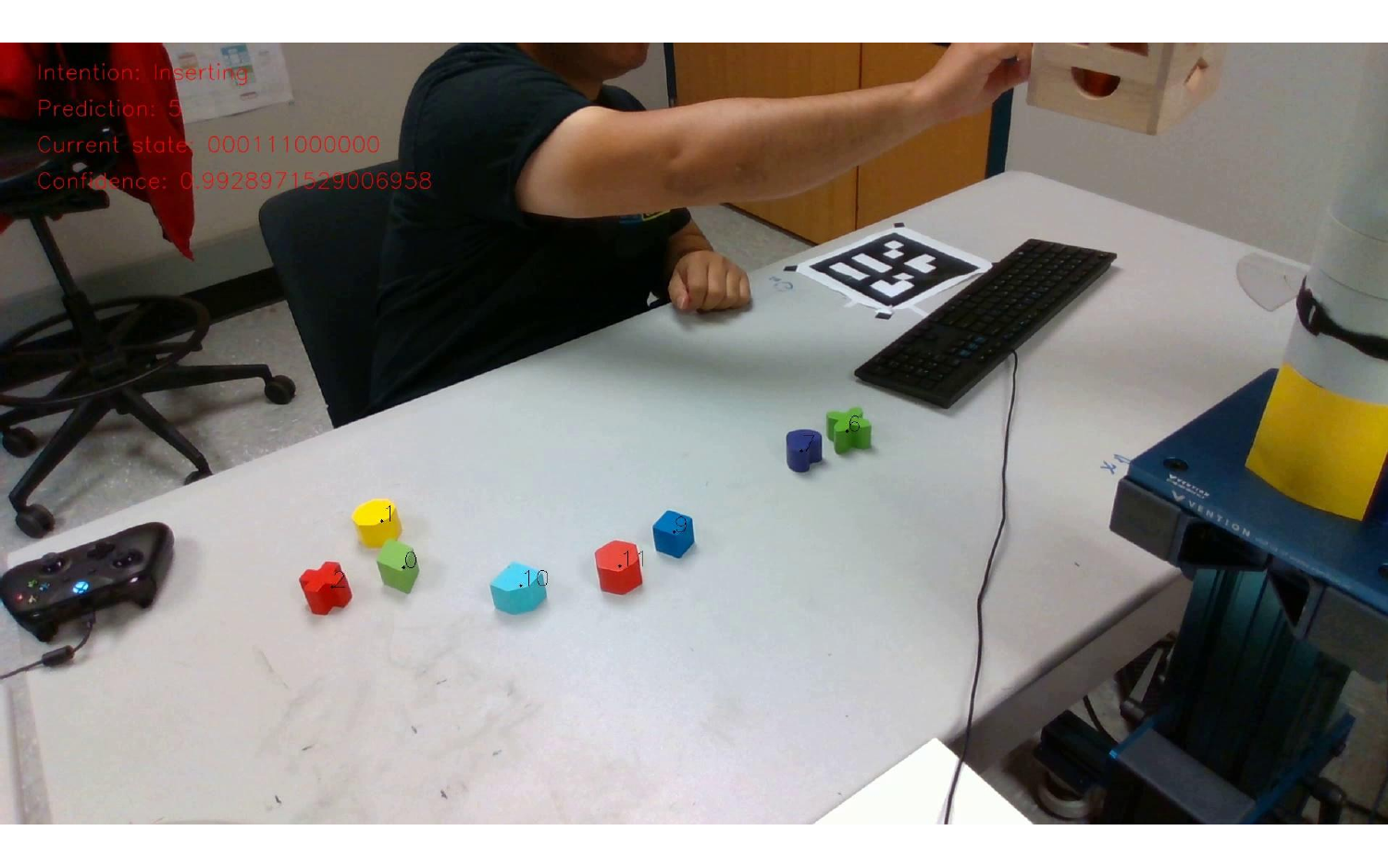}\label{fig:3_4}}\hfill
\subfigure[]{\includegraphics[width=0.12\linewidth]{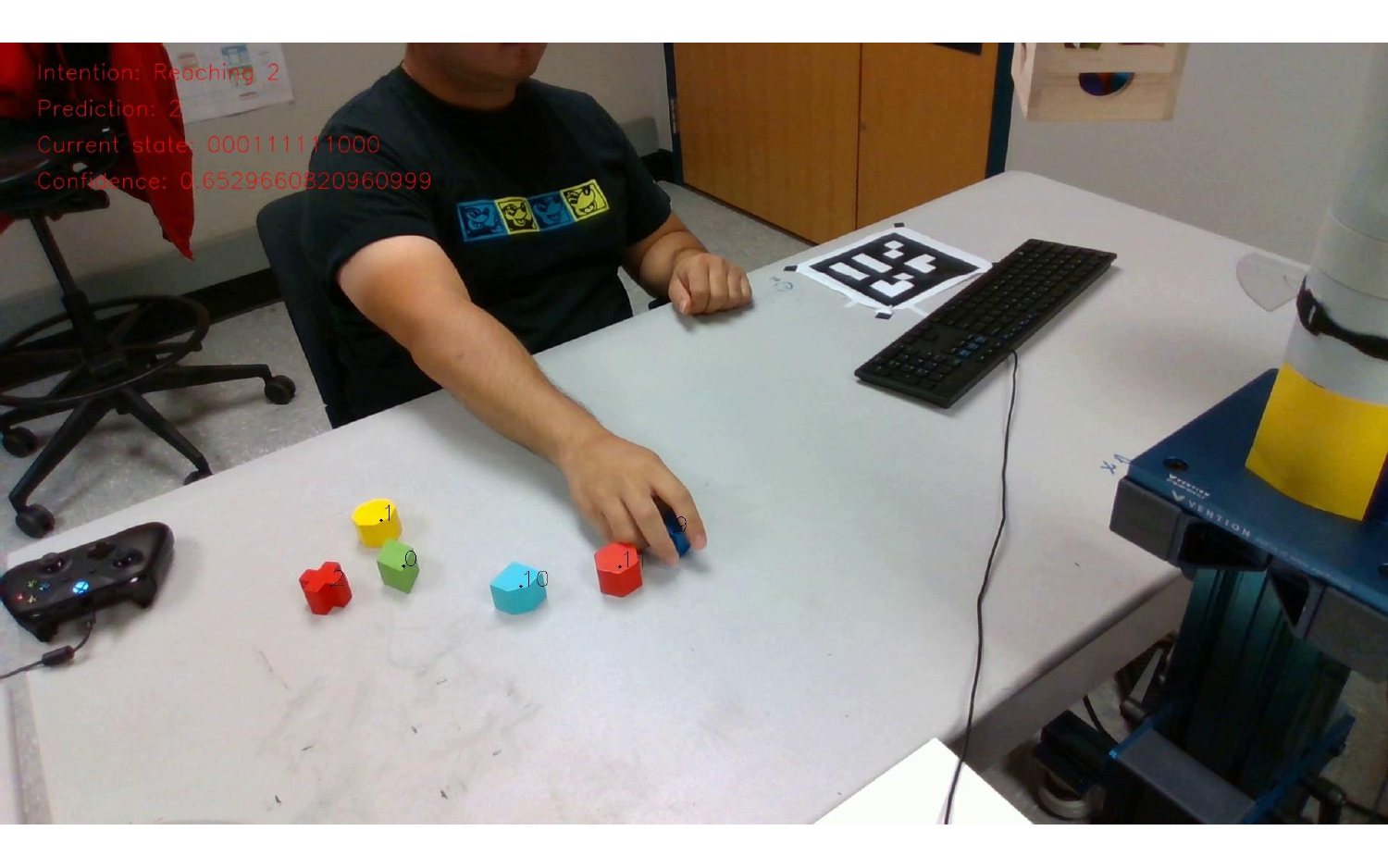}\label{fig:3_5}}\hfill
\subfigure[]{\includegraphics[width=0.12\linewidth]{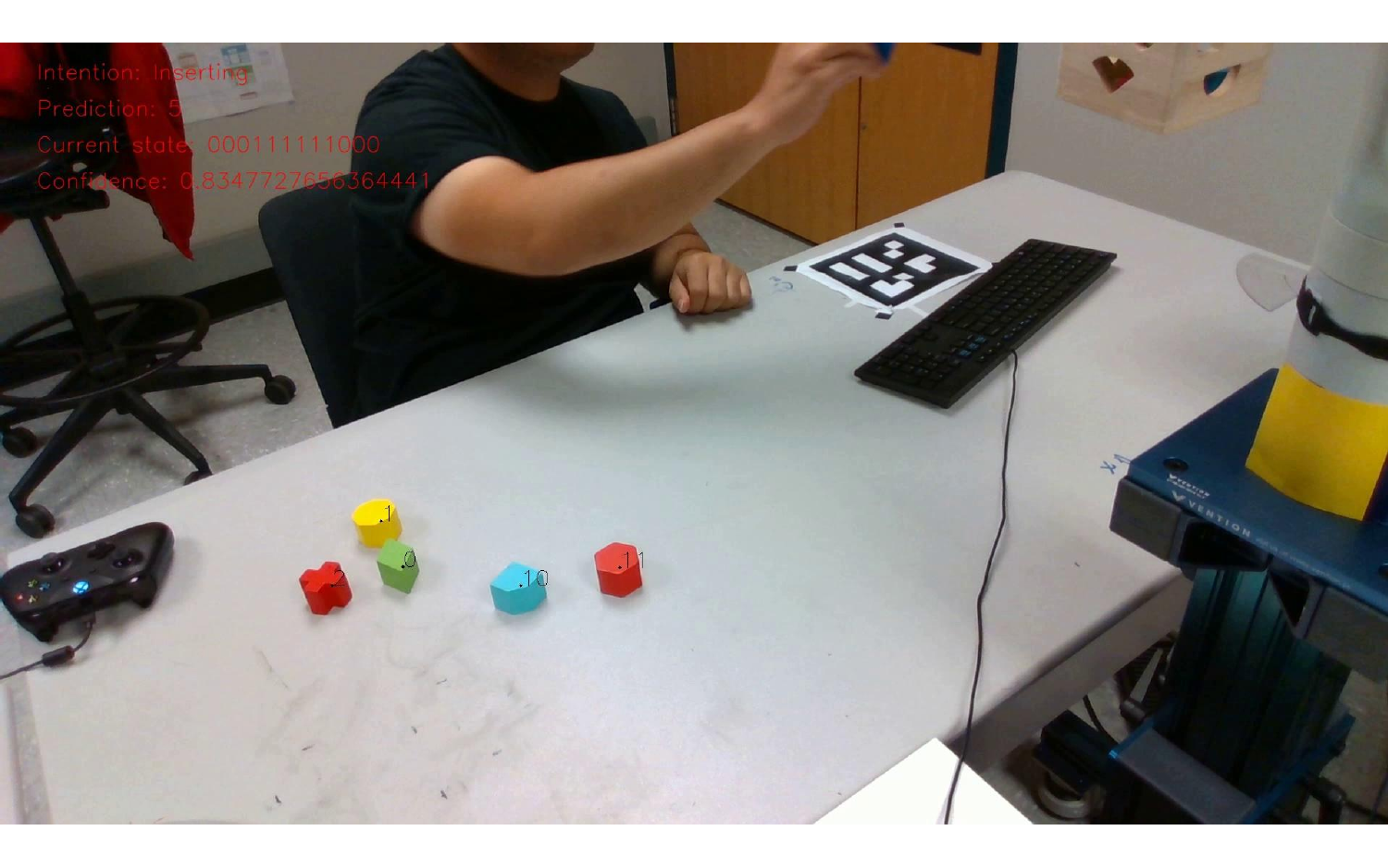}\label{fig:3_6}}\hfill
\subfigure[]{\includegraphics[width=0.12\linewidth]{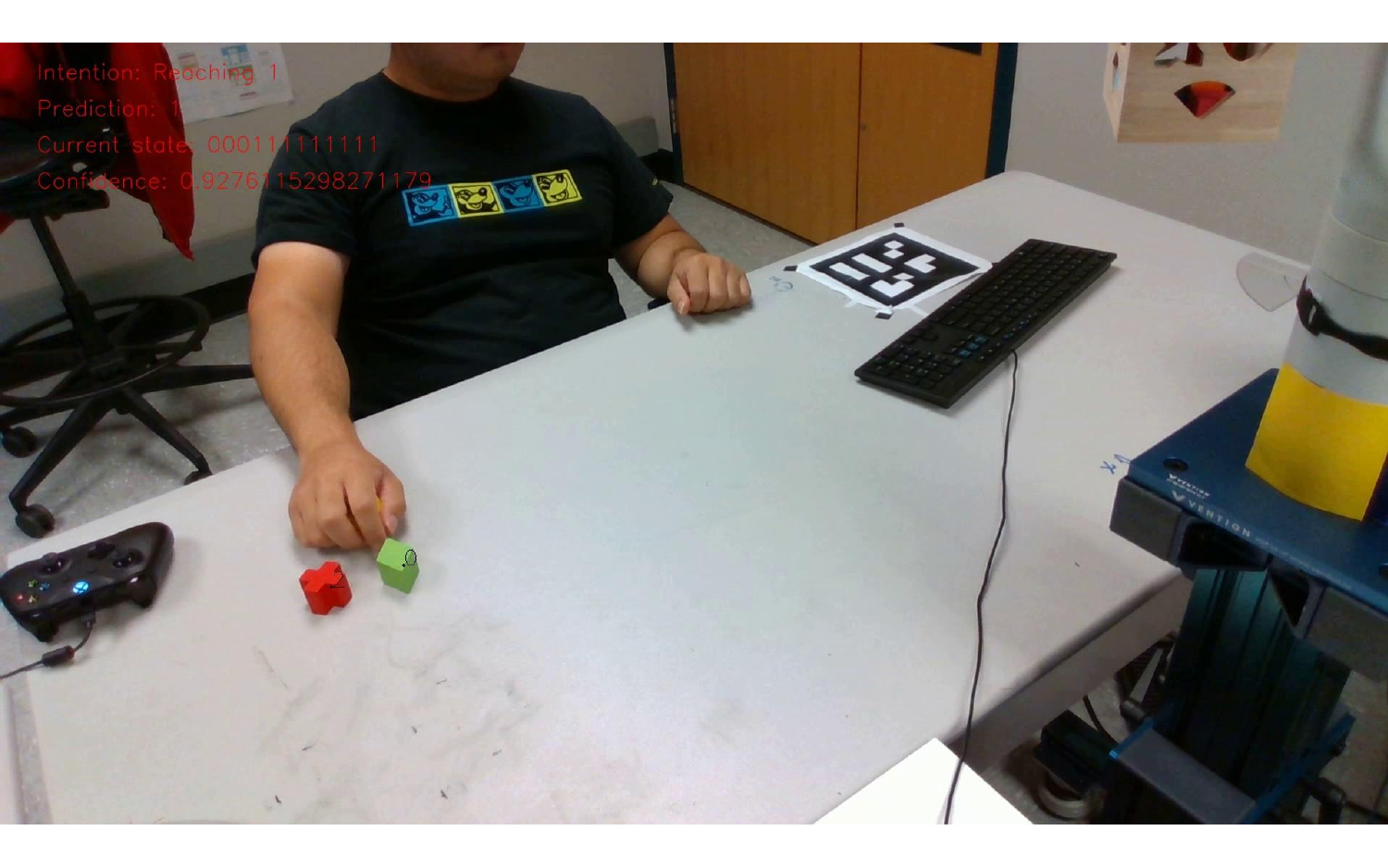}\label{fig:3_7}}\hfill
\subfigure[]{\includegraphics[width=0.12\linewidth]{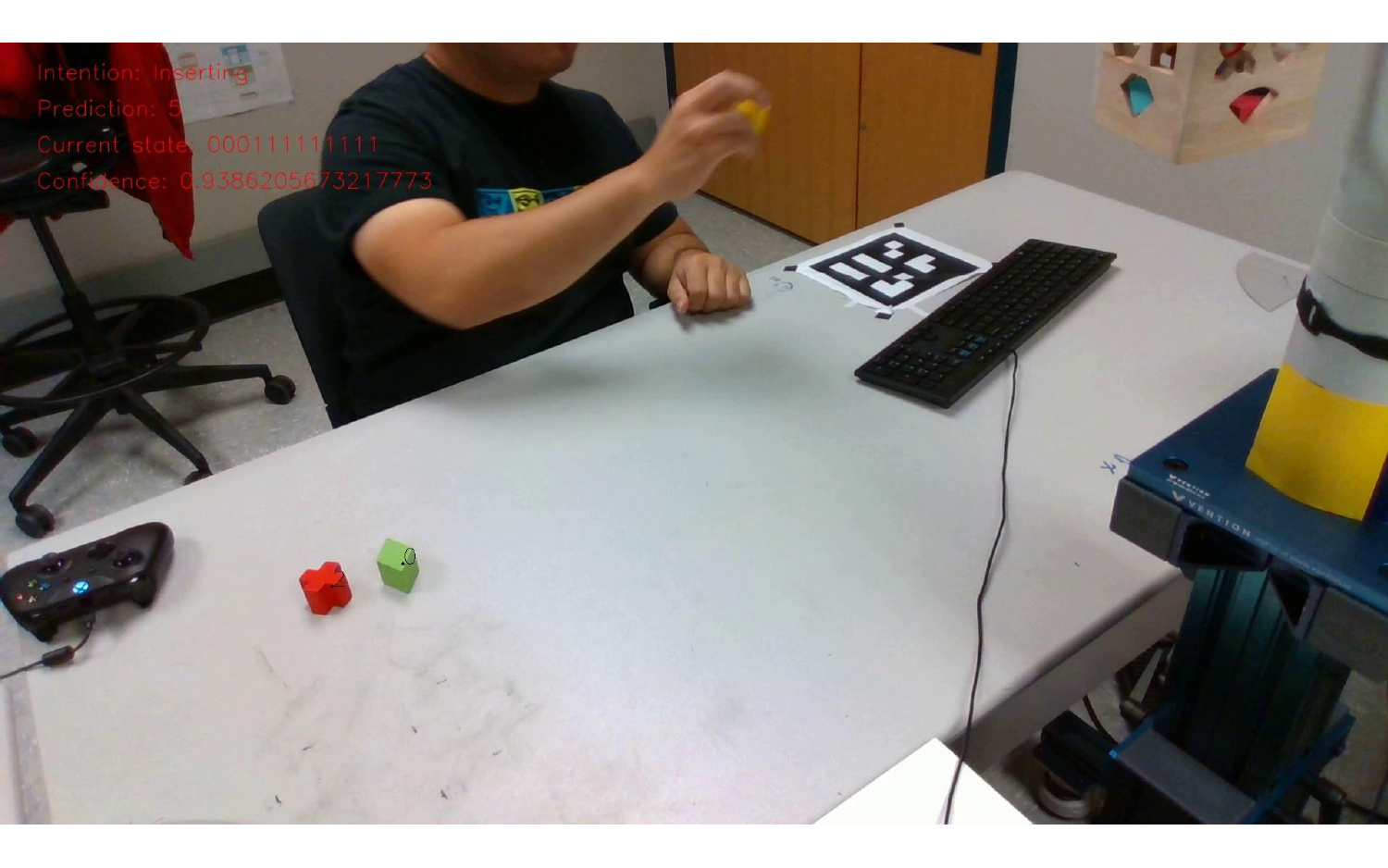}\label{fig:3_8}}
\vspace{-10pt}\\
\subfigure[]{\includegraphics[width=0.12\linewidth]{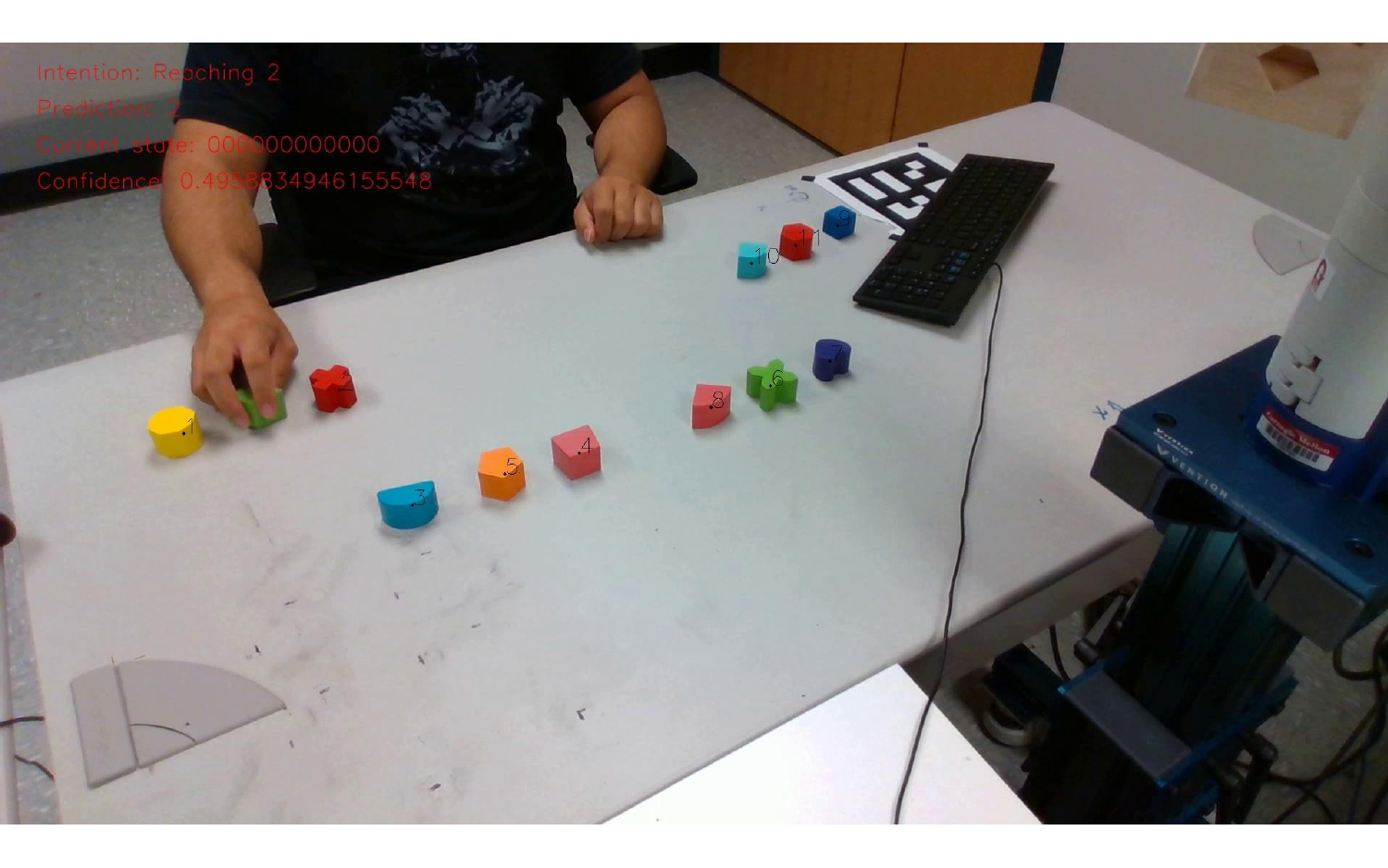}\label{fig:4_1}}\hfill
\subfigure[]{\includegraphics[width=0.12\linewidth]{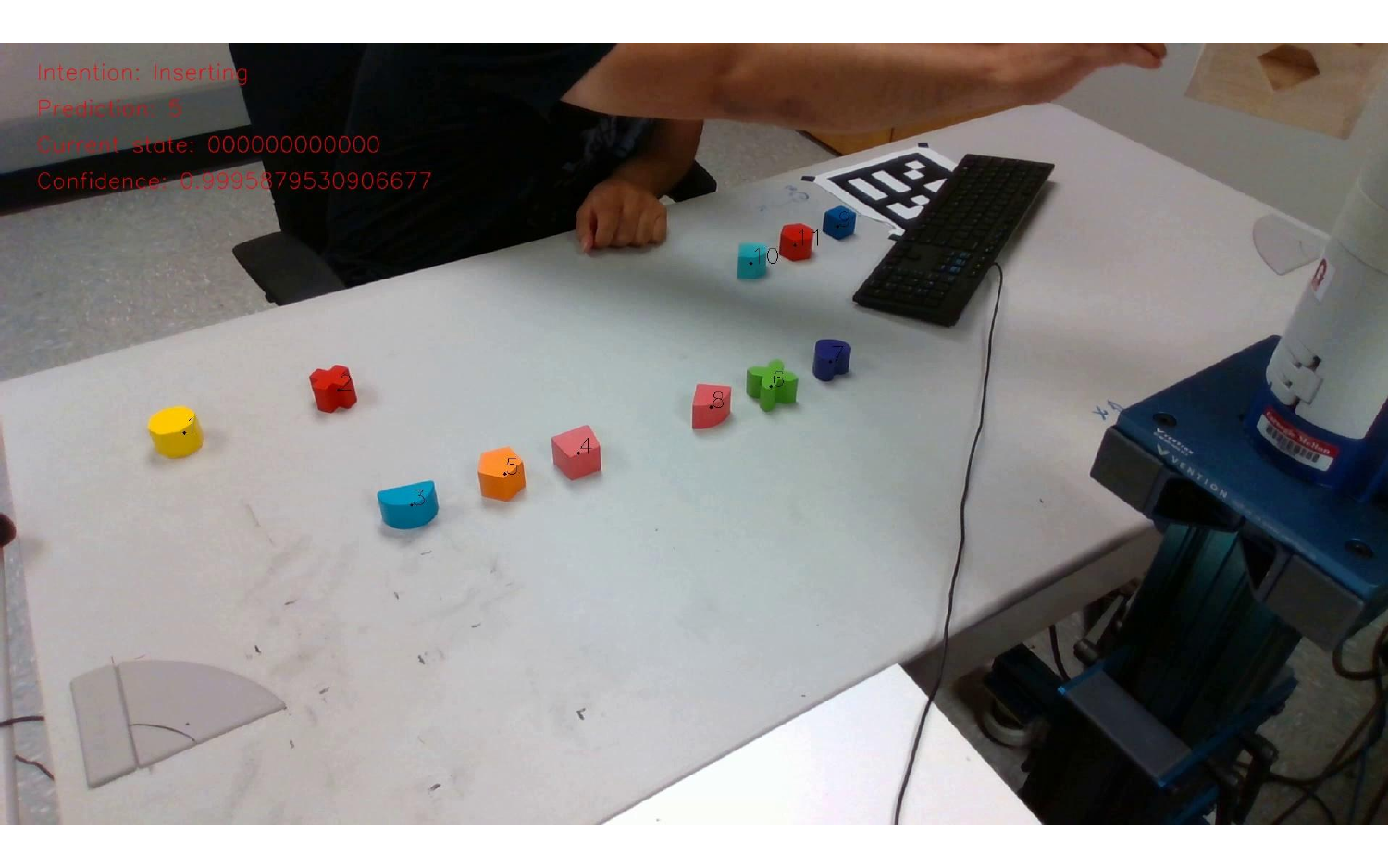}\label{fig:4_2}}\hfill
\subfigure[]{\includegraphics[width=0.12\linewidth]{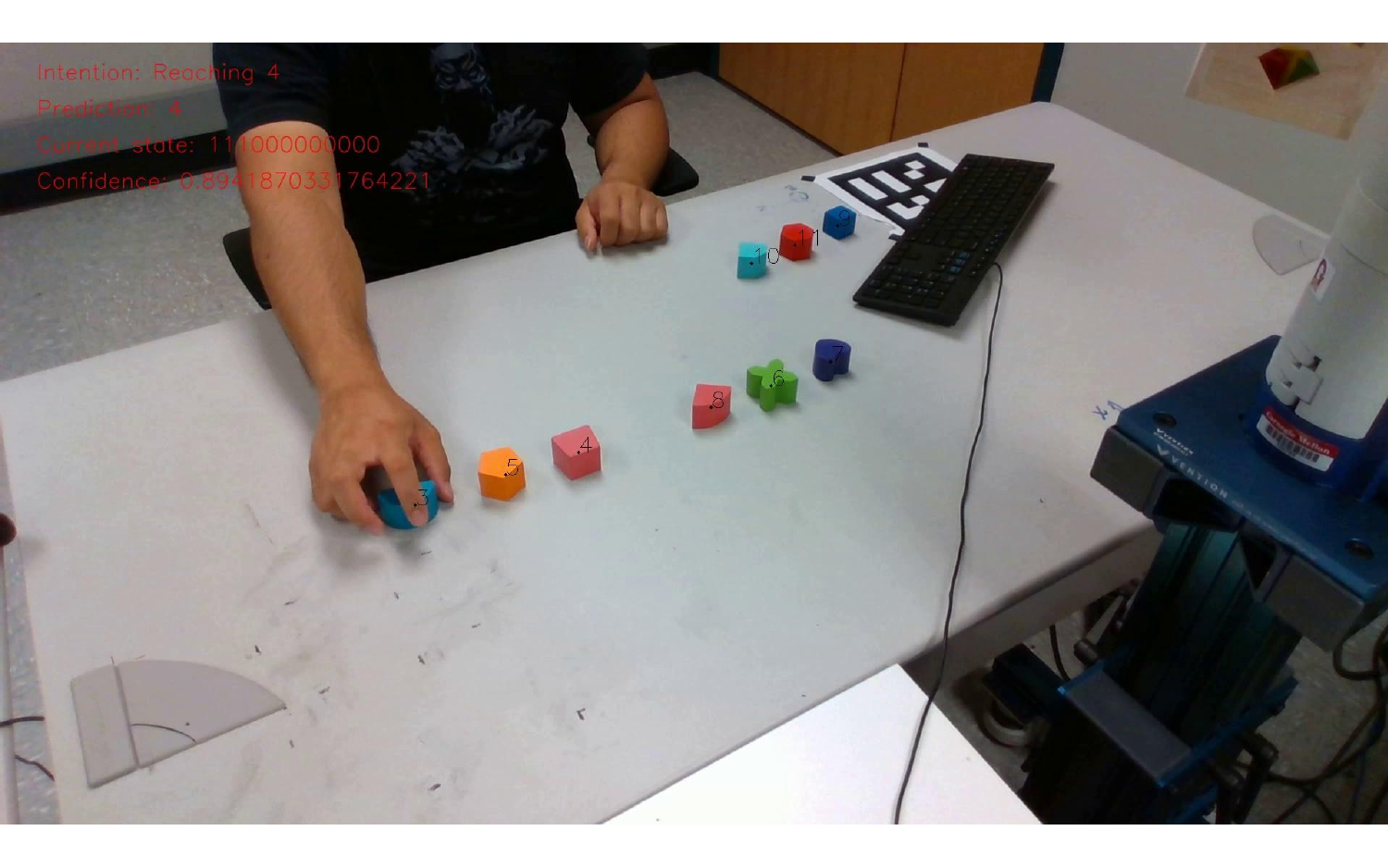}\label{fig:4_3}}\hfill
\subfigure[]{\includegraphics[width=0.12\linewidth]{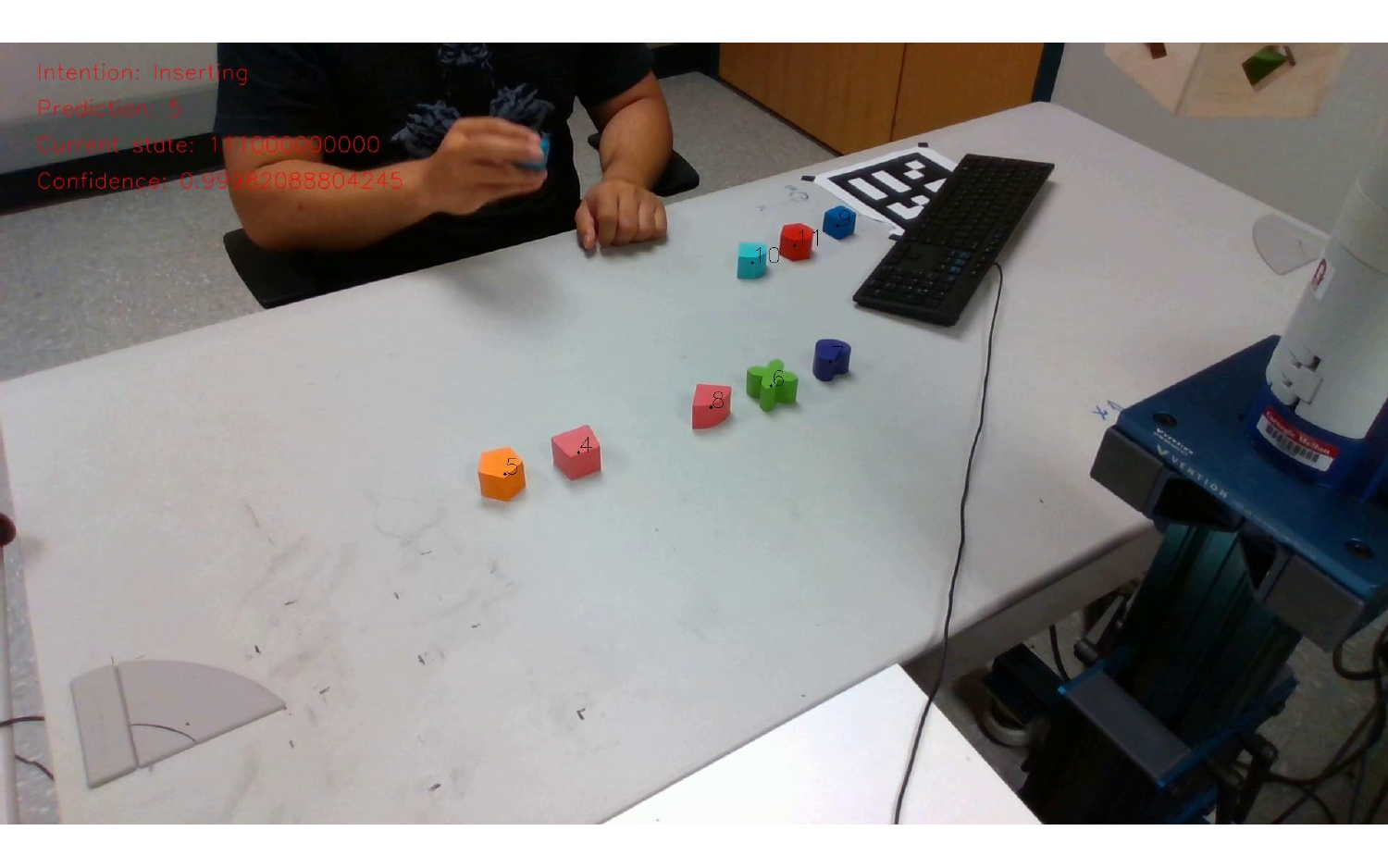}\label{fig:4_4}}\hfill
\subfigure[]{\includegraphics[width=0.12\linewidth]{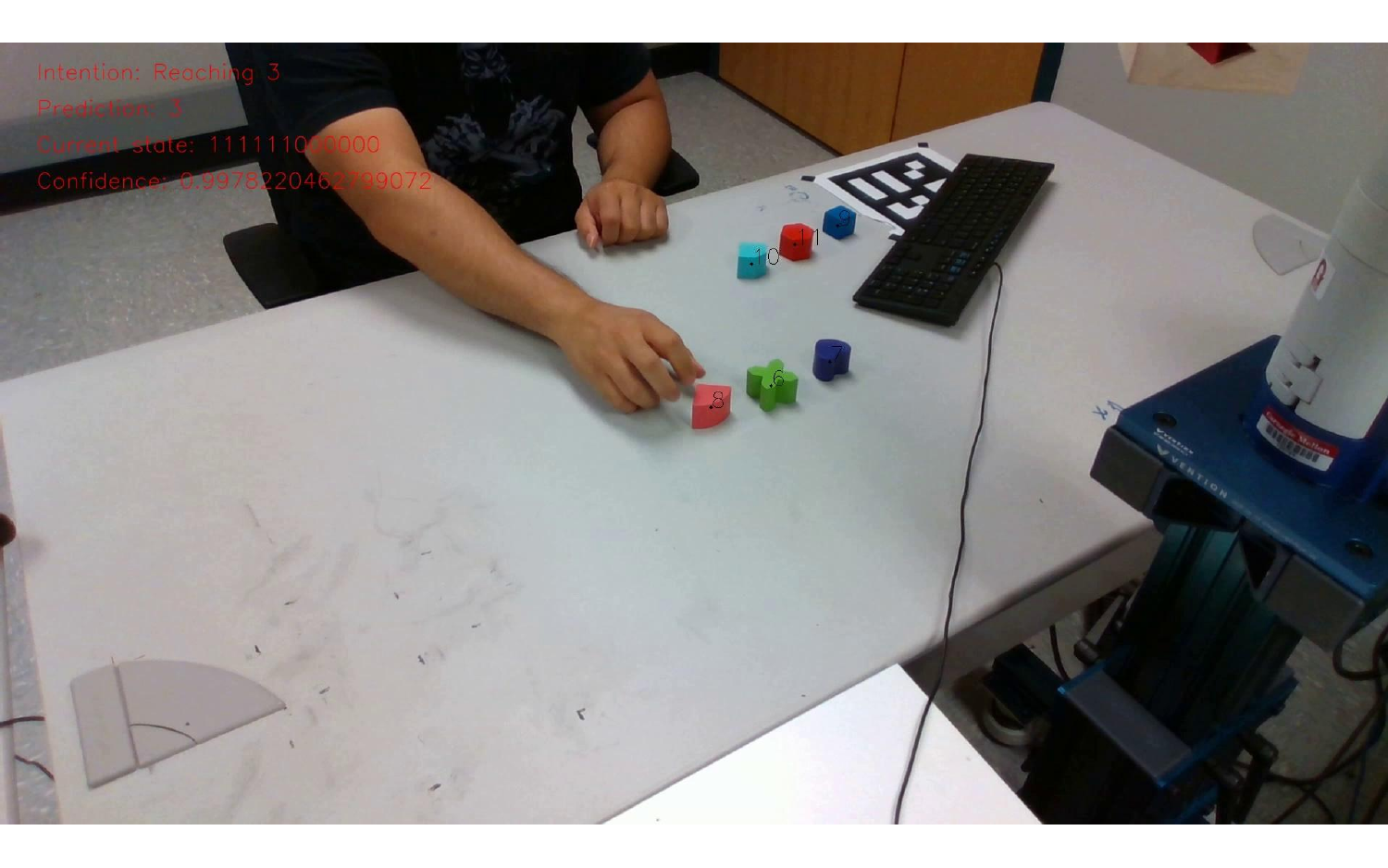}\label{fig:4_5}}\hfill
\subfigure[]{\includegraphics[width=0.12\linewidth]{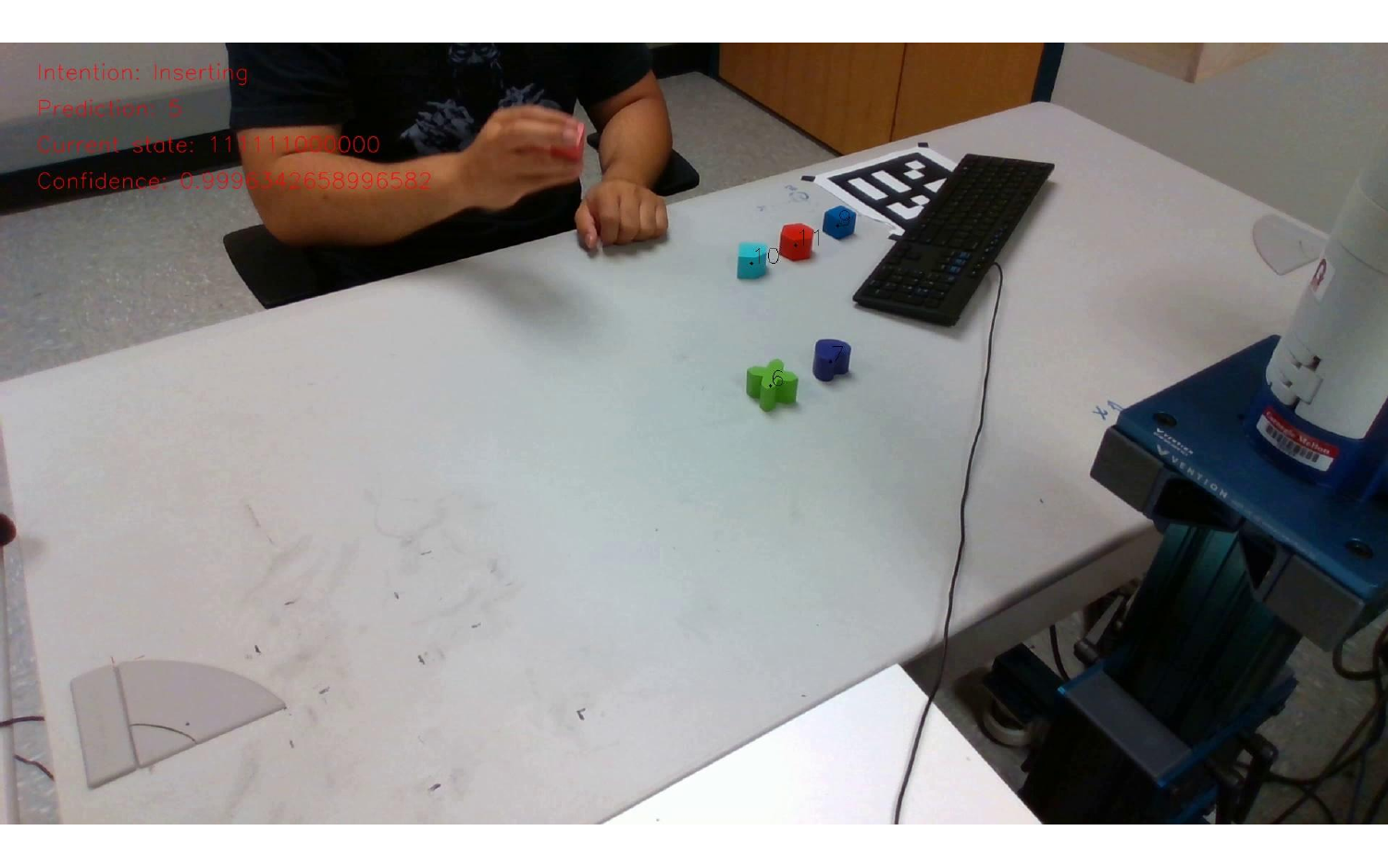}\label{fig:4_6}}\hfill
\subfigure[]{\includegraphics[width=0.12\linewidth]{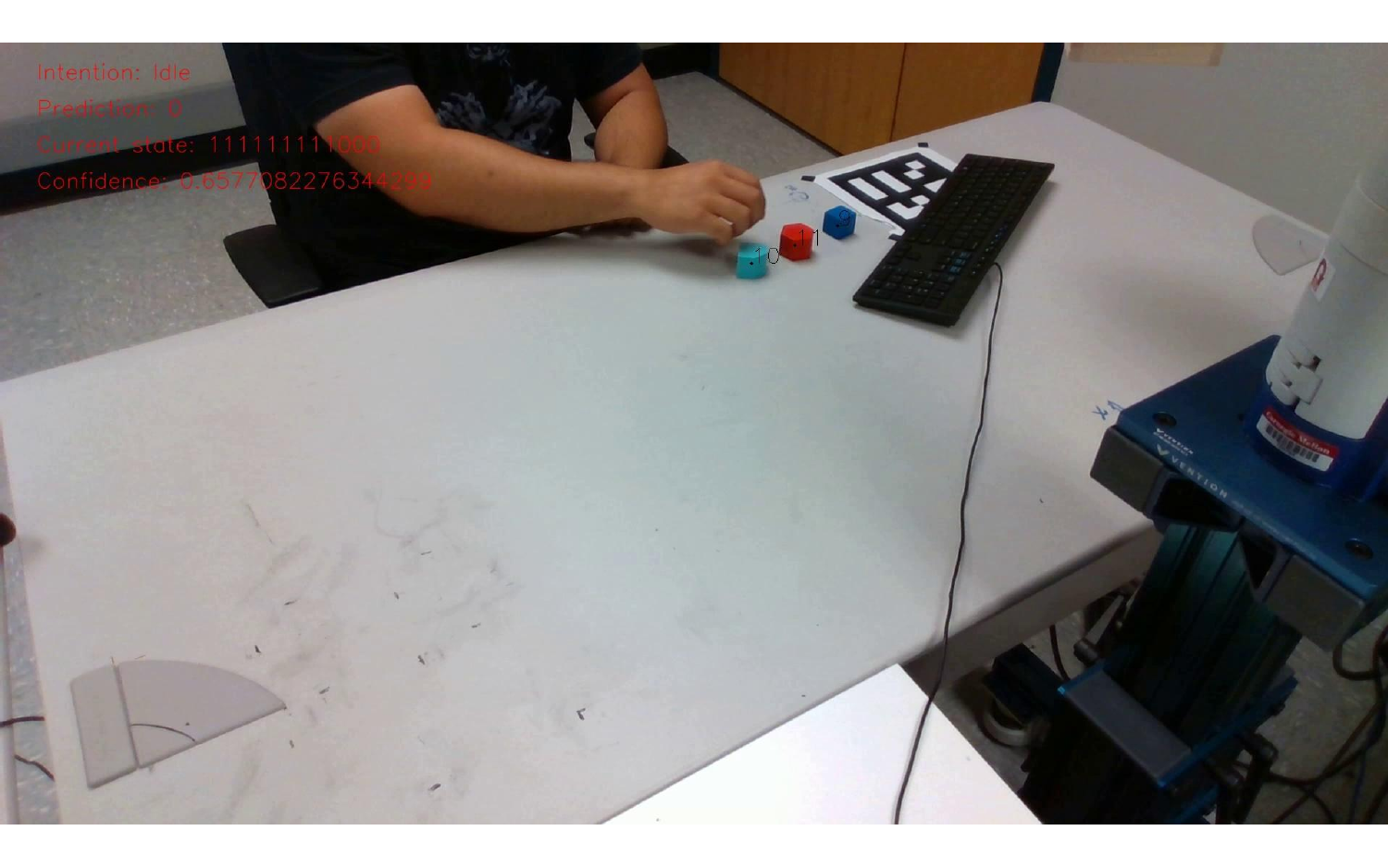}\label{fig:4_7}}\hfill
\subfigure[]{\includegraphics[width=0.12\linewidth]{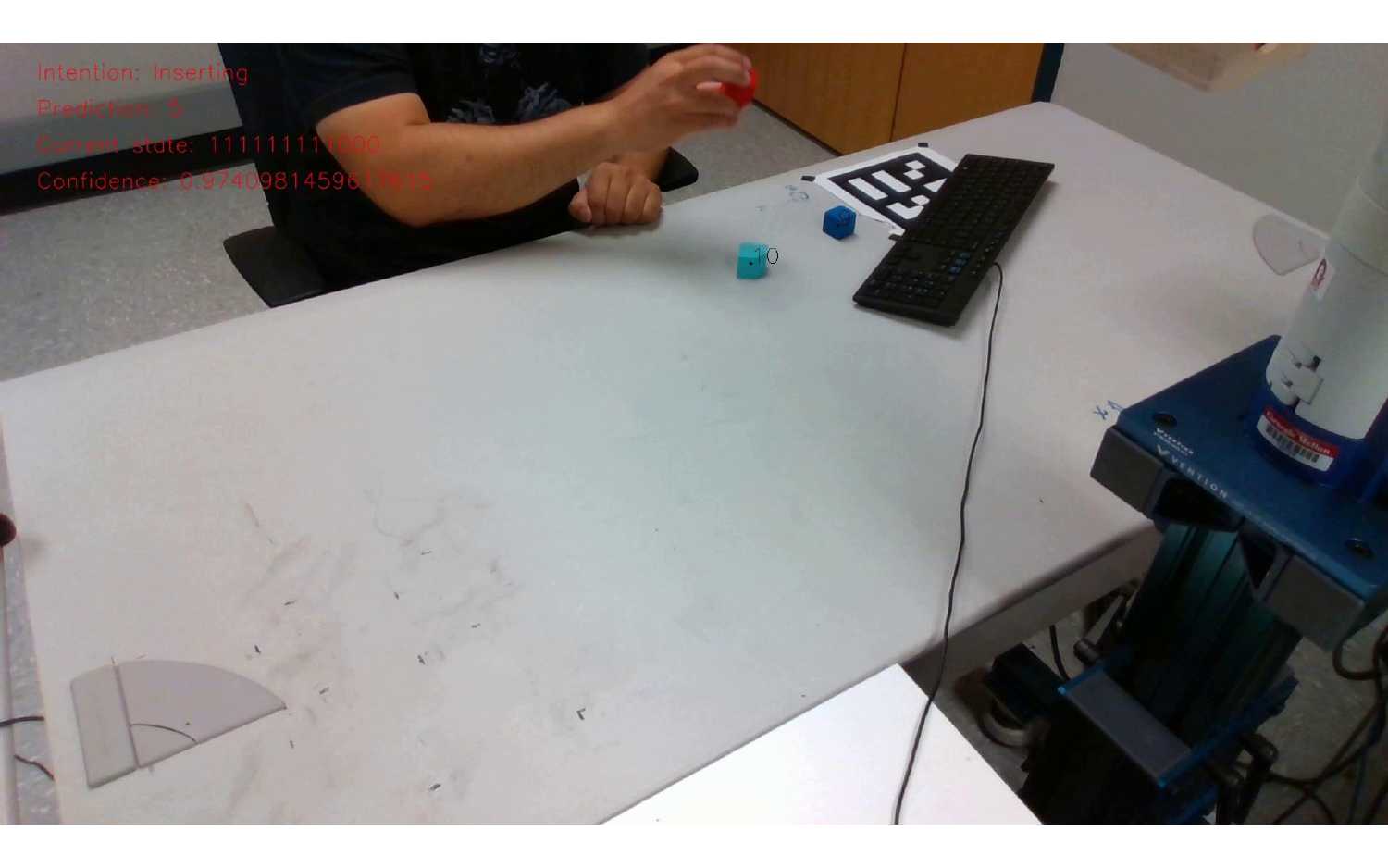}\label{fig:4_8}}
\vspace{-10pt}\\
\subfigure[]{\includegraphics[width=0.12\linewidth]{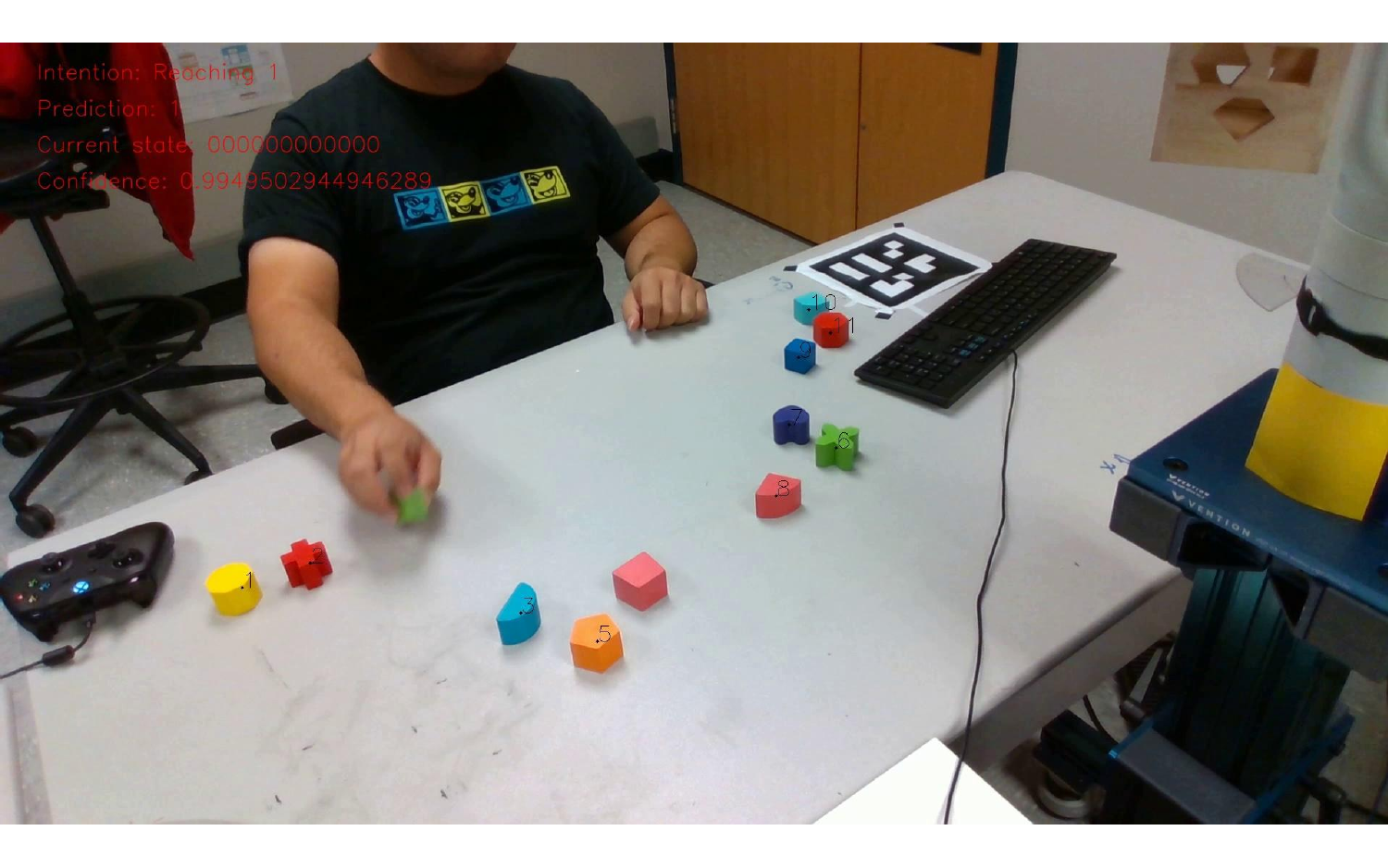}\label{fig:5_1}}\hfill
\subfigure[]{\includegraphics[width=0.12\linewidth]{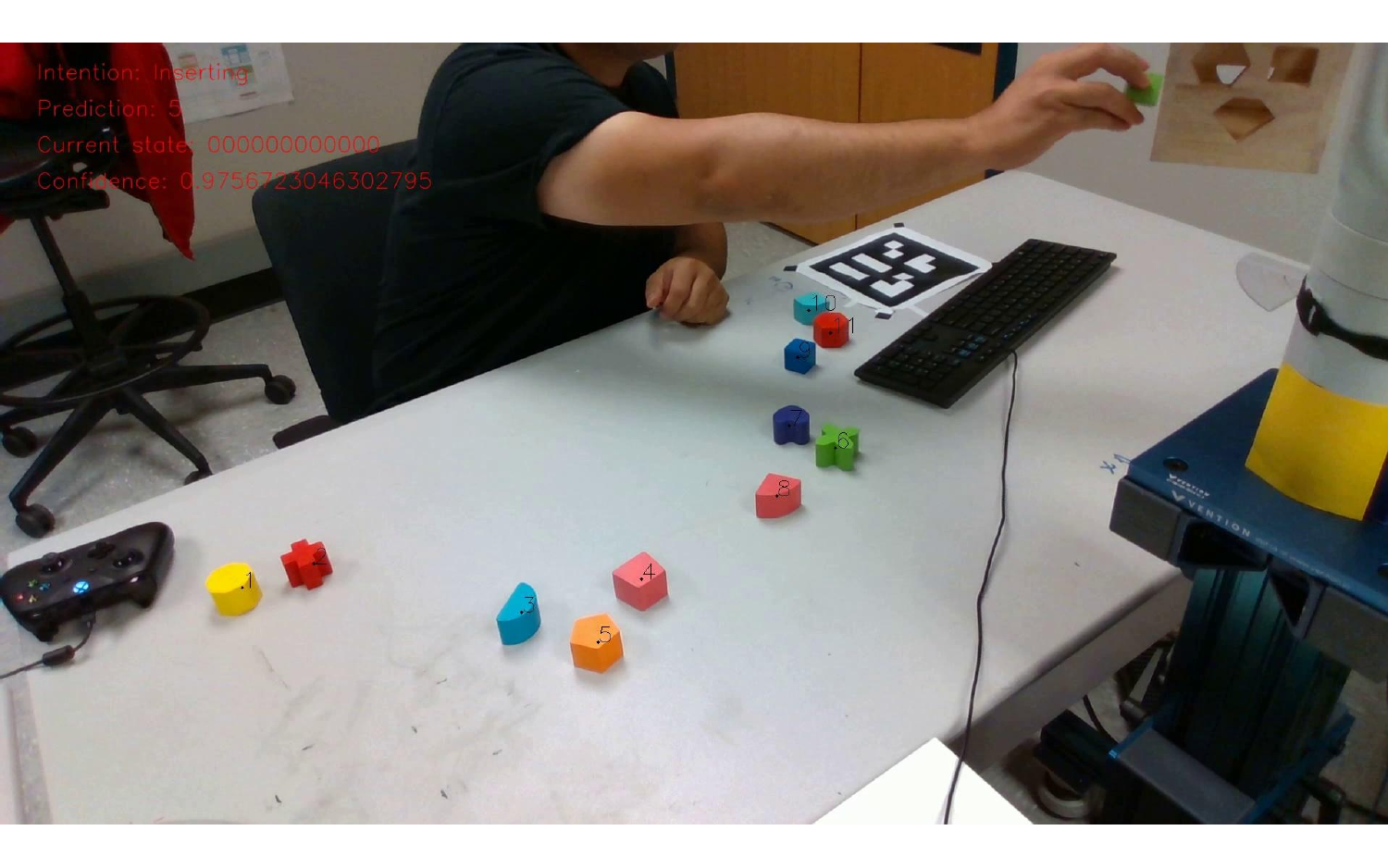}\label{fig:5_2}}\hfill
\subfigure[]{\includegraphics[width=0.12\linewidth]{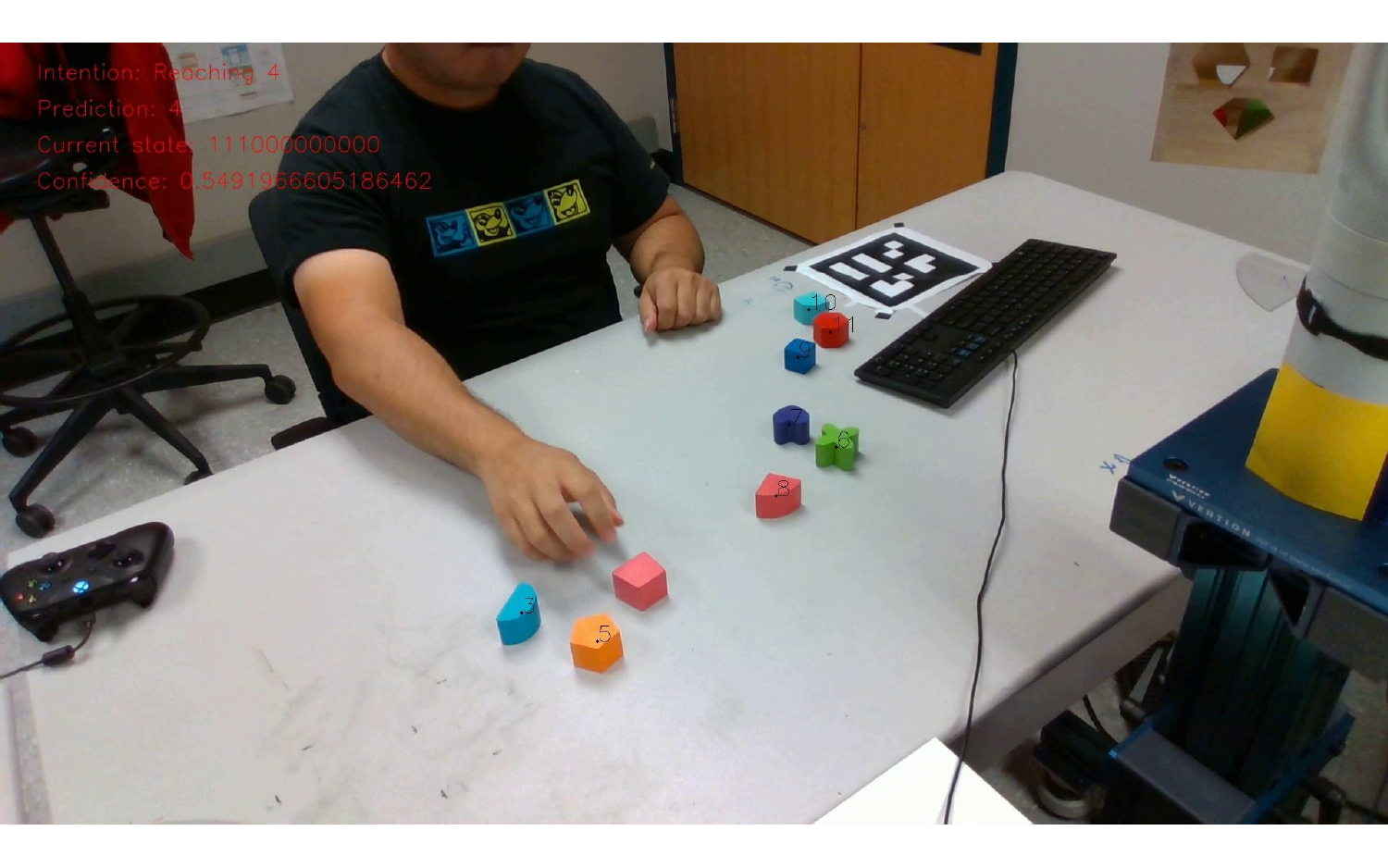}\label{fig:5_3}}\hfill
\subfigure[]{\includegraphics[width=0.12\linewidth]{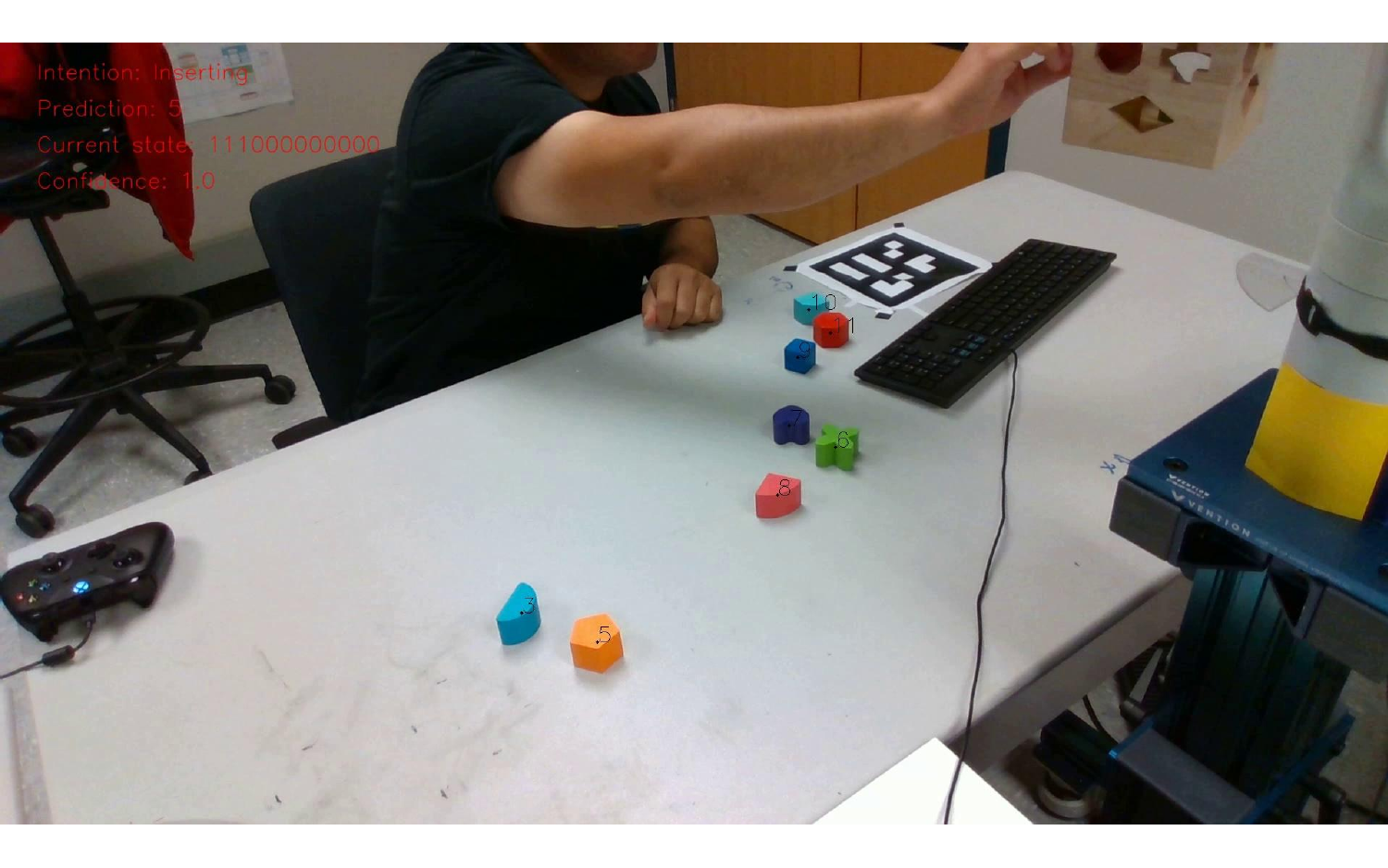}\label{fig:5_4}}\hfill
\subfigure[]{\includegraphics[width=0.12\linewidth]{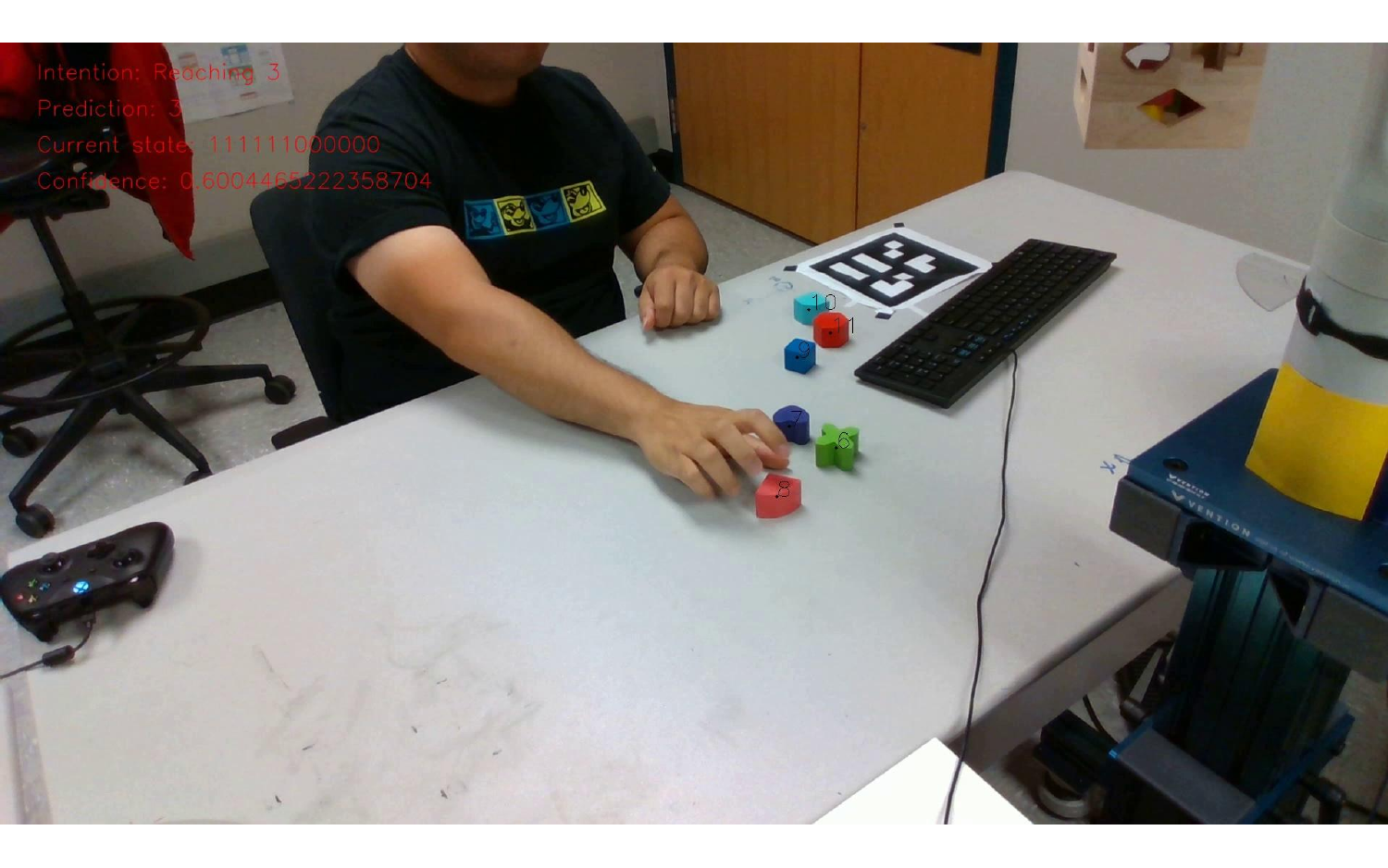}\label{fig:5_5}}\hfill
\subfigure[]{\includegraphics[width=0.12\linewidth]{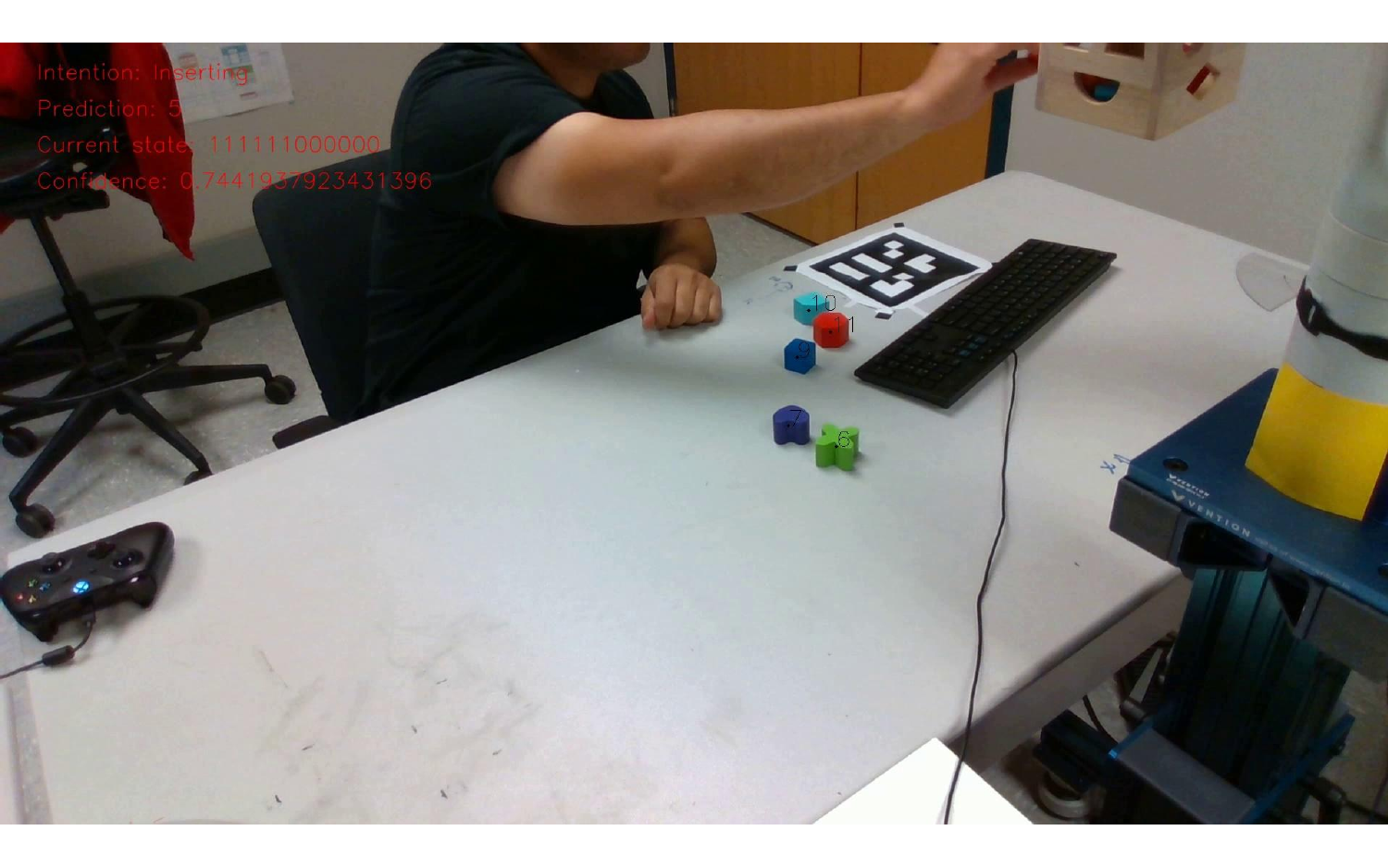}\label{fig:5_6}}\hfill
\subfigure[]{\includegraphics[width=0.12\linewidth]{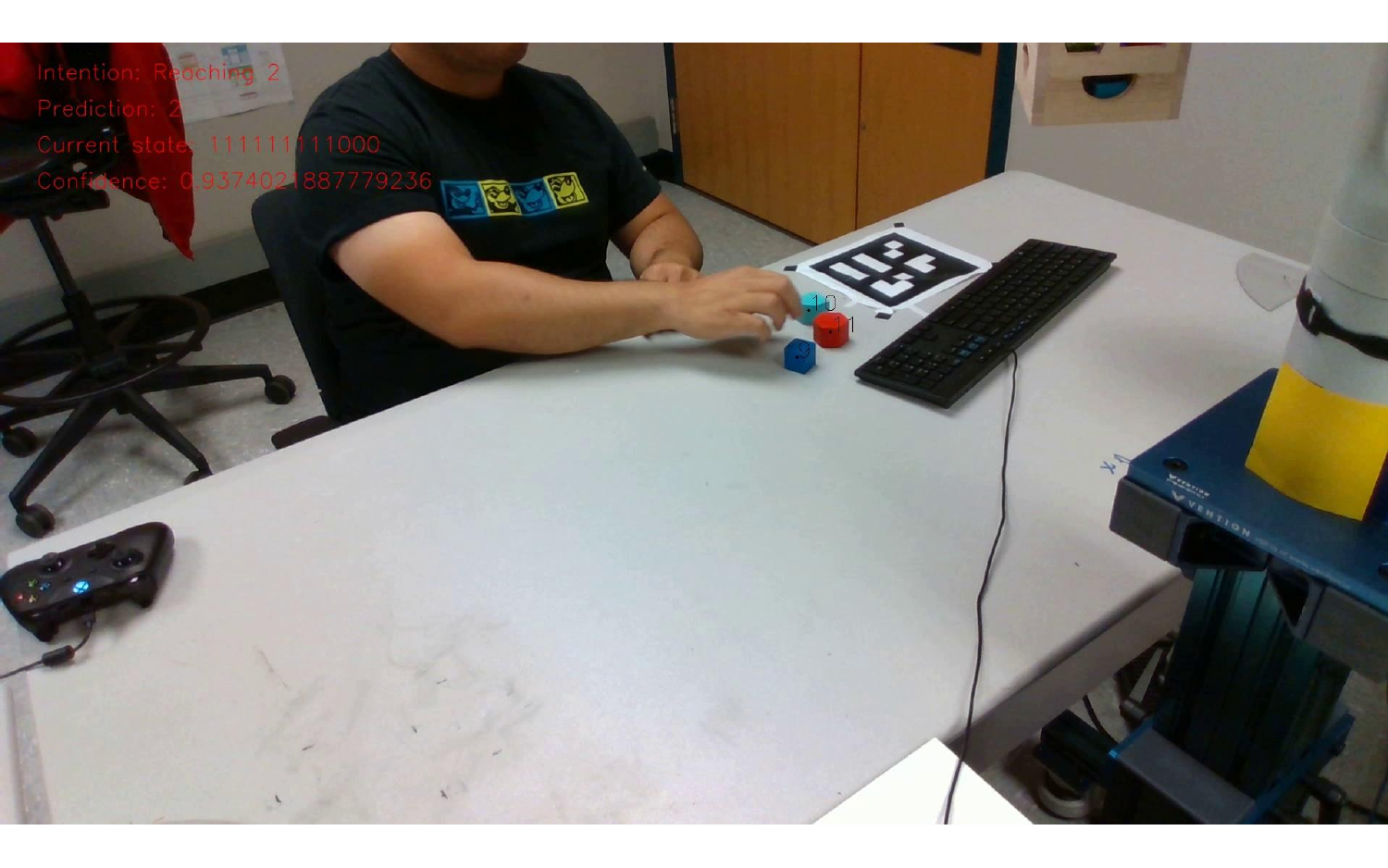}\label{fig:5_7}}\hfill
\subfigure[]{\includegraphics[width=0.12\linewidth]{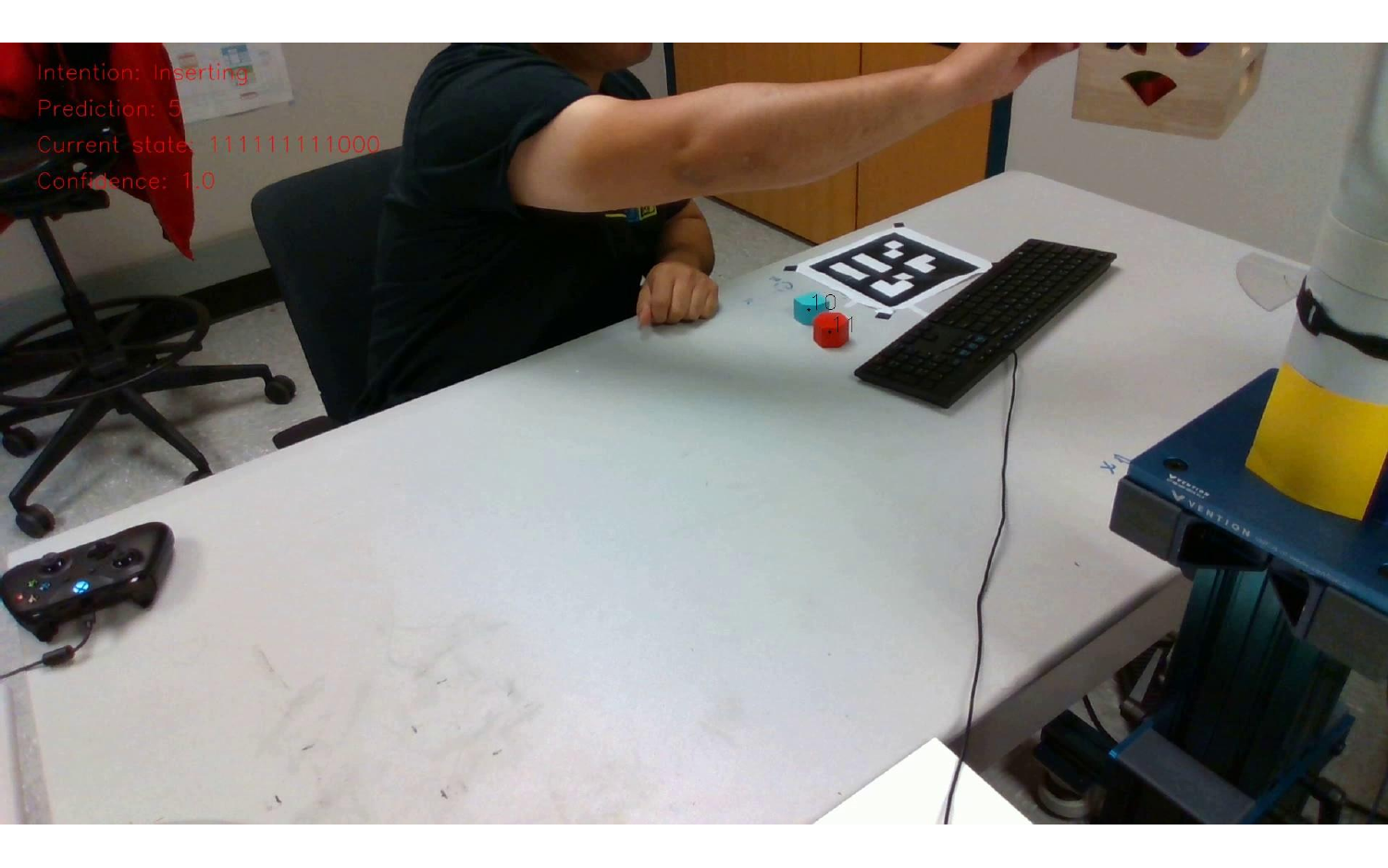}\label{fig:5_8}}
\vspace{-10pt}\\
\subfigure[]{\includegraphics[width=0.12\linewidth]{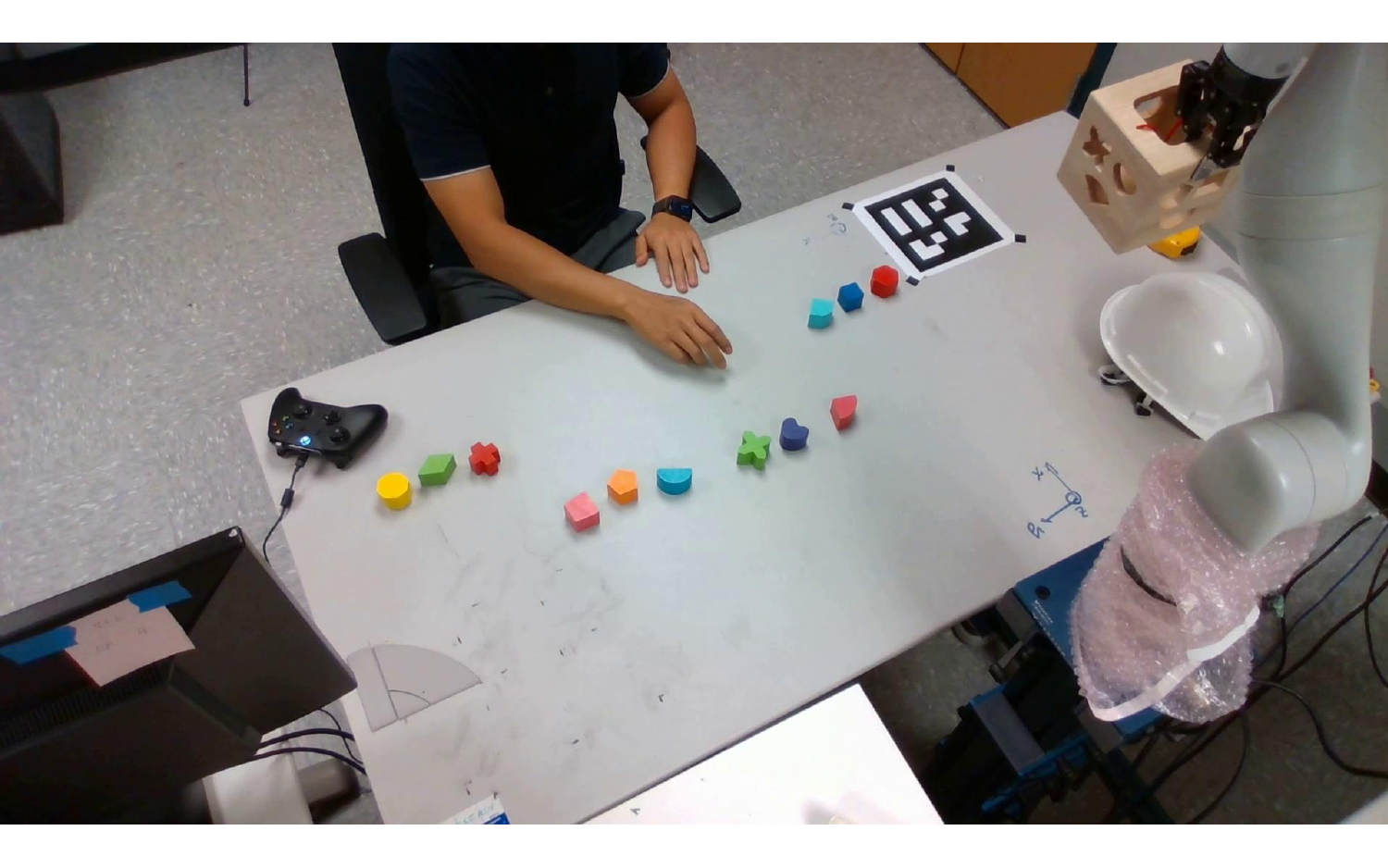}\label{fig:6_1}}\hfill
\subfigure[]{\includegraphics[width=0.12\linewidth]{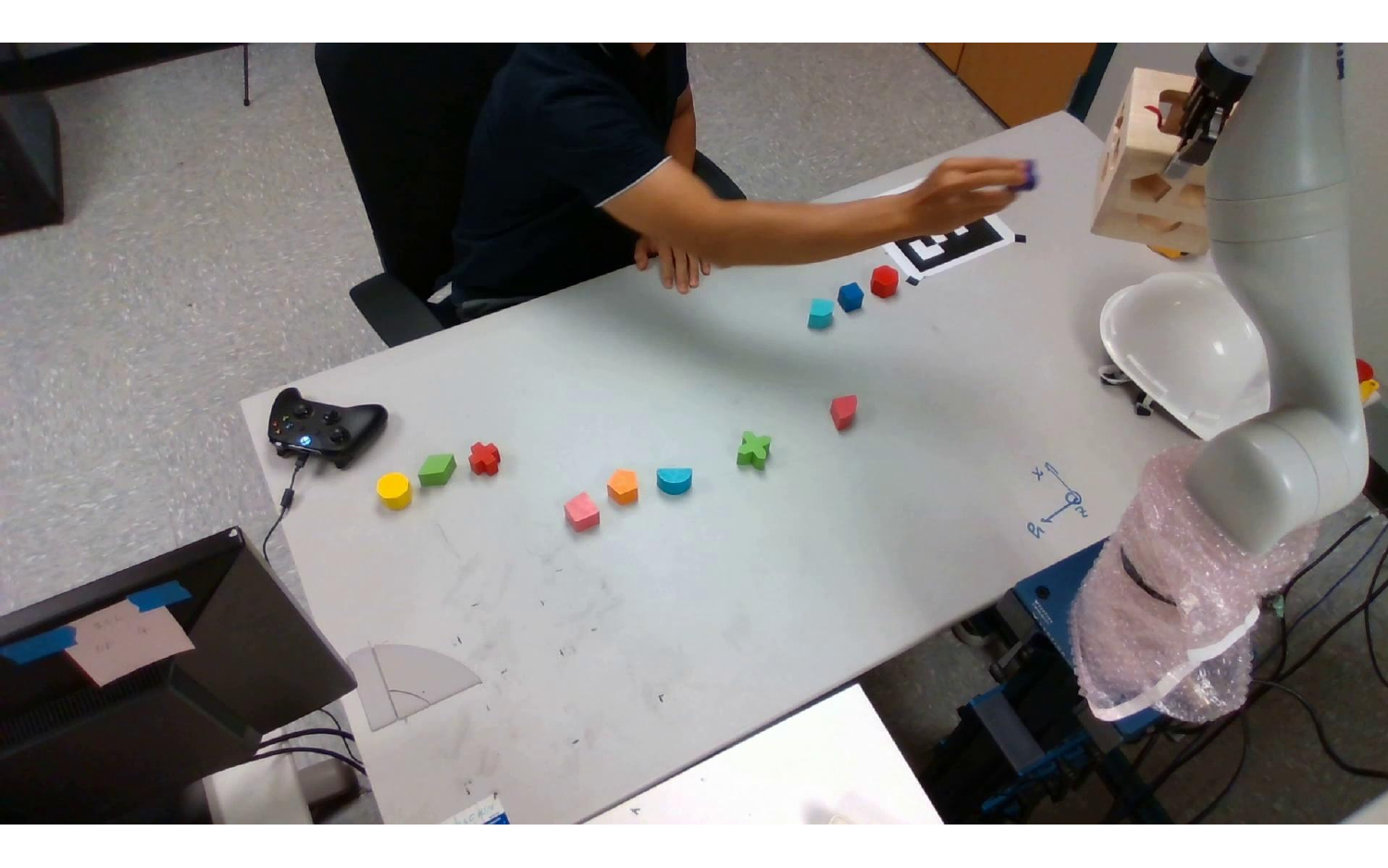}\label{fig:6_2}}\hfill
\subfigure[]{\includegraphics[width=0.12\linewidth]{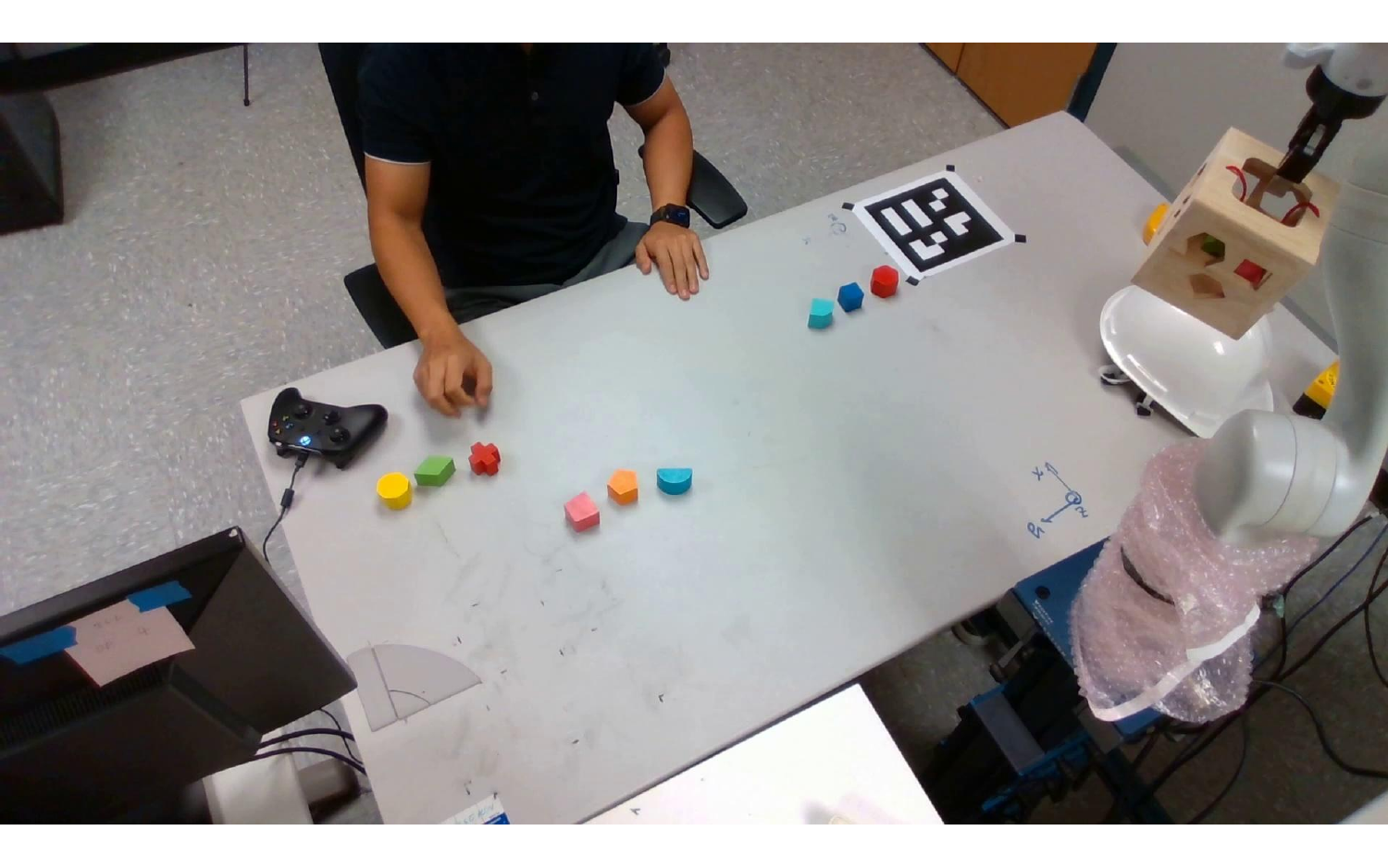}\label{fig:6_3}}\hfill
\subfigure[]{\includegraphics[width=0.12\linewidth]{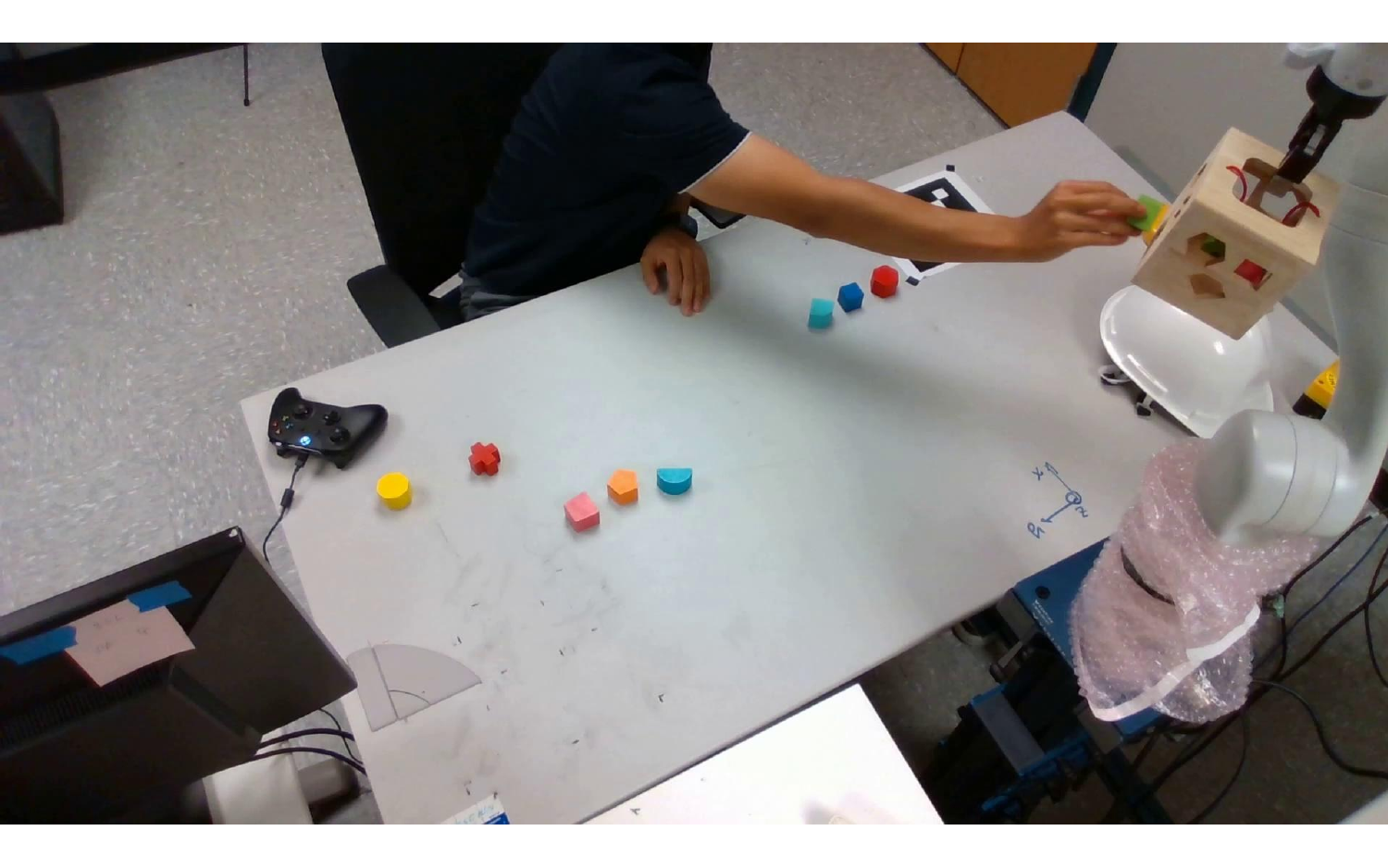}\label{fig:6_4}}\hfill
\subfigure[]{\includegraphics[width=0.12\linewidth]{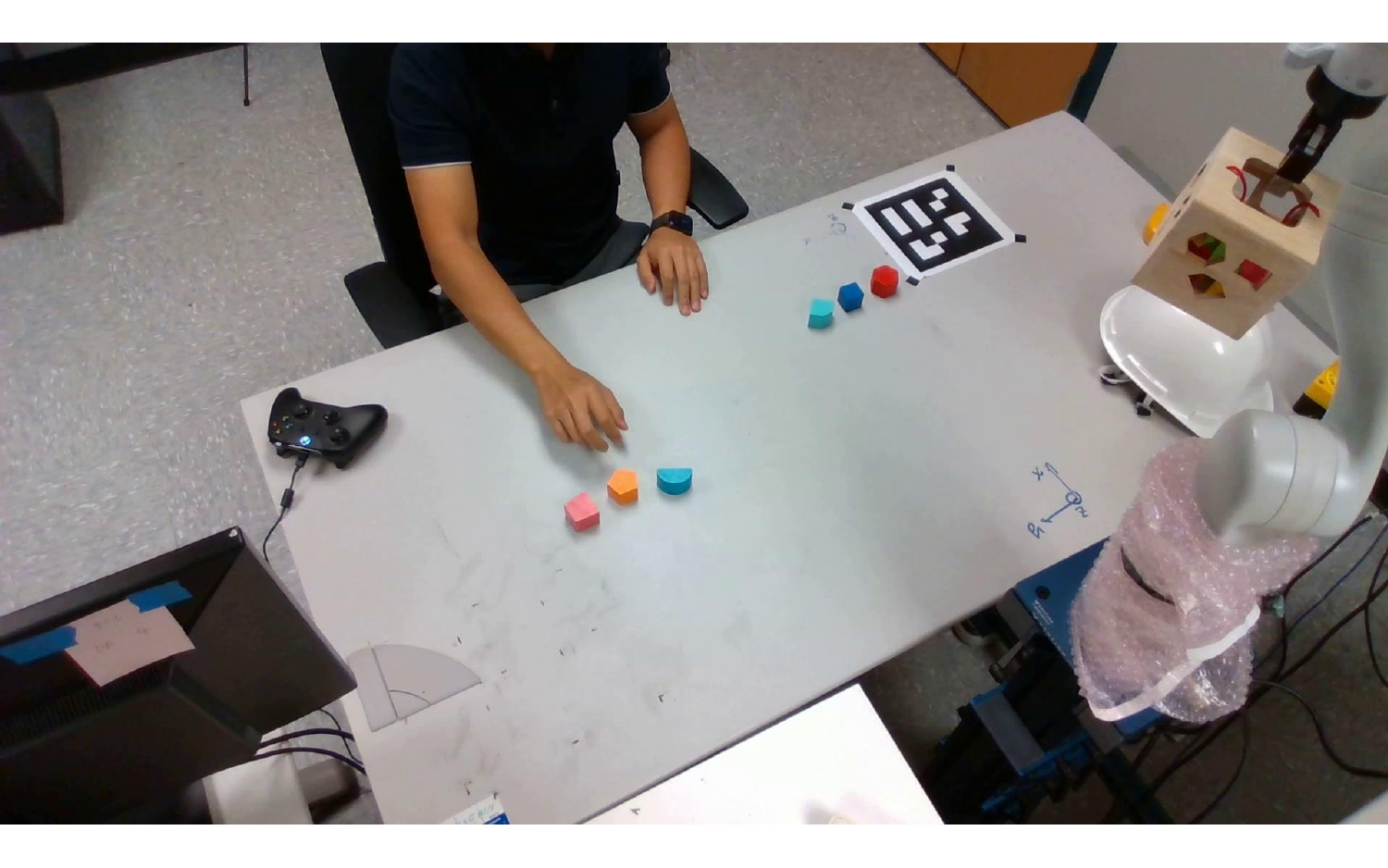}\label{fig:6_5}}\hfill
\subfigure[]{\includegraphics[width=0.12\linewidth]{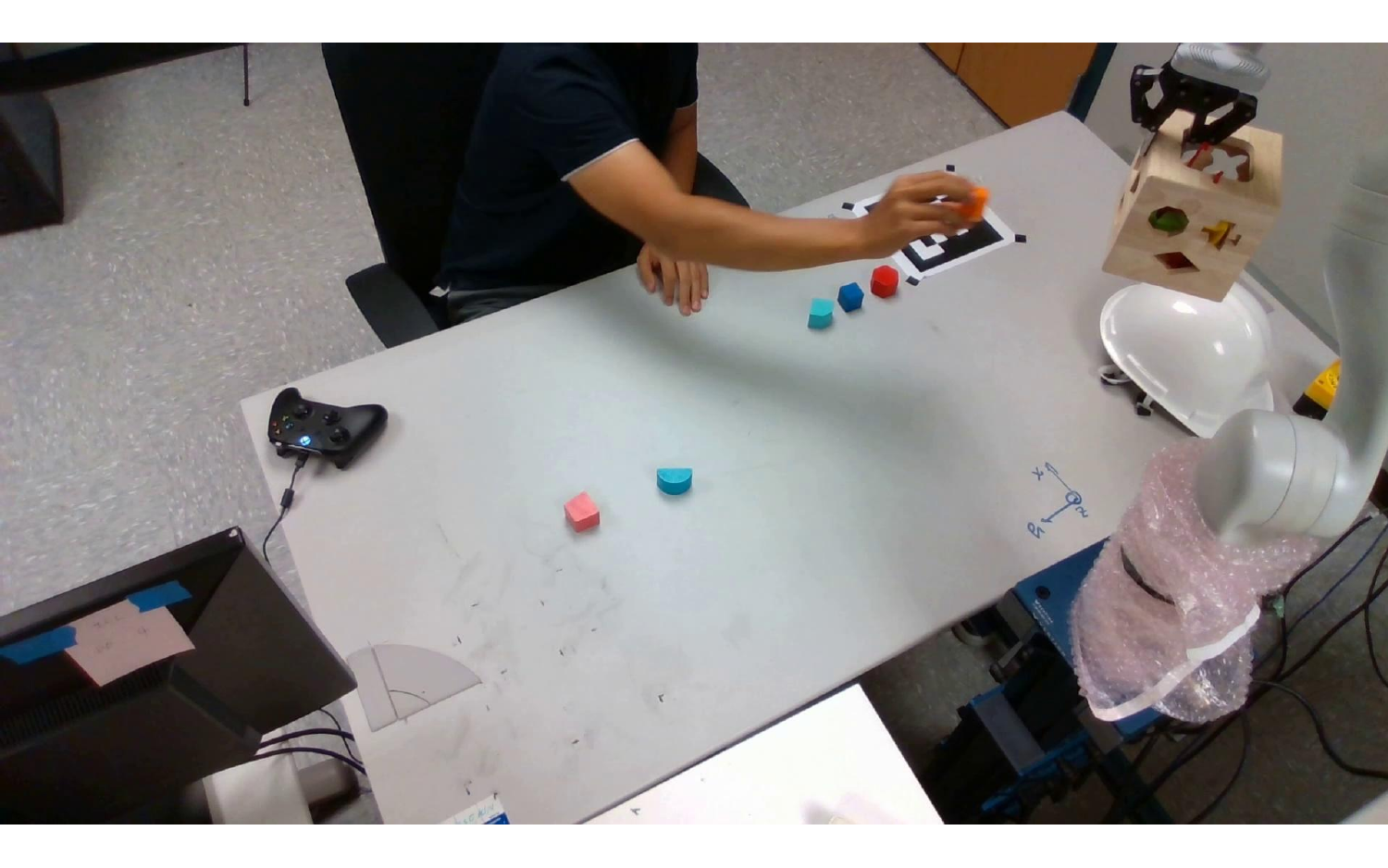}\label{fig:6_6}}\hfill
\subfigure[]{\includegraphics[width=0.12\linewidth]{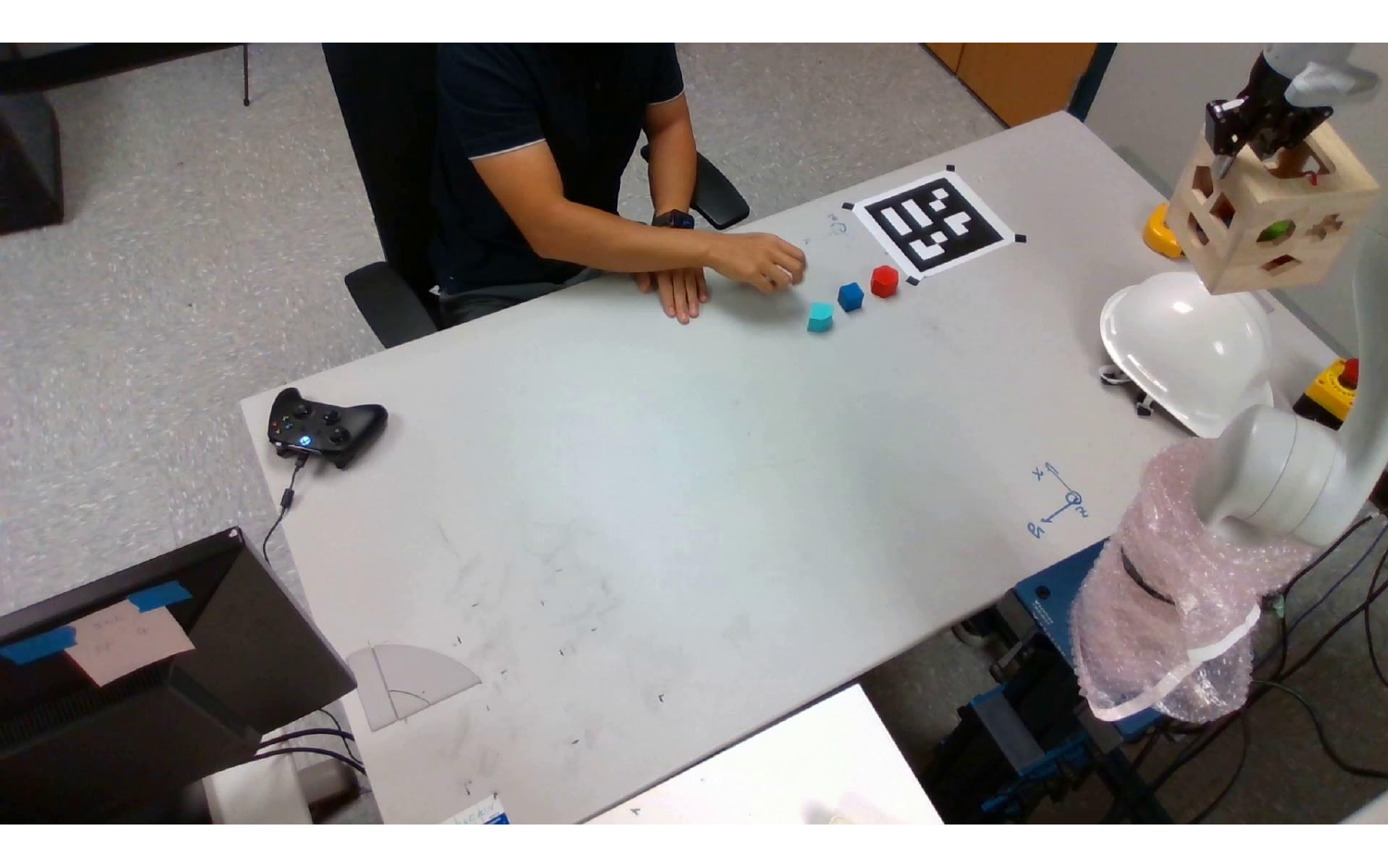}\label{fig:6_7}}\hfill
\subfigure[]{\includegraphics[width=0.12\linewidth]{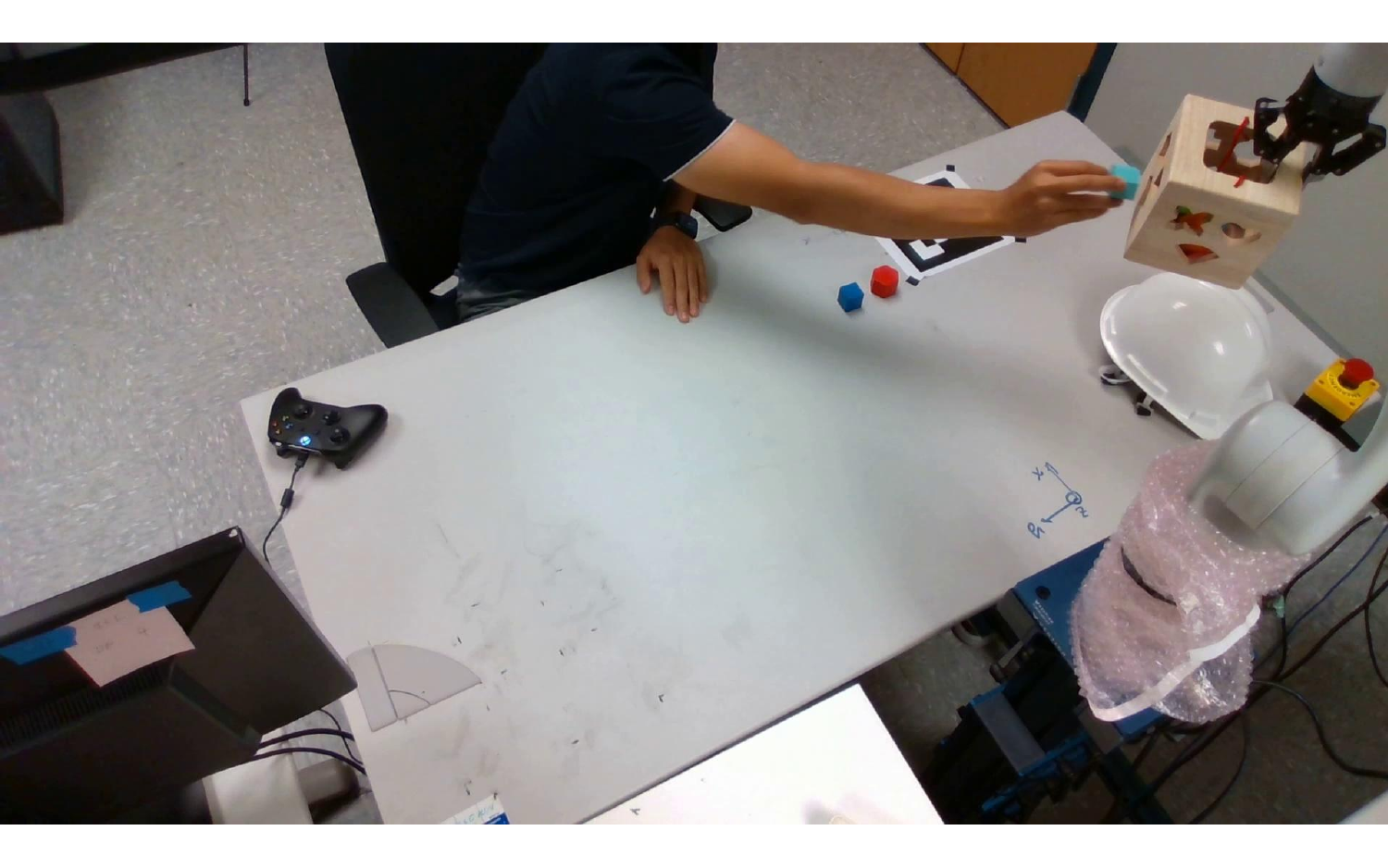}\label{fig:6_8}}
\vspace{-10pt}\\
\subfigure[]{\includegraphics[width=0.12\linewidth]{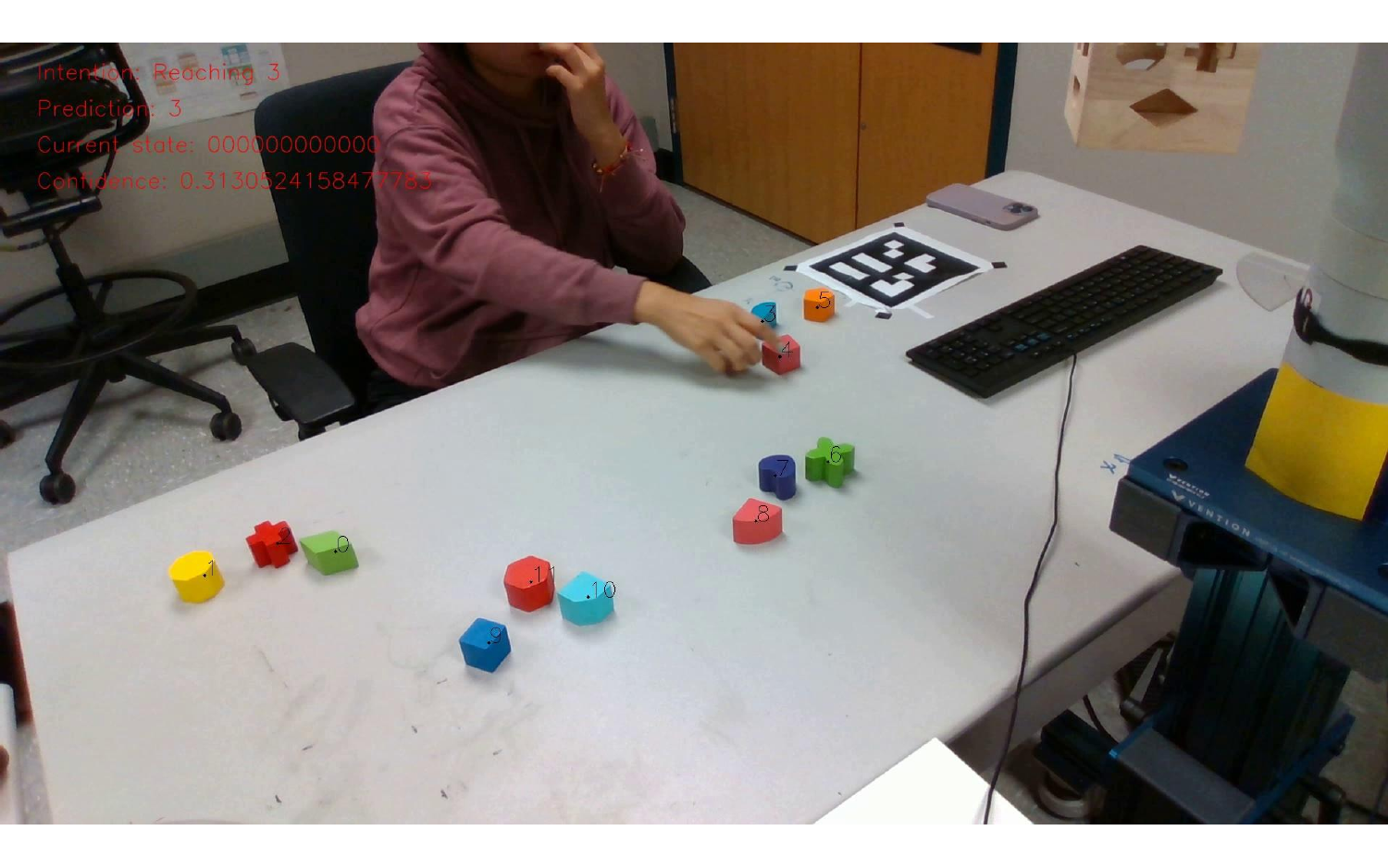}\label{fig:7_1}}\hfill
\subfigure[]{\includegraphics[width=0.12\linewidth]{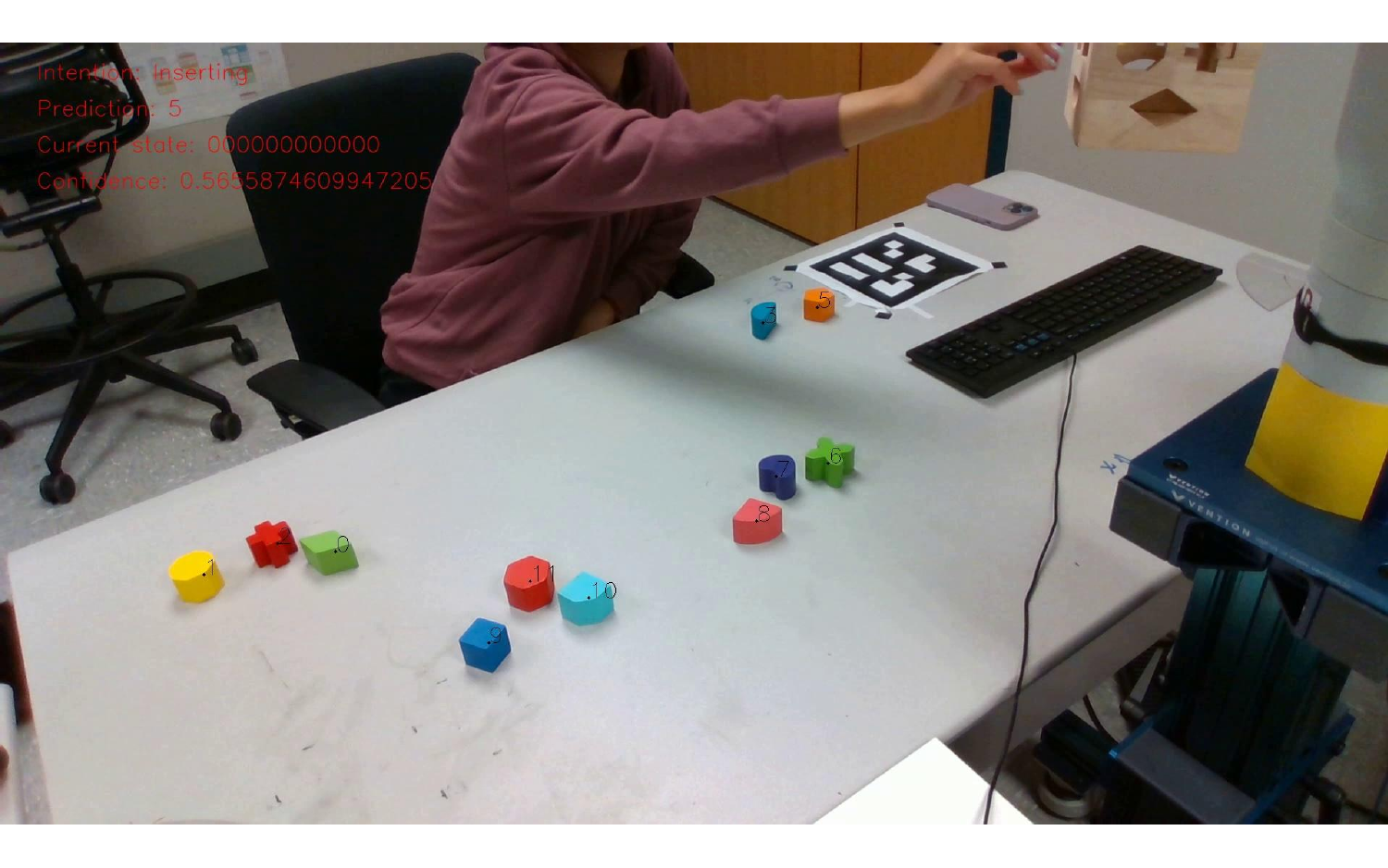}\label{fig:7_2}}\hfill
\subfigure[]{\includegraphics[width=0.12\linewidth]{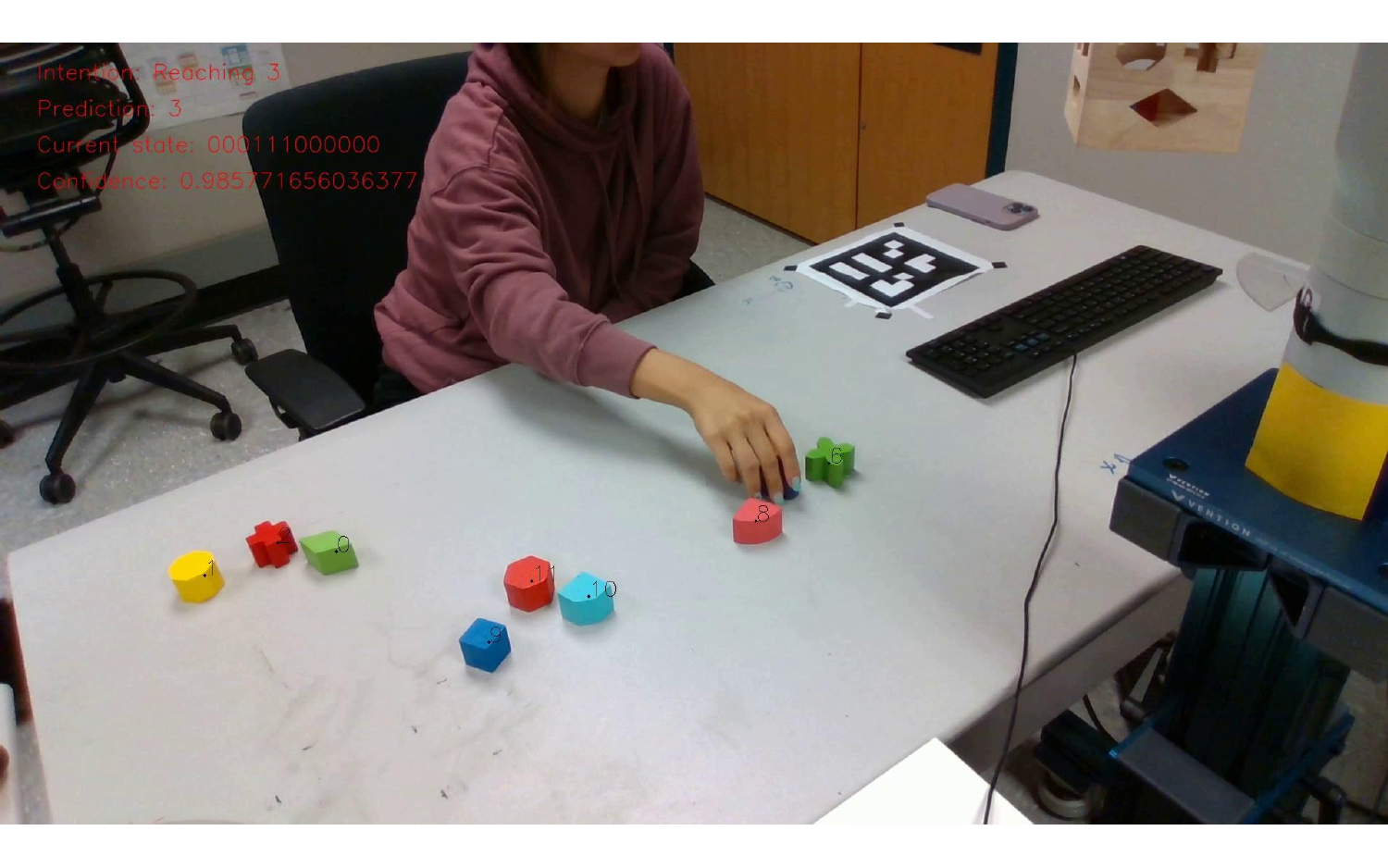}\label{fig:7_3}}\hfill
\subfigure[]{\includegraphics[width=0.12\linewidth]{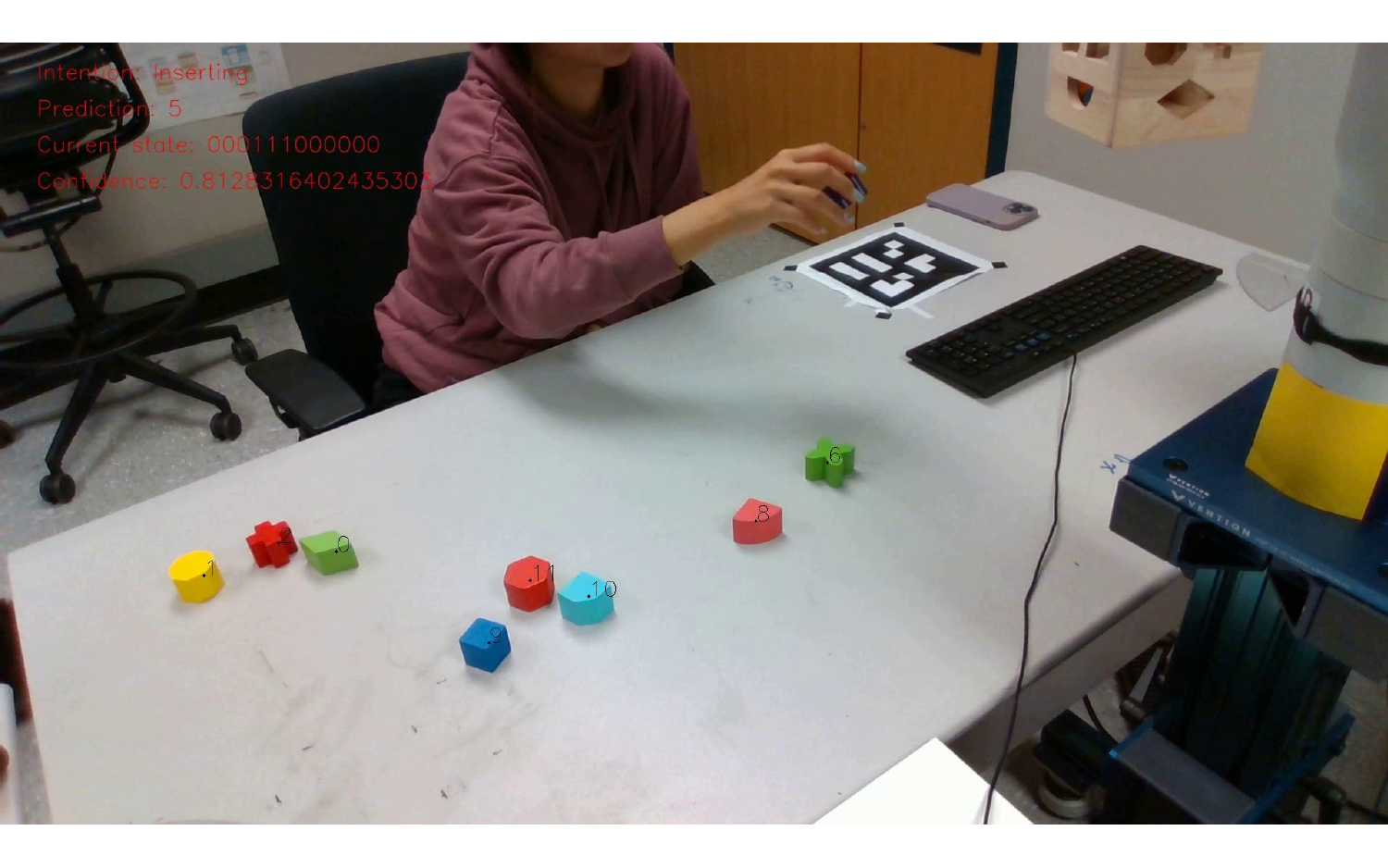}\label{fig:7_4}}\hfill
\subfigure[]{\includegraphics[width=0.12\linewidth]{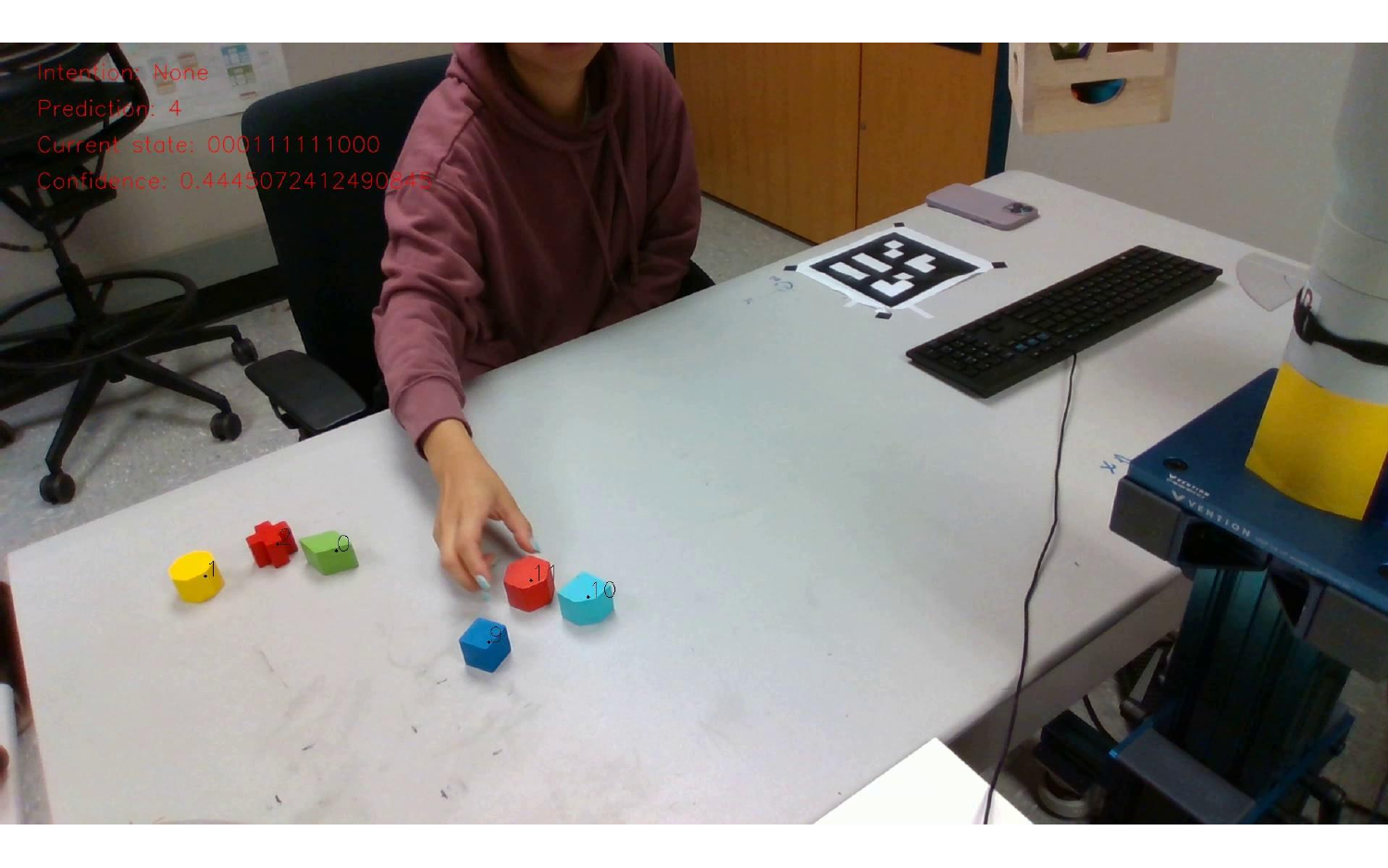}\label{fig:7_5}}\hfill
\subfigure[]{\includegraphics[width=0.12\linewidth]{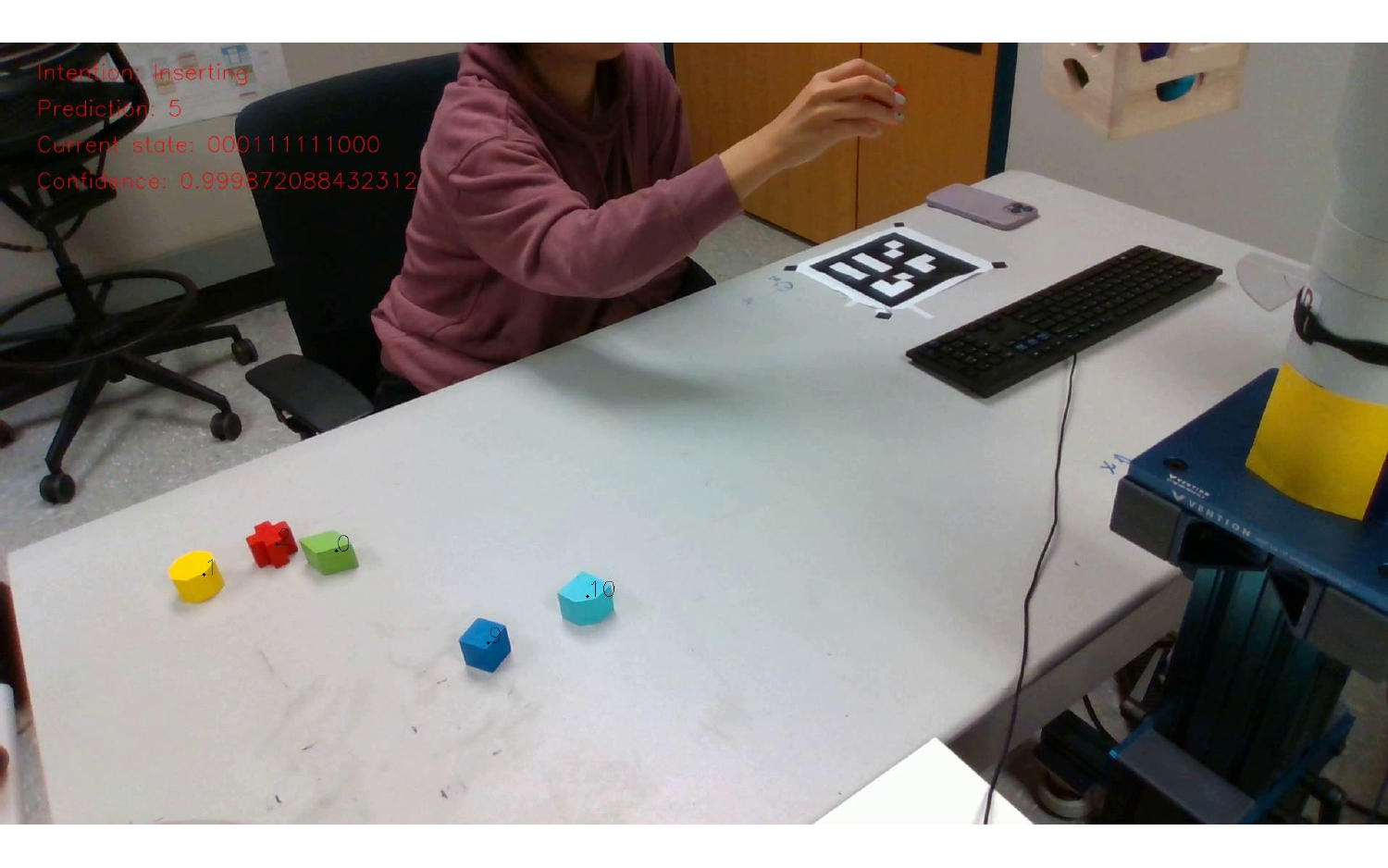}\label{fig:7_6}}\hfill
\subfigure[]{\includegraphics[width=0.12\linewidth]{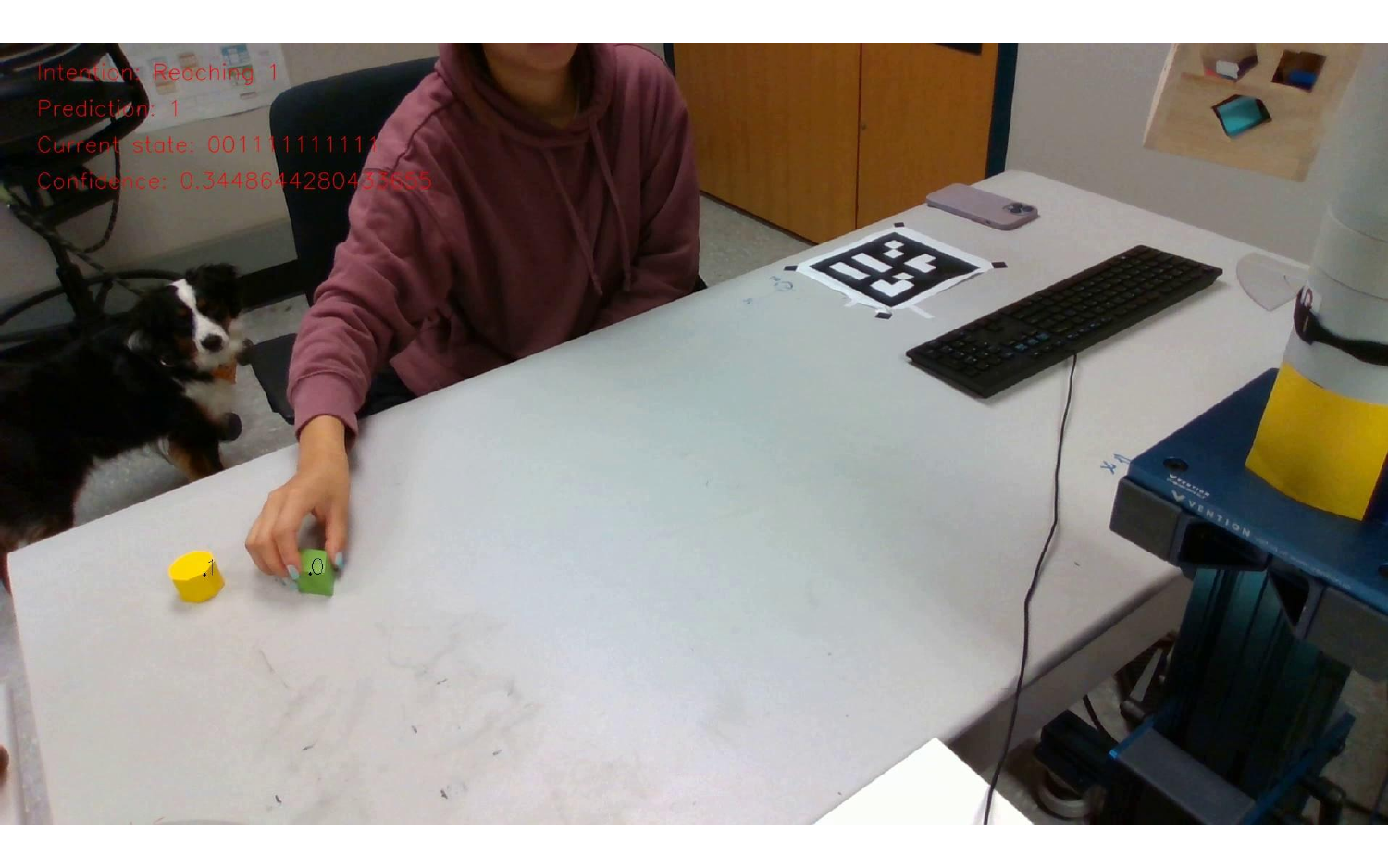}\label{fig:7_7}}\hfill
\subfigure[]{\includegraphics[width=0.12\linewidth]{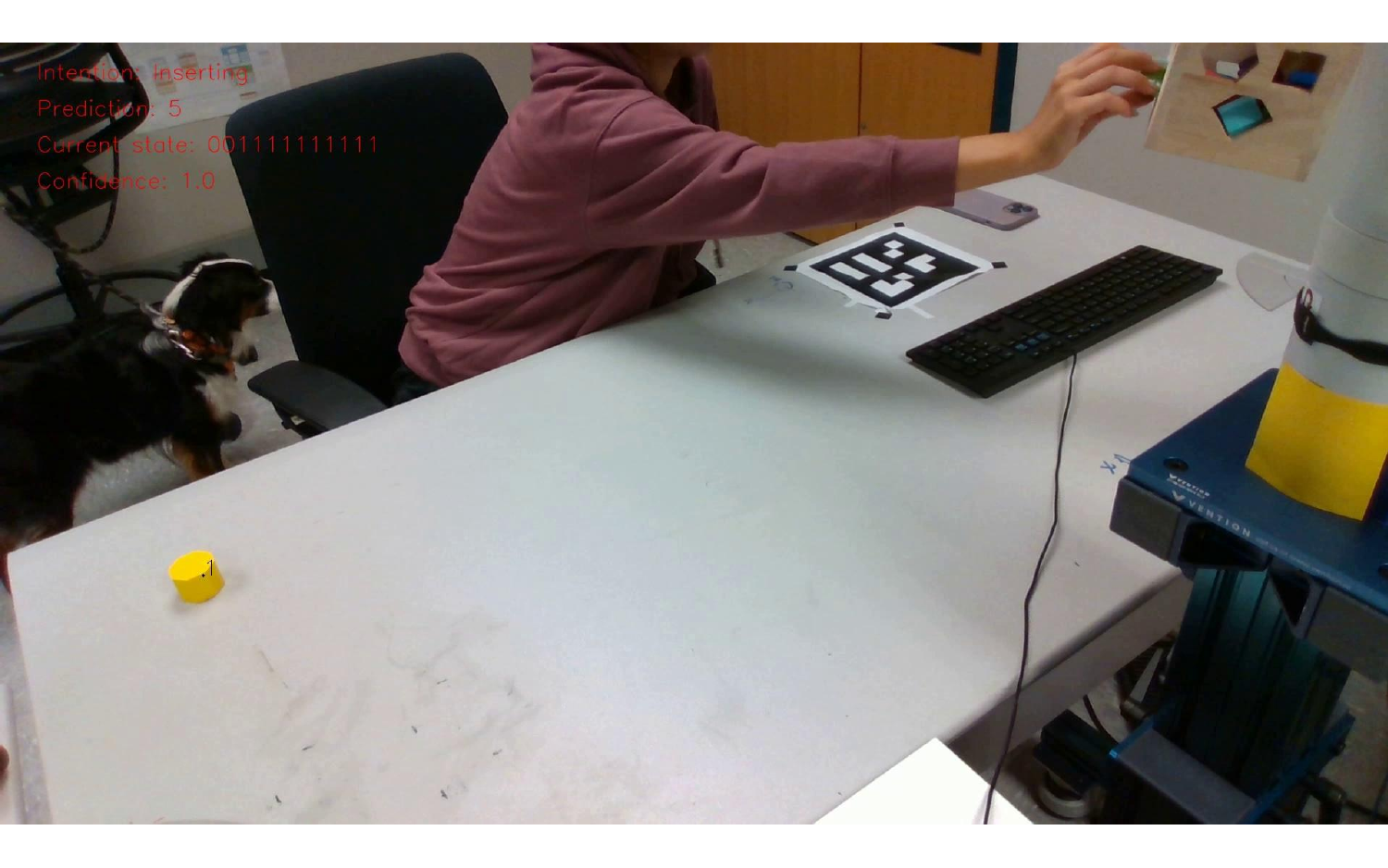}\label{fig:7_8}}\\
\vspace{-10pt}
    \caption{\footnotesize Illustration of the human-robot co-assembly. Each row is a human subject performing the co-assembly. Columns 1, 3, 5, 7: Picking up a block. Columns 2, 4, 6, 8: Inserting a block. \label{fig:robust_collab}}
    \vspace{-20pt}
\end{figure*}

\subsubsection{Human Preferences}
We want to show that our solution is robust to different human preferences.
Recall that there is flexibility in the task for the human to decide the order of insertion (\ie the order of the surfaces and the order of blocks within a surface).
Therefore, the robot should be able to accommodate individual preferences when it is assisting.
\Cref{fig:1_1,fig:1_2,fig:1_3,fig:1_4,fig:1_5,fig:1_6,fig:1_7,fig:1_8,fig:2_1,fig:2_2,fig:2_3,fig:2_4,fig:2_5,fig:2_6,fig:2_7,fig:2_8,fig:3_1,fig:3_2,fig:3_3,fig:3_4,fig:3_5,fig:3_6,fig:3_7,fig:3_8} show the same human collaborating with the robot on the assembly task. In the first row of \cref{fig:robust_collab}, the human first inserts the blocks in the second from the right cluster. Then, he inserts the second from the left cluster, followed by the right-most cluster. And eventually, he inserts the left-most cluster.
Differently in the second row of \cref{fig:robust_collab}, the human inserts the left-most cluster after finishing the second from the right cluster. Then, he inserts the right-most cluster. Lastly, he inserts the second from the left cluster.
In the third row of \cref{fig:robust_collab}, the human simply inserts blocks from the right-most cluster to the left-most.
We can see the human chooses the insertion order based on his own preference. 
Moreover, the order of insertion within a cluster/surface is randomly determined. 
The results demonstrate that the robot can successfully collaborate with humans and our solution is robust to individual preferences.
The robot understands the task well given the task graph and the intention prediction is robustly identifying the human operations.

\subsubsection{Environment Setups}
In real-world applications, the environment is constantly changing.
It is critical that the robot can collaborate robustly under different environment setups.
We consider the environment setup specifically as how the blocks are placed on the table.
\Cref{fig:2_1,fig:2_2,fig:2_3,fig:2_4,fig:2_5,fig:2_6,fig:2_7,fig:2_8,fig:4_1,fig:4_2,fig:4_3,fig:4_4,fig:4_5,fig:4_6,fig:4_7,fig:4_8,fig:5_1,fig:5_2,fig:5_3,fig:5_4,fig:5_5,fig:5_6,fig:5_7,fig:5_8} show the same human collaborating with the robot on the assembly task with different environment setups.
The results demonstrate that although the blocks are placed differently, the robot can correctly recognize the human intention and track the task progress.
Therefore, it can robustly help the human on the co-assembly task.

\subsubsection{Different Humans}
In real manufacturing, a robotic system needs to work with different human workers.
It is important that the robot can collaborate robustly with different people.
\Cref{fig:5_1,fig:5_2,fig:5_3,fig:5_4,fig:5_5,fig:5_6,fig:5_7,fig:5_8,fig:6_1,fig:6_2,fig:6_3,fig:6_4,fig:6_5,fig:6_6,fig:6_7,fig:6_8,fig:7_1,fig:7_2,fig:7_3,fig:7_4,fig:7_5,fig:7_6,fig:7_7,fig:7_8} show different humans collaborating with the robot on the co-assembly task.
Each person is doing the task differently, but because of IADA, the robot learns a robust prediction model from scarce data and robustly identifies human intention.
And based on the robust prediction, it assists different people in the co-assembly task.

\subsection{Efficient Collaboration}
In terms of task efficiency, we mainly compare the time for completing the task.
\Cref{table:time} shows the comparison of execution times between the baseline and our HRC framework. 
The table compares the average, standard deviation, minimum, and maximum time for all trials.
In the baseline, the robot passively waits for human commands.
While in our HRC framework, the robot proactively collaborates based on its inference of human intention and task progress.

\begin{table}
\centering
\begin{center}
\caption{\footnotesize Comparison of assemble time (\si{\second}).}\label{table:time}
\begin{tabular}{c  |  c  c  c  } 
\toprule 
 &  & \multicolumn{1}{c}{Baseline} & \multicolumn{1}{c}{Ours}\\  
\midrule
\multirow{2}{*}{Complete Task} & \multirow{1}{*}{Avg}
   & \multirow{1}{*}{77.80} & \multirow{1}{*}{\textbf{66.33}}\\
   & \multirow{1}{*}{Std} & \multirow{1}{*}{\textbf{11.88}} & \multirow{1}{*}{13.46}\\
\multirow{2}{*}{(\ie 12 blocks on 4 surfaces)} & \multirow{1}{*}{Min}
   & \multirow{1}{*}{66} & \multirow{1}{*}{\textbf{51}}\\
   & \multirow{1}{*}{Max} & \multirow{1}{*}{{96}} & \multirow{1}{*}{\textbf{90}}\\
\midrule
\multirow{2}{*}{Single Surface} & \multirow{1}{*}{Avg}
   & \multirow{1}{*}{18.62} & \multirow{1}{*}{\textbf{15.25}}\\
   & \multirow{1}{*}{Std} & \multirow{1}{*}{\textbf{2.73}} & \multirow{1}{*}{3.34}\\
\multirow{2}{*}{(\ie 3 blocks on 1 surface)} & \multirow{1}{*}{Min}
   & \multirow{1}{*}{15} & \multirow{1}{*}{\textbf{10}}\\
   & \multirow{1}{*}{Max} & \multirow{1}{*}{{23}} & \multirow{1}{*}{\textbf{22}}\\
\midrule
\multirow{2}{*}{Single Block} & \multirow{1}{*}{Avg}
   &  \multirow{1}{*}{8.64}  & \multirow{1}{*}{\textbf{6.36}}  \\
   & \multirow{1}{*}{Std} & \multirow{1}{*}{\textbf{1.68}} & \multirow{1}{*}{2.44}\\
\multirow{2}{*}{(\ie 1 block on 1 surface)} & \multirow{1}{*}{Min}
   & \multirow{1}{*}{6} & \multirow{1}{*}{\textbf{3}}\\
   & \multirow{1}{*}{Max} & \multirow{1}{*}{\textbf{12}} & \multirow{1}{*}{13}\\
\bottomrule
\end{tabular}
\vspace{-30pt}
\end{center}
\end{table}

First, we compare the time for inserting one single piece.
In the baseline, the human first controls the robot to move. 
After the robot finishes the motion, the human then grabs the block and inserts it.
It generally takes 2-3 seconds for the robot to move to the target position.
But in our solution, the human does not need to explicitly command the robot.
The human operator directly reaches the intended block, and the robot acts accordingly.
The robot recognizes human intention $0.5\sim1$\si{\second}, on average, before the human really grabs the block (rough estimation from recorded videos).
Therefore, we can see the time for inserting one piece is shorter by approximately 2 seconds as shown in \cref{table:time}.
After finishing one block, the human needs to finish the surface before moving to the next. 
Therefore, the robot does not need to move until the human finishes the surface.
As a result, \cref{table:time} indicates that our solution takes a shorter time than the baseline by a similar amount of time difference.
Similarly, it also takes a shorter time for completing the entire task with our HRC solution.
Since it includes multiple surfaces, we can see our solution takes a shorter time by a much larger amount.
In addition, we observe that the minimum and maximum execution times for our HRC framework are generally shorter than the baseline.
This indicates that our framework indeed improves the task efficiency.
However, \cref{table:time} shows that our framework has larger standard deviation. 
This might due to the fact that the baseline is a more consistent way that most people are familiar with.
It takes time to get used to collaborating with a proactive robot partner.
Therefore, the performance differs between trials.
Overall, the result in \cref{table:time} demonstrates that our HRC solution improves task efficiency through proactive collaboration.

\begin{figure}
\subfigure[]{\includegraphics[width=0.32\linewidth]{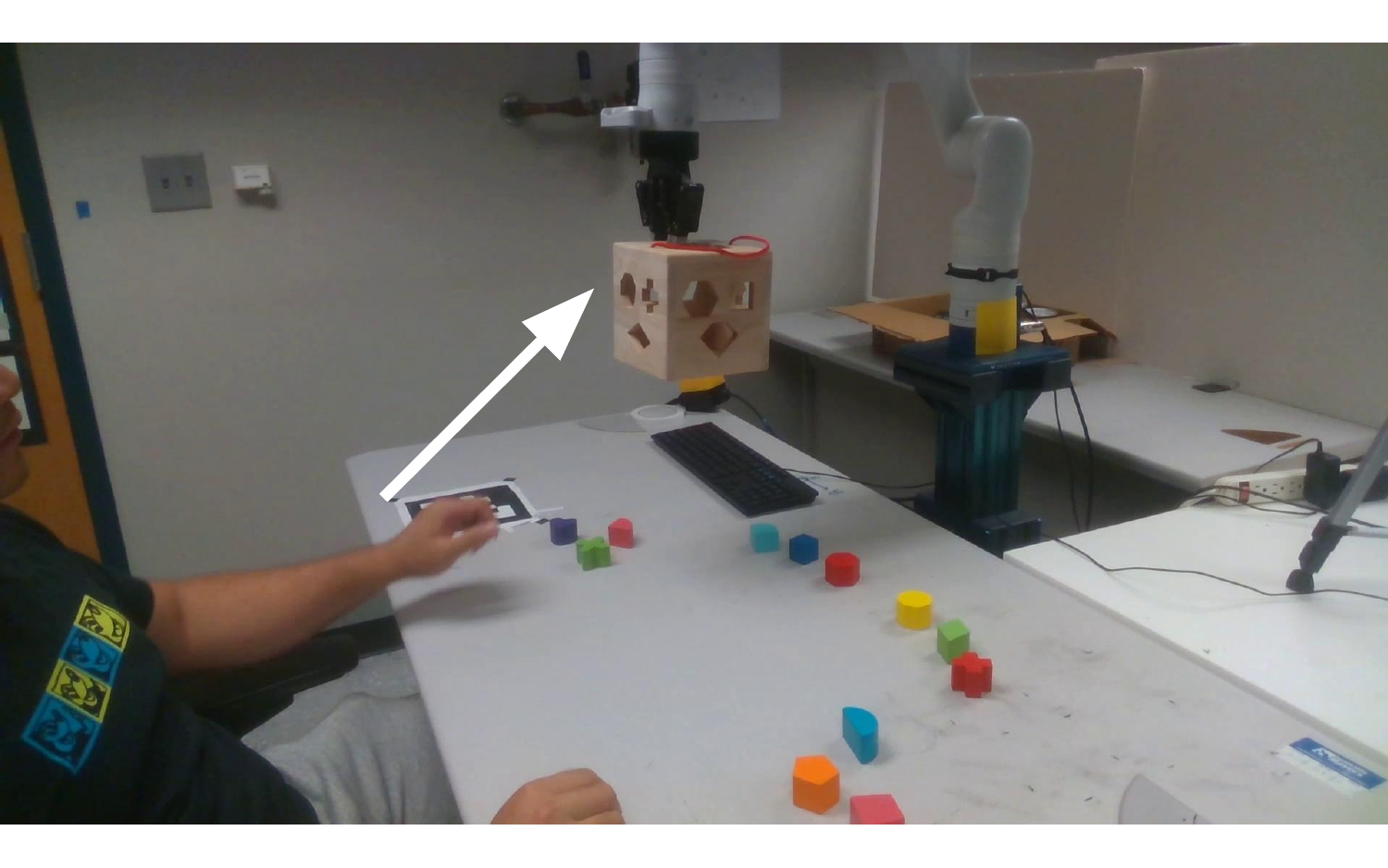}\label{fig:static1}}\hfill
\subfigure[]{\includegraphics[width=0.32\linewidth]{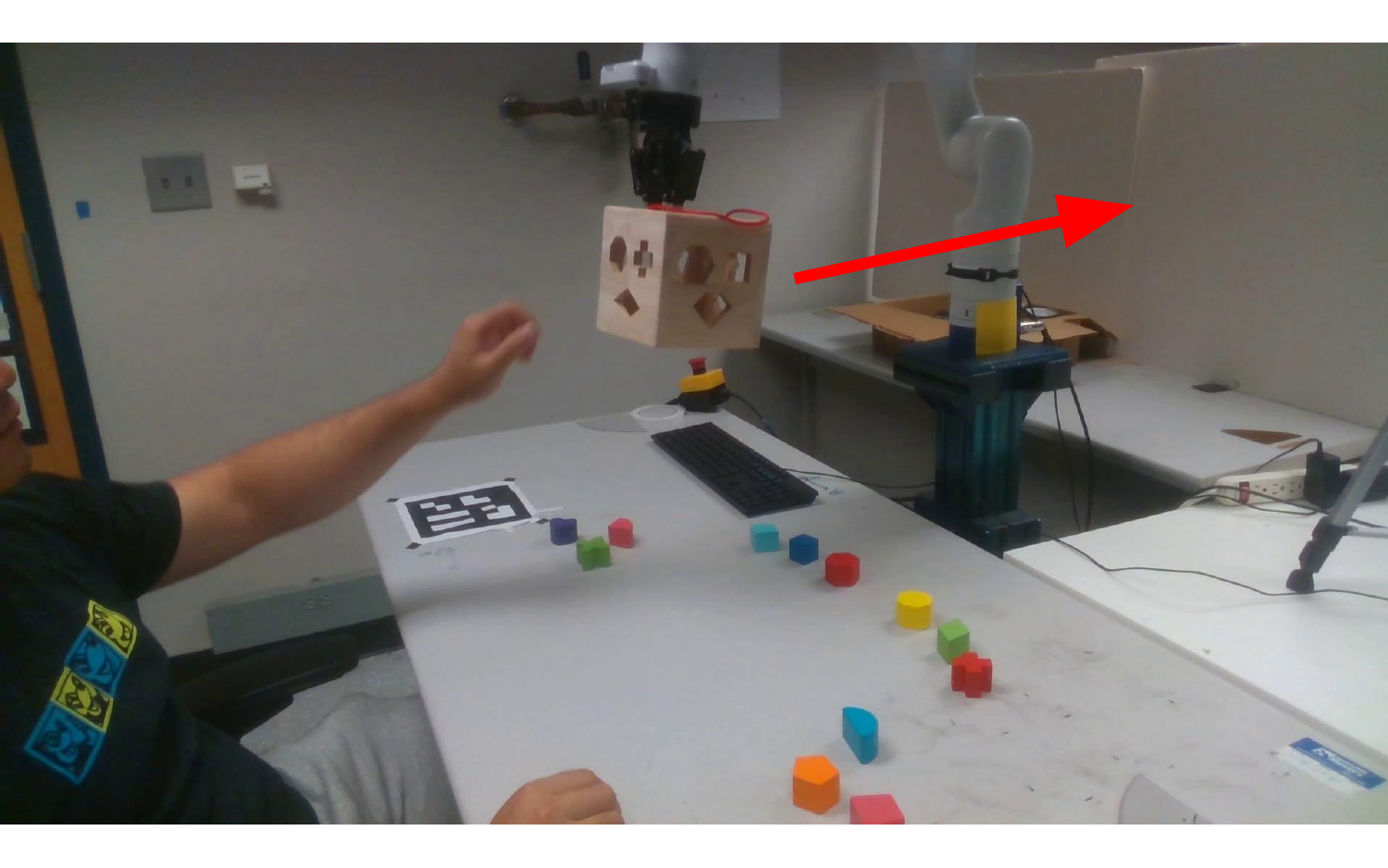}\label{fig:static2}}\hfill
\subfigure[]{\includegraphics[width=0.32\linewidth]{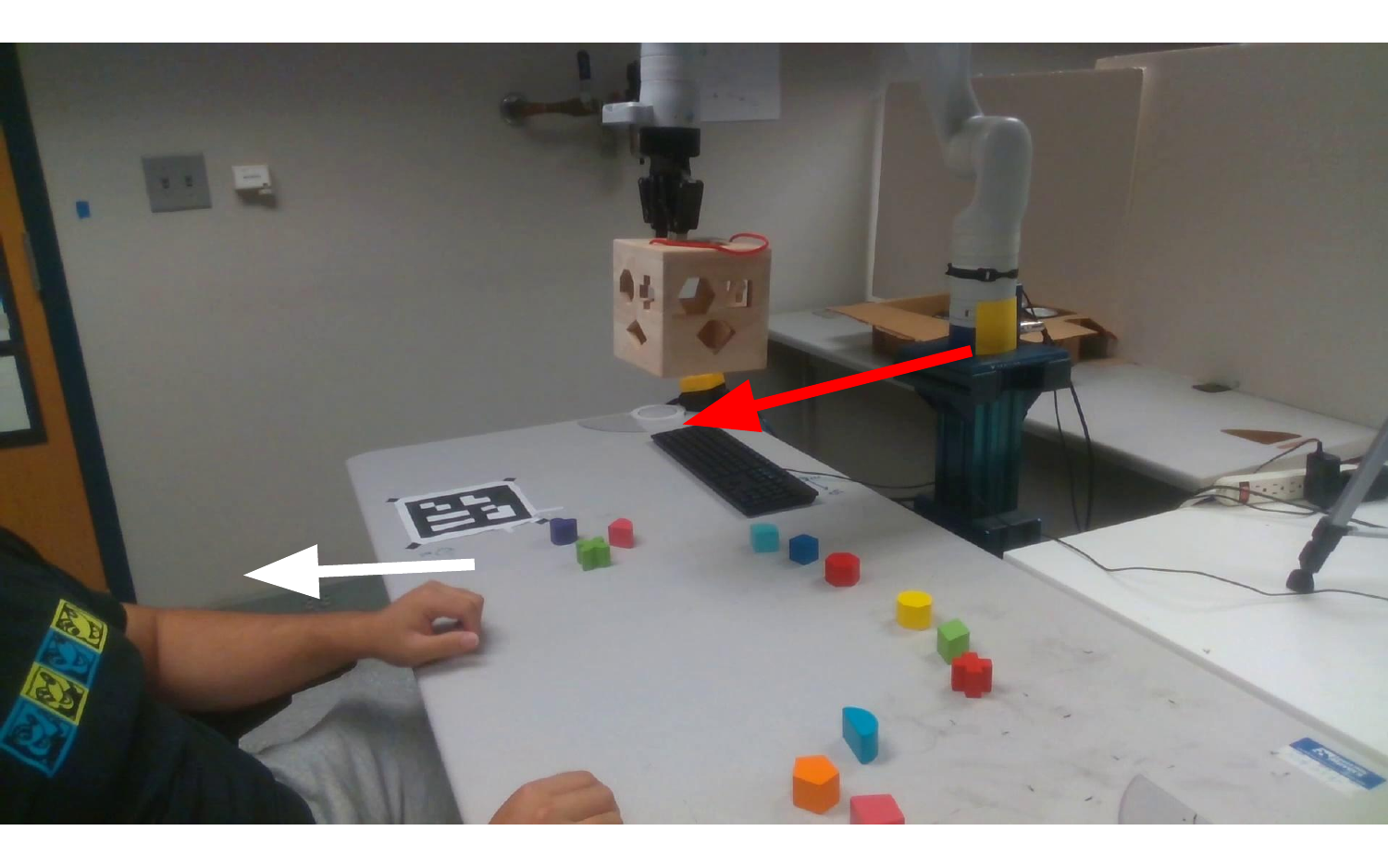}\label{fig:static3}}
\vspace{-10pt}\\
\subfigure[]{\includegraphics[width=0.32\linewidth]{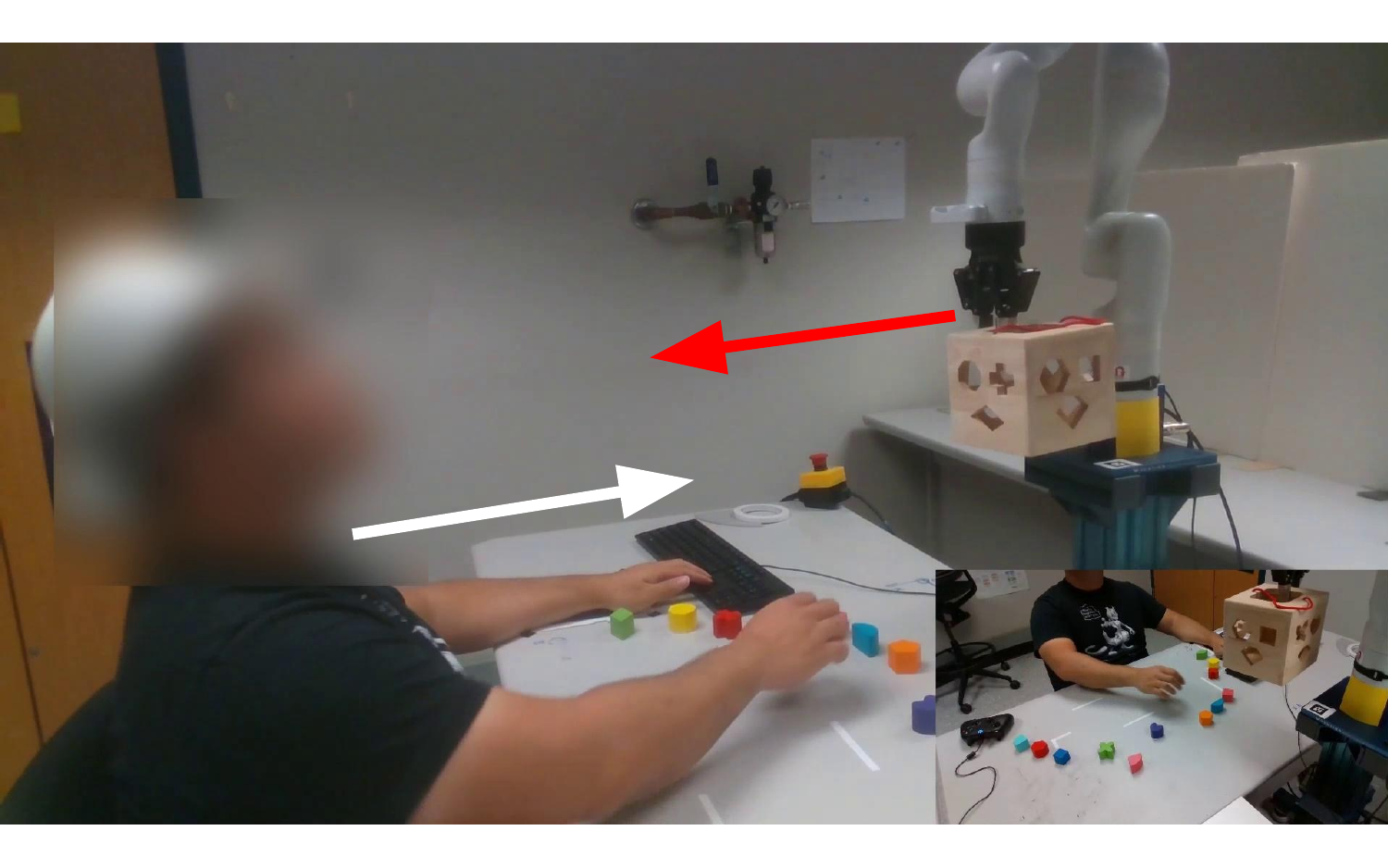}\label{fig:move1}}\hfill
\subfigure[]{\includegraphics[width=0.32\linewidth]{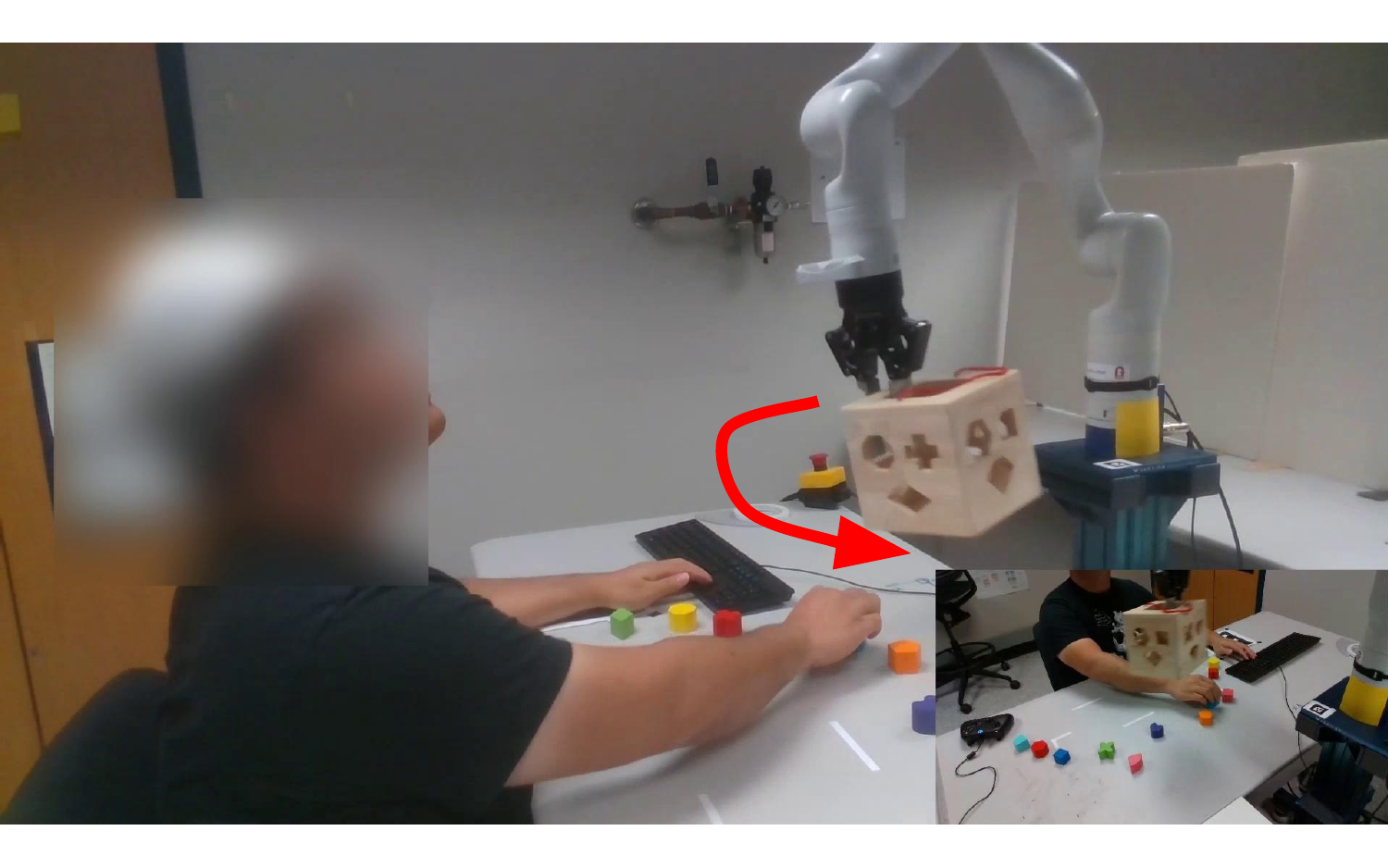}\label{fig:move2}}\hfill
\subfigure[]{\includegraphics[width=0.32\linewidth]{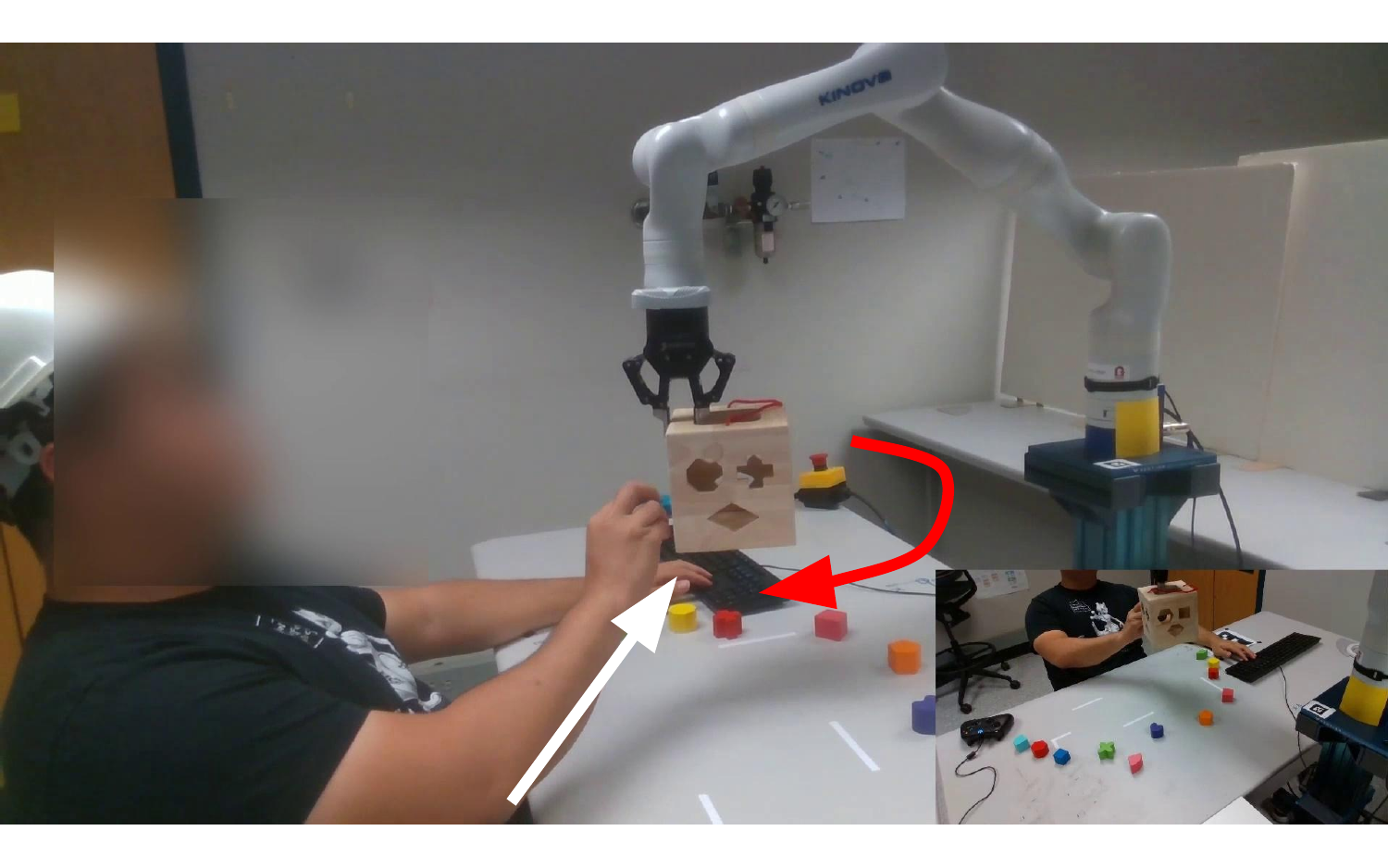}\label{fig:move3}}
\vspace{-10pt}\\
\subfigure[]{\includegraphics[width=0.32\linewidth]{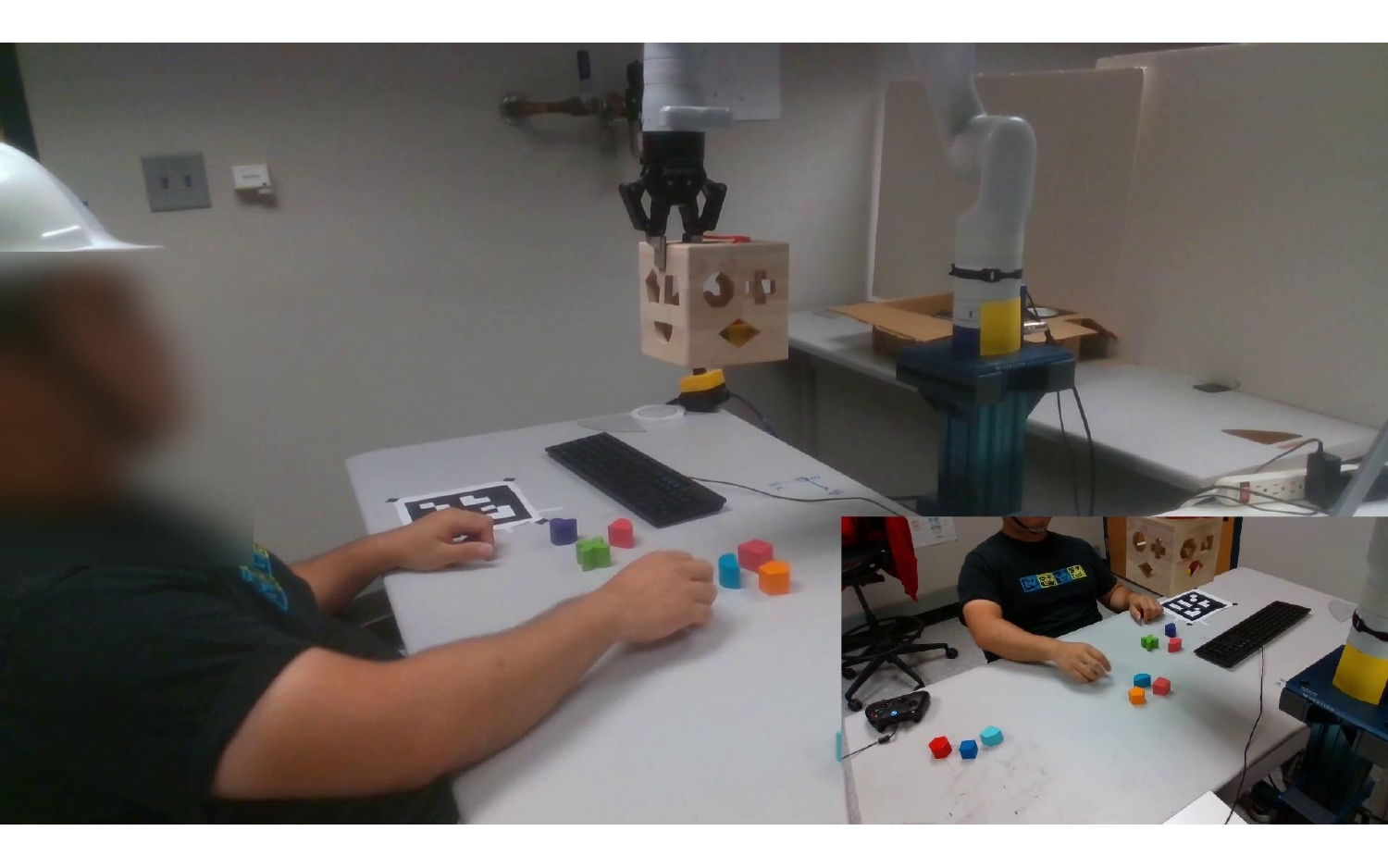}\label{fig:small1}}\hfill
\subfigure[]{\includegraphics[width=0.32\linewidth]{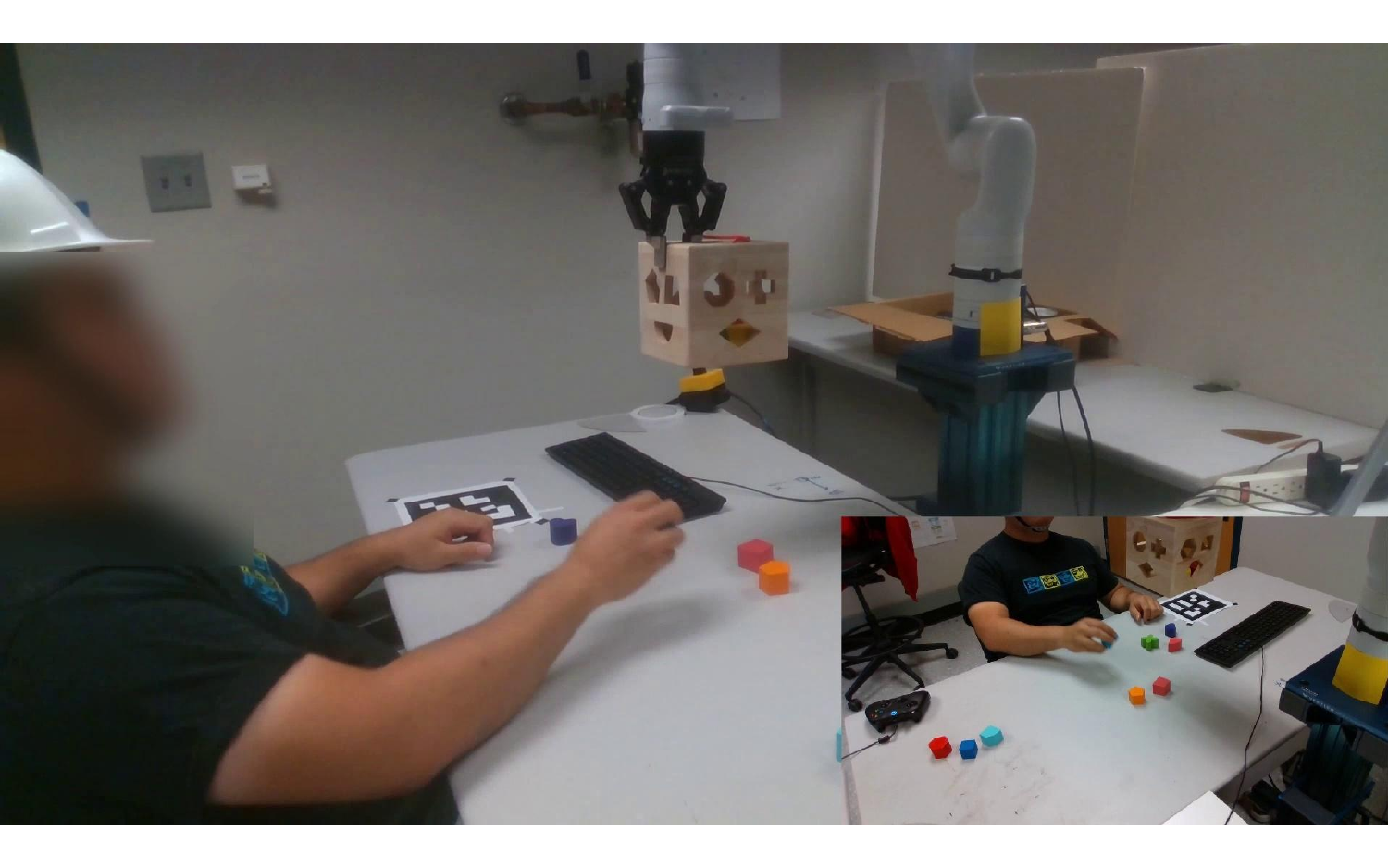}\label{fig:small2}}\hfill
\subfigure[]{\includegraphics[width=0.32\linewidth]{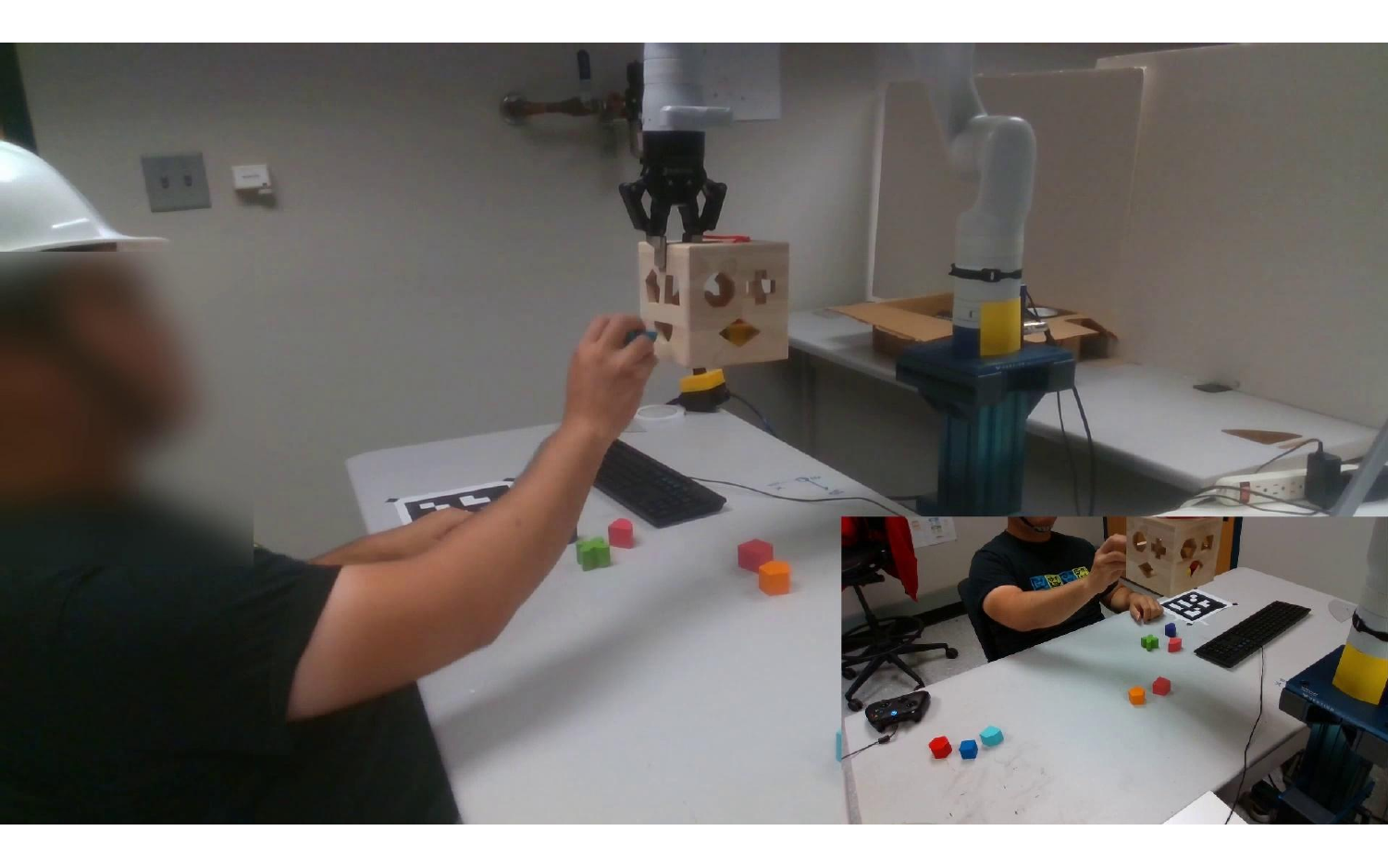}\label{fig:small3}}
\vspace{-10pt}\\
\subfigure[]{\includegraphics[width=0.32\linewidth]{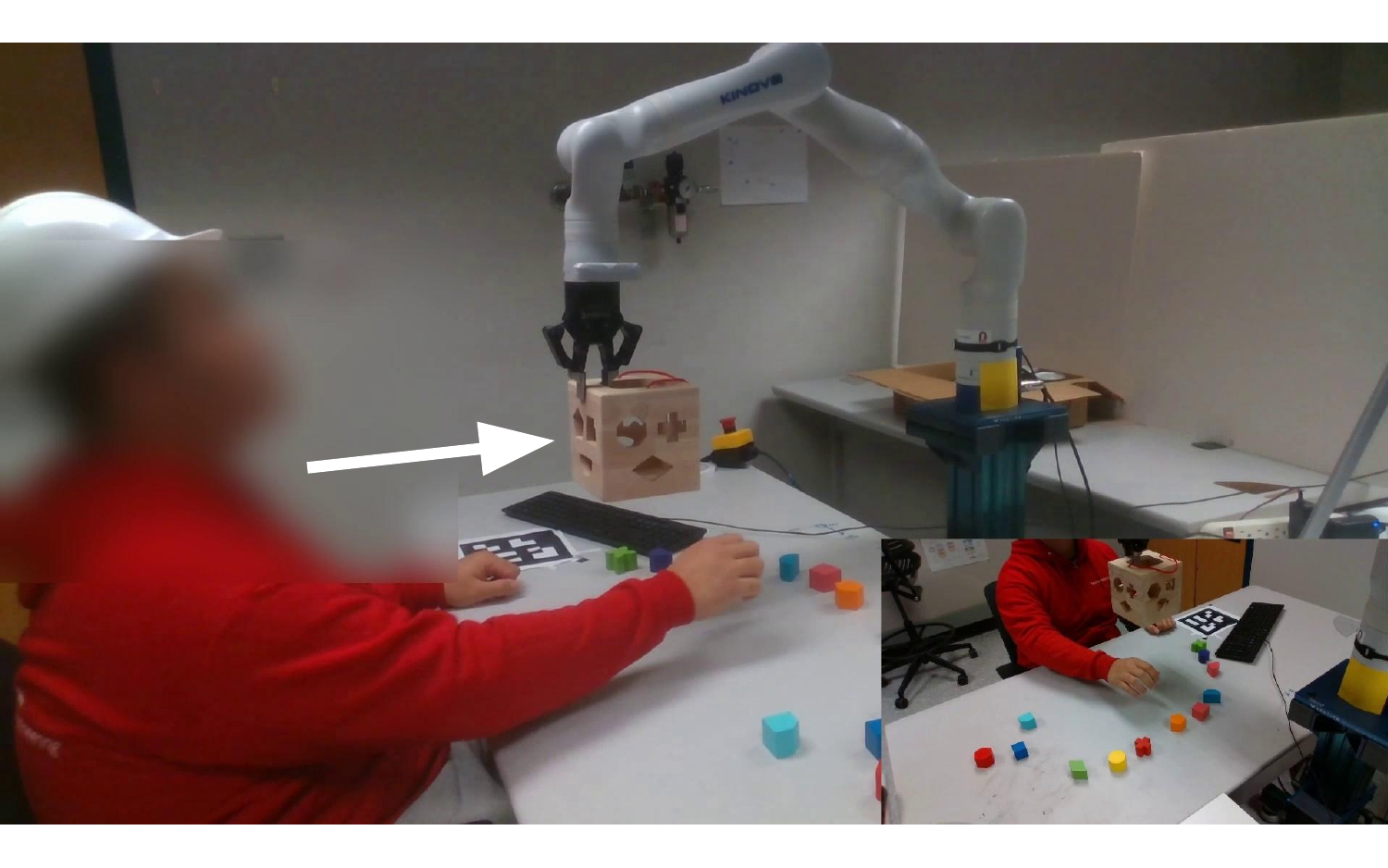}\label{fig:big1}}\hfill
\subfigure[]{\includegraphics[width=0.32\linewidth]{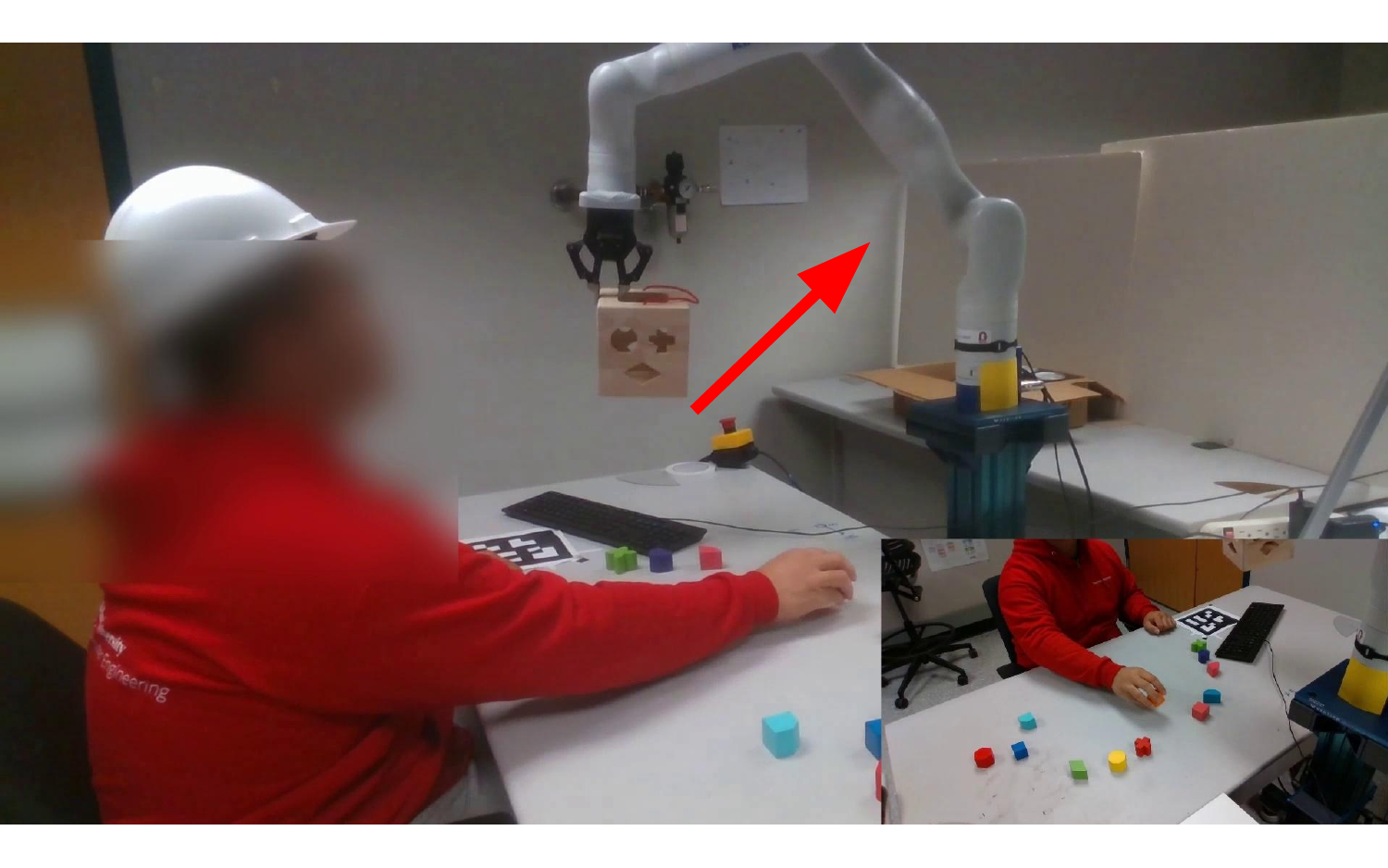}\label{fig:big2}}\hfill
\subfigure[]{\includegraphics[width=0.32\linewidth]{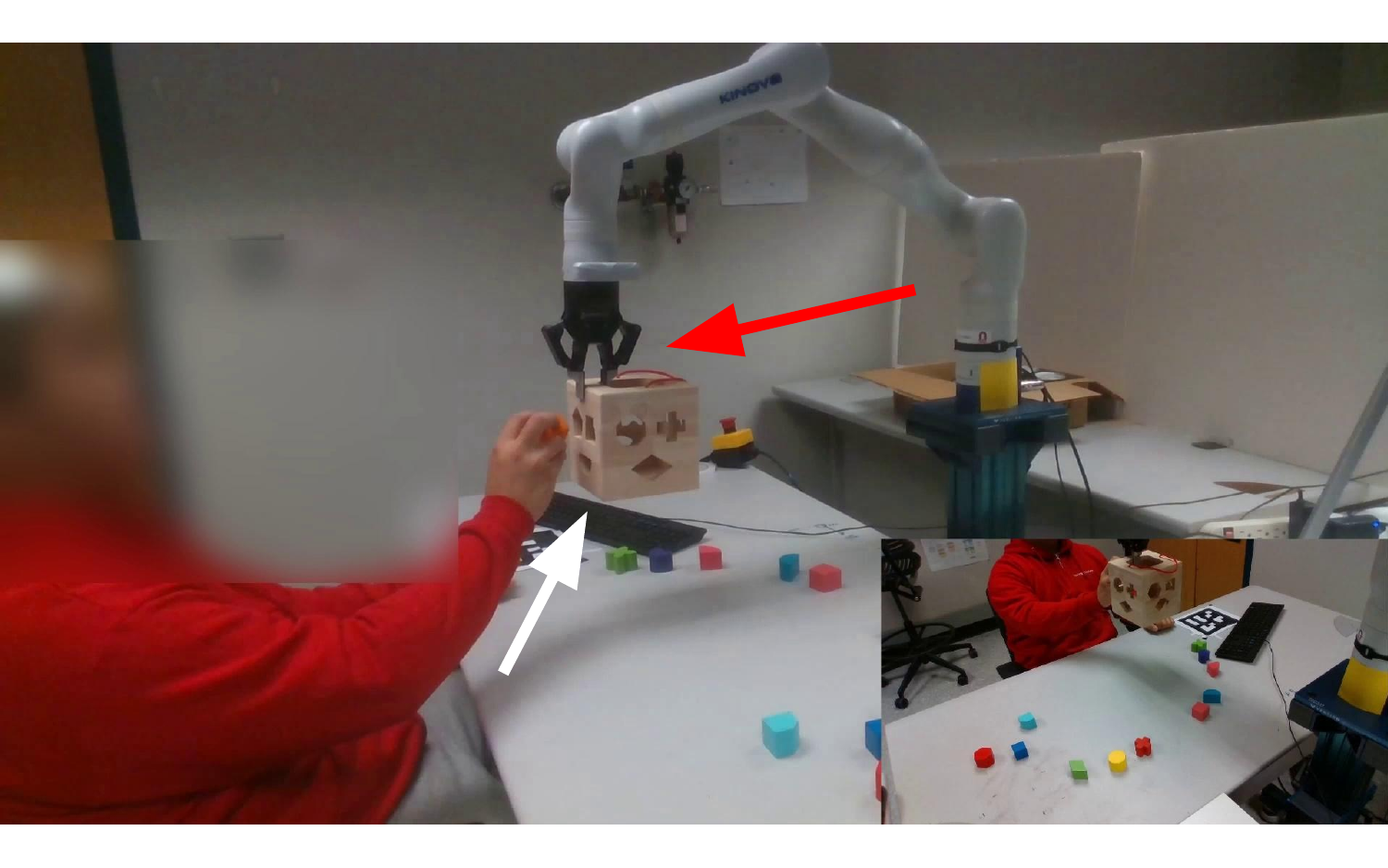}\label{fig:big3}}
\vspace{-5pt}
    \caption{\footnotesize Illustration of safe HRC. Red arrow: robot motion. White arrow: human motion. \label{fig:safe_collab}}
    \vspace{-20pt}
\end{figure}

\subsection{Safe Collaboration}
\begin{figure}
    \centering
    \input{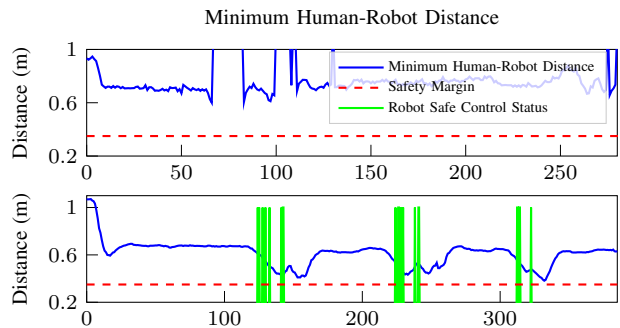}
    \vspace{-10pt}
    \caption{Minimum distance profiles between the human and robot. Top: task in \cref{fig:small1,fig:small2,fig:small3}. Bottom: task in \cref{fig:big1,fig:big2,fig:big3}. \label{fig:small_dist}}
    \vspace{-20pt}
\end{figure}

The proposed HRC framework is actively collaborating with the human based on its understanding of the human and task. 
Therefore, safety becomes a key concern.
It is critical to ensure that the robot does not harm (\ie collide) the human during the collaboration.
Due to the nature of co-assembly, it is necessary for the robot to get close to certain human body parts, \eg hands.
Intuitively, the robot should not collide with any human body parts that are not involved in the collaboration but allow close contact with body parts that are involved.
\Cref{fig:safety_specification} illustrates our safety specification in the co-assembly task. 
The red capsules represent the human body parts that should be avoided by the robot.
The green capsules indicate the parts that are allowed for contact with the robot.
We assume that humans prefer right-hand insertion, and thus, the right forearm and right hand are allowed for contact as shown in \cref{fig:safety_specification} while all other parts should be avoided by the robot. Note that the safety specification in \cref{fig:safety_specification} can be easily reconfigured for customized tasks.
The safety margin is set as $d_{min}=0.35$\si{\meter}, which is shown as the red dotted line in \cref{fig:small_dist}.
We define the safety index as $\Phi=d_{min}^2-D^2-\lambda \dot{D}$, where $D=d(q,E)$ is the monitored distance between the robot and environmental objects and $\lambda$ is a tunable parameter.

\Cref{fig:static1,fig:static2,fig:static3} demonstrate active collision avoidance by the robot.
The human tries to touch the robot using his left hand (\cref{fig:static1}).
Due to the safety specification, the robot controller monitors the environmental change (\ie human motion) and reacts to it immediately by moving away (\cref{fig:static2}). 
After the human is away (\cref{fig:static3}), the robot then goes back to its original position.
This demonstrates that the safety controller is commanding the robot to avoid collision according to our safety design.

The second row of \cref{fig:safe_collab} illustrates a more complex situation.
As shown in \cref{fig:move1}, the robot recognizes the human intention and is delivering the corresponding container surface to the human.
However, before the robot settles down, the human decides to start the task early and reaches the block while the robot is still going toward him.
In this case, we can see that the robot aborts the ongoing task immediately, and avoids the collision by going sideway (\cref{fig:move2}).
And after the human gets the block, the robot then moves to the front of the human and displays the corresponding surface for easy insertion (\cref{fig:move3}).
This demonstrates that by adding the safeguard (\ie safety controller in \cref{fig:controller}) in the control loop, the robot can respond to unexpected changes in real time.
This behavior demonstrates that our controller always prioritizes safety. 
If any potentially dangerous thing could happen, the controller would abort the task to ensure safety.
But once the hazard is gone, it commands the robot to get back to the original task.

\Cref{fig:small1,fig:small2,fig:small3,fig:big1,fig:big2,fig:big3} demonstrate safety when collaborating with different workers.
\Cref{fig:small1,fig:small2,fig:small3} show a conservative worker, who keeps himself away from the robot and only uses his arm for operation.
On the other hand, \cref{fig:big1,fig:big2,fig:big3} illustrate a proactive worker, who leans his whole body toward the robot for easy operation.
From the experiments, we can see the robot distinguishes the different behavior and act accordingly to ensure safety.
When collaborating with the conservative worker, the robot has much smaller avoiding motion since the worker's hand is allowed for close contact, and the other body parts are far away.
When collaborating with the proactive worker, the robot has a larger motion to avoid a potential collision with the worker's head.
\Cref{fig:small_dist} displays the minimum distance between the human and the robot during the collaboration.
Since the conservative worker keeps himself away from the robot, the distance profile is constant throughout the task and the robot does not need to move too much to ensure safety.
However, the proactive worker leans his body forward, and thus, the distance dramatically decreases as shown in the bottom figure of \cref{fig:small_dist}.
The green plot in \cref{fig:small_dist} displays the triggering status of the robust safety controller.
As the distance gets closer to the safety margin, the safety controller is triggered and moves the robot away to avoid potential collision as illustrated in \cref{fig:big2}.
The results demonstrate that the hierarchical controller can ensure interactive safety while accomplishing the task for HRC.


\section{CONCLUSIONS}
This paper presents an integrated HRC framework to improve collaboration efficiency while guaranteeing interactive safety. 
Our framework has an intention prediction module to understand human behavior and task information.
In particular, it leverages prior task knowledge and human-in-the-loop learning to tackle the data scarcity problem in intention prediction.
Based on the inference of human intention and task progress, the system plans the appropriate collaboration action ahead of time. 
The developed framework has a hierarchical controller, which includes a task controller and a robust safety controller, to safely execute the planned action. 
The controller monitors the robot action in real-time and ensures interactive safety.
The proposed framework is applied to a human-robot co-assembly task.
The experiment results demonstrate that our framework is robust and can improve task efficiency while guaranteeing interactive safety.

\bibliographystyle{ifacconf}
\bibliography{ifacconf}      
\end{document}